\documentclass[10pt]{article} % For LaTeX2e
\usepackage[preprint]{tmlr}

\usepackage{amsmath,amsfonts,bm}

\def\eqref#1{equation~\ref{#1}}
\def\1{\bm{1}}

\DeclareMathAlphabet{\mathsfit}{\encodingdefault}{\sfdefault}{m}{sl}
\SetMathAlphabet{\mathsfit}{bold}{\encodingdefault}{\sfdefault}{bx}{n}

\def\eqref#1{(\ref{#1})}

\usepackage{hyperref}
\usepackage{url}
\usepackage{graphicx}
\usepackage{booktabs}
\usepackage{float}

\usepackage{algorithm}
\usepackage{algpseudocode}
\usepackage{subcaption}
\usepackage{makecell}
\usepackage{booktabs}
\usepackage{multirow}

\newcommand{\hdr}[3]{\makecell{\scriptsize #1\\\scriptsize #2\\\scriptsize #3}}

\title{TriScore: Post-Hoc Out-of-Distribution Detection with \\ Energy, Boundary Probes, and Transform Consistency}

\author{\name Ege Gursoy \email egursoy@laas.fr \\
      \addr LAAS-CNRS
      \AND
      \name Nicolas Mansard \email nmansard@laas.fr \\
      \addr LAAS-CNRS\\
      ANITI}

\def\month{09}  % Insert correct month for camera-ready version
\def\year{2026} % Insert correct year for camera-ready version
\def\openreview{https://openreview.net/forum?id=H5hyDKANup} % OpenReview link is inserted by the venue/camera-ready process.

\begin{document}

\maketitle

\begin{abstract}

Post-hoc out-of-distribution detection usually relies on a single signal extracted from a frozen classifier, such as confidence, energy, features, or gradients. This dependence becomes fragile under classifier-level adversarial shift, where the shift can preserve the detector’s chosen cue while changing the model’s decision behavior. We investigate whether this failure mode can be mitigated by combining complementary detection criteria instead of selecting a single score. We introduce TriScore, a post-hoc OOD detector that combines three ID-calibrated signals from the same frozen backbone: an energy prior, a boundary-fragility score, and a transform-consistency score. These signals are standardized on ID validation data and fused by an input-dependent ID-residual gate, requiring no OOD labels, auxiliary model, training, or stored feature bank. Across four ImageNet-scale backbones, TriScore achieves the best mean AUROC among 15 post-hoc baselines under a non-adaptive adversarial-shift protocol, while remaining competitive on standard OOD benchmarks. Ablations show that the gains come from combining complementary cues rather than optimizing a single detector signal. These results support multi-criterion post-hoc detection as a practical route toward more robust OOD detection under classifier-level shift. The source code is publicly available at \url{github.com/gursoyege/TriScore}.

\end{abstract}

\section{Introduction}

% Vision classifiers are often deployed in environments with inputs outside the training distribution. A useful OOD detector should identify such inputs without retraining the classifier, since many practical models are frozen backbones reused across tasks. Post-hoc OOD detection addresses this case. The detector observes the logits, features, or gradients of a fixed classifier and outputs a scalar score, where larger values indicate stronger evidence of OOD input.

Modern vision systems often reuse a trained backbone across datasets, tasks, and deployment settings. Post-hoc out-of-distribution (OOD) detection methods help in those cases, where the classifier has already been fixed and the detector must rely on the outputs or internal signals of the existing classifier without retraining it or using OOD data.

Many post-hoc detectors are built around a primary cue, such as softmax confidence~\citep{hendrycks2017baseline}, logit energy~\citep{liu2020energy}, activation statistics~\citep{sun2021react,djurisic2023ash}, feature distance~\citep{sun2022knn}, or gradient magnitude~\citep{huang2021gradnorm}. Each cue emphasizes a different property of the same classifier. A detector built around only one of these cues therefore has little redundancy when a shift happens to preserve that particular statistic, even when other classifier signals are affected.

This lack of redundancy is most visible under classifier-level adversarial shift. These are not detector-adaptive attacks, but they provide a demanding stress test for post-hoc detectors, since the image remains close to an in-distribution sample while the classifier response changes by design. In this setting, several strong baselines become difficult to distinguish from random guessing on at least one backbone, and the backbone on which a baseline collapses differs across baselines.

TriScore addresses this lack of redundancy, extracting three signals from the same frozen classifier. The energy branch measures abnormal logit evidence after subtraction of a class-mean logit vector estimated from ID calibration data. The Boundary-Probe branch estimates local decision fragility from a top-$k$ logit margin and the gradient of that margin at a late feature representation. The transform-consistency branch measures whether mild deterministic views lead to stable predictive distributions. These branches are complementary as they fail for different reasons: Energy is cheap and strong on many semantic shifts, Boundary-Probe responds to locally fragile decisions even when the logit magnitude remains plausible, Transform consistency detects predictions that are unstable under benign views.

\begin{figure}[t]
  \centering
  \includegraphics[width=1.0\linewidth]{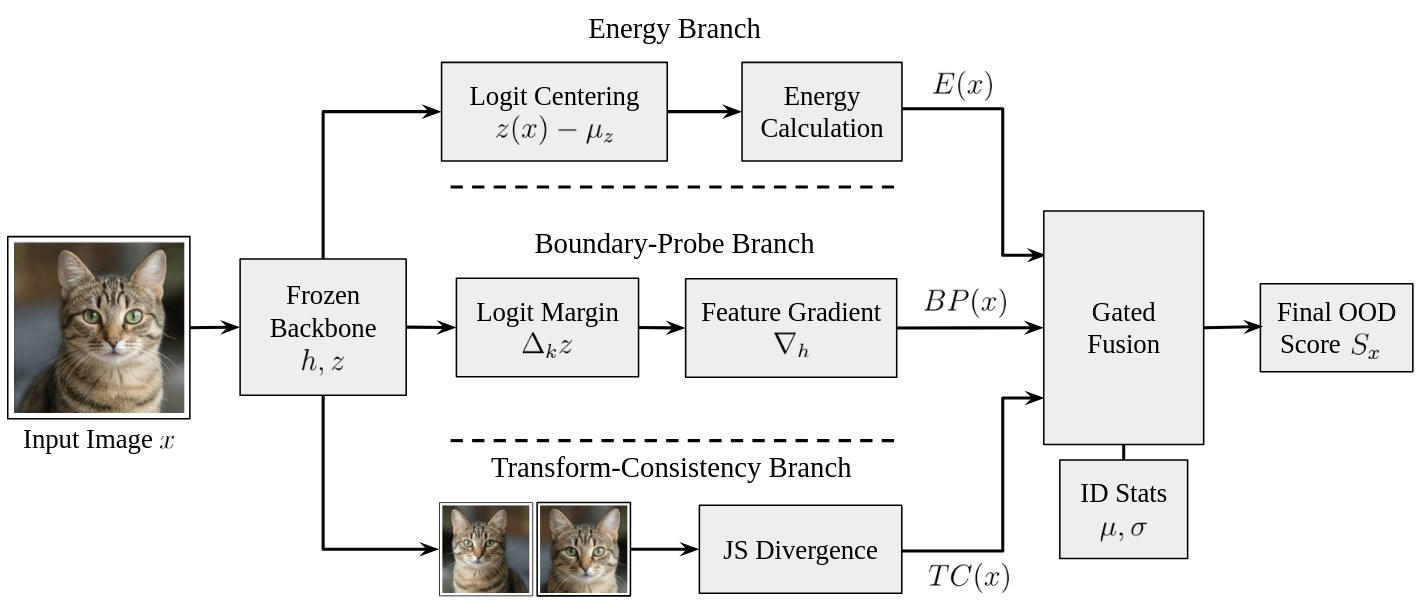}
  \caption{TriScore attaches to a frozen classifier. It computes class-centered energy, Boundary-Probe fragility, and transform consistency, then fuses the standardized branch scores using an ID-residual gate. Calibration uses ID validation data only.}
  \label{fig:overview}
\end{figure}

After the three signals are computed, TriScore fuses them with a calibrated rule. Each branch is calibrated on ID validation data, which determines the typical behavior for that branch. At test time, the final score gives more weight to branches whose standardized score exceeds the ID-typical range. Calibration does not use OOD examples, and the method does not train an auxiliary model. An overview of our method is shown in Figure~\ref{fig:overview}.

We evaluate TriScore in the OpenOOD framework on four ImageNet-scale backbones: ResNet-50, RegNet, ViT-B/16, and Swin-T~\citep{he2016deep,radosavovic2020regnet,dosovitskiy2021vit,liu2021swin}. The evaluation uses near-OOD and far-OOD datasets ~\citep{bitterwolf2023ninco,vaze2022ssb,vanhorn2018inaturalist,cimpoi2014dtd,wang2022vim,zhang2023openood15}, and adversarial-shift data ~\citep{goodfellow2014explaining,kurakin2018adversarial,madry2017towards,dong2018boosting,zheng2019distributionally,croce2020reliable,carlini2017towards,moosavi2016deepfool,andriushchenko2020square,athalye2018obfuscated}.

TriScore provides a stronger adversarial-shift robustness floor than the individual post-hoc methods. The per-attack analysis shows that this result is not driven by a single attack, and the transformer results show lower cross-attack variance than the closest competitors.

% The main finding is that TriScore provides a stronger adversarial-shift robustness floor than the individual post-hoc methods. It achieves the best mean AUROC among 15 baselines on all four backbones and the best mean FPR@95 on three of four backbones in this protocol. The per-attack analysis shows that this result is not driven by a single attack, and the transformer results show lower cross-attack variance than the closest competitors.

Our contribution is a practical post-hoc detector for frozen classifiers, together with a controlled study of how energy, boundary fragility, and consistency under deterministic transforms interact under clean and adversarial shift. We make the following claims: (i) TriScore combines three frozen-backbone signals under ID-only calibration and remains compatible with standard OpenOOD post-hoc evaluation.
(ii) under our non-adaptive adversarial-shift protocol, TriScore gives the best mean AUROC among 15 post-hoc baselines on all four ImageNet-scale backbones, and the best mean FPR@95 on three of four.
(iii) the gain is broad across attack families. TriScore wins 27 of 40 attack$\times$backbone AUROC cells and 19 of 40 FPR@95 cells, and it has lower cross-attack AUROC standard deviation on multiple backbones.
(iv) the ablation study shows that the proposed branches improve over their closest predecessor-style alternatives and that the fusion provides an ID-only way to combine them.

\section{Related Work}
\label{sec:related}

Post-hoc OOD detection scores a frozen classifier without parameter changes. The earliest and still widely used baseline is maximum softmax probability, which treats low maximum confidence as OOD evidence~\citep{hendrycks2017baseline}. ODIN improves this output-space view with temperature scaling and a small input perturbation~\citep{liang2018odin}. Later work showed that logits often carry stronger OOD information than normalized softmax probabilities. The maximum logit score uses the largest pre-softmax logit~\citep{hendrycks2022scaling}, energy-based detection uses log-sum-exp free energy~\citep{liu2020energy}, and GEN studies generalized entropy scores at scale~\citep{liu2023gen}. These methods are efficient and architecture-agnostic, which makes them strong baselines for any post-hoc method.

A second line of work uses internal network information rather than only final logits. GradNorm uses gradients from a uniform-target objective as an OOD signal~\citep{huang2021gradnorm}. ReAct and ASH modify late activations by clipping or sparsification before scoring~\citep{sun2021react,djurisic2023ash}. DICE changes the effective classifier weights at test time~\citep{sun2022dice}, and SCALE revisits activation shaping through activation rescaling~\citep{xu2024scale}. Feature-space methods use the geometry of ID representations. KNN measures nearest-neighbor distance in deep feature space~\citep{sun2022knn}. ViM combines the residual outside a principal ID feature subspace with the original logits~\citep{wang2022vim}. NCI uses neural-collapse-inspired class geometry~\citep{liu2025nci}, while CADRef measures class-aware feature discrepancies relative to classifier-weight directions~\citep{ling2025cadref}. TriScore is motivated by the fact that these cues are useful, but each one is incomplete on its own.

The consistency branch is related to test-time augmentation and distributional disagreement. Test-time augmentation often improves prediction or calibration under benign shifts~\citep{shanmugam2021better}, and Jensen-Shannon divergence is a standard symmetric comparison between predictive distributions~\citep{lin1991divergence}. We use this idea in a post-hoc setting. Our branch compares a base prediction with two mild deterministic views and treats distributional instability as OOD evidence.

Several OOD methods alter the classifier during training. Outlier Exposure trains with auxiliary outlier data~\citep{hendrycks2019oe}, and energy-based losses can also serve as training objectives~\citep{liu2020energy}. These approaches are highly effective when retraining and auxiliary OOD data are allowed. They are outside our deployment setting, where the classifier is frozen and the detector must be added after training.

Robustness-oriented OOD detection is closest to our motivation. PRO searches locally around each input for a perturbation that minimizes the OOD score, then uses the resulting score as a robustness-aware detector~\citep{chen2025pro}. TriScore instead uses fixed forward passes and late-feature backward passes rather than iterative test-time search. The adversarial stress tests in our evaluation draw from the standard attack literature, with FGSM, BIM, PGD and momentum variants, AutoAttack and APGD, C\&W, DeepFool, Square, and BPDA-based attacks~\citep{goodfellow2014explaining,kurakin2018adversarial,madry2017towards,dong2018boosting,zheng2019distributionally,croce2020reliable,carlini2017towards,moosavi2016deepfool,andriushchenko2020square,athalye2018obfuscated}. Certified robust OOD detection has also been studied~\citep{meinke2022provably}, but our work makes an empirical post-hoc claim, not a certified one.

We follow current OpenOOD and full-spectrum evaluation practice~\citep{yang2022openood,zhang2023openood15,yang2022fsood}. For adversarial generation, the evaluation uses the OpenOOD attack-data interface which uses AutoAttack and Foolbox to generate these attacks~\citep{rauber2017foolbox,croce2020reliable}. ImageNet-scale near-OOD evaluation uses SSB-Hard and NINCO~\citep{vaze2022ssb,bitterwolf2023ninco}, and far-OOD evaluation uses iNaturalist, Textures, and OpenImage-O~\citep{vanhorn2018inaturalist,cimpoi2014dtd,wang2022vim,zhang2023openood15}. Smaller-scale OSR and generalized OOD studies use CIFAR-10, Tiny ImageNet, and MNIST~\citep{krizhevsky2009learning,le2015tiny,lecun2002gradient,yang2021oodsurvey,vaze2021open}. Since post-hoc detector behavior changes with architecture, we evaluate both convolutional and transformer backbones, with ResNet, RegNet, ViT, and Swin~\citep{he2016deep,radosavovic2020regnet,dosovitskiy2021vit,liu2021swin}.

In our method, the energy branch adapts EBO~\citep{liu2020energy} with subtracting an ID-estimated class-mean logit vector (Eq.~\eqref{eq:energy_def}). The Boundary-Probe is based on standard top-\(k_m\) margin, the specificity of the branch is the ratio of the margin to its late-feature gradient as one fragility scalar (Eq.~\eqref{eq:bp_distance}), which differs from GradNorm's uniform-target gradient magnitude~\citep{huang2021gradnorm} and from margin-only scores. The consistency branch adapts test-time augmentation and Jensen-Shannon divergence~\citep{shanmugam2021better,lin1991divergence} to a post-hoc OOD score. Another new element is the fusion, which is a per-sample, OOD-label-free combiner keyed on how far each standardized branch exceeds its ID-tail quantile (Section~\ref{subsec:egf}). Our method contributes through the integration and design of the ID-residual gate, together with the empirical finding (Section~\ref{subsec:adv_results}) that diverse ID-only fusion provides a robust detection method. Appendix~\ref{app:altfusion} further investigates whether the same baseline can be achieved by fusing other existing detectors.

\section{Methodology}
\label{sec:method}

We consider post-hoc OOD detection for a frozen image classifier
\(f:\mathbb{R}^{C\times H\times W}\to\mathbb{R}^{K}\) with logits
\(z(x)\in\mathbb{R}^{K}\). TriScore assigns a scalar score \(S(x)\), where larger values indicate stronger evidence that \(x\) is OOD. The classifier weights are not updated, and calibration uses only an ID validation split. This setting matches common deployment constraints in which a trained classifier is available, but OOD samples and auxiliary outlier data are not.

We combine three calibrated views of the same frozen classifier. The Energy (E) branch measures whether the logit vector looks atypical after removing persistent class-wise offsets. The Boundary-Probe (BP) branch estimates local decision fragility by combining a top-\(k\) margin with the gradient of that margin at a late feature representation. The Transform-Consistency (TC) branch measures whether the predictive distribution changes under mild deterministic views. These branches are normalized on ID data and fused with an ID-residual gate that gives more weight to whichever branch is most atypical for the current input.

In the rest of this section, we write \(z(x)\in\mathbb{R}^K\) for the frozen classifier logits, \(\mathcal{D}_{\mathrm{ID}}\) for the ID validation split used for all calibration, and \(\mathbb{T}_{\mathrm{geo}}=\{\mathrm{id},\mathrm{flip},\mathrm{crop}_{0.9}\}\) for the deterministic transform pool (Section~\ref{subsec:consistency}) shared by the Boundary-Probe and transform-consistency branches. Every calibration constant is an empirical statistic of a single branch on \(\mathcal{D}_{\mathrm{ID}}\) such as, the class-mean logit vector \(\mu_z\) (Eq.~\eqref{eq:energy_def}), the branch means and standard deviations \((\mu_\cdot,\sigma_\cdot)\) used for z-scoring (Eq.~\eqref{eq:zscore}), and the branch quantile thresholds \(\kappa_\cdot\) used by the gate (empirical \(p_\cdot\) quantiles of the standardized branch on \(\mathcal{D}_{\mathrm{ID}}\)). No OOD or adversarial data enters any of these estimates, and an ID-only sensitivity study is given in Appendix~\ref{app:within_ablation} and~\ref{app:sensitivity}.

Two assumptions underlie the method: 

\textbf{Assumption 1:} The ID calibration split $\mathcal{D}_{\mathrm{ID}}$ should represent the deployment-time ID distribution, since it determines the branch means, standard deviations, and quantiles. In implementation we use a small standard-deviation floor $\sigma_{\min}=10^{-6}$ so that a nearly constant ID branch cannot create unstable standardized scores. 

\textbf{Assumption 2:} The transform pool should be mostly label preserving for ID inputs. Horizontal flips and mild crops are usually safe for natural images, but not for all classes, especially classes involving orientation, digits, or readable text. We therefore treat transform consistency as an empirical cue rather than an invariance guarantee.

\subsection{Energy on class-centered logits}
\label{subsec:energy}

An energy signal is a scalar summary of the full logit vector. Instead of keeping only the largest softmax probability, it uses log-sum-exp to combine the evidence assigned to all classes. For classifier logits \(z(x)\), the standard raw energy score is
\begin{equation}
    E_{\mathrm{raw}}(x;T)
    = -T \log \sum_{k=1}^{K}\exp\!\Big(\frac{z_k(x)}{T}\Big)
    \label{eq:energy_raw}
\end{equation}
This is the standard energy-based OOD formulation used by EBO~\citep{liu2020energy}. It is efficient as it only requires the logits already produced by the classifier, and it is stronger than maximum softmax probability in many large-scale settings as it preserves logit-scale information~\citep{hendrycks2022scaling}.

The limitation is that raw logits in a frozen classifier can include persistent class-wise offsets inherited from training. Some classes receive systematically larger logits on ID data even when the evidence for a particular input is not stronger. A raw energy score can therefore mix input-level OOD evidence with reusable class-level logit bias. This is related to the logit-scale calibration effects studied in recent OOD work~\citep{djurisic2024logitscaling}.

Our addition is to keep the standard energy form but remove the ID-estimated class-wise offset before computing it. The calibration mean \(\mu_z\), centered logits \(\tilde z(x)\), and final TriScore energy branch \(E(x;T)\) are
\begin{equation}
\begin{gathered}
    \mu_z = \mathbb{E}_{x\in\mathcal{D}_{\mathrm{ID}}}[z(x)] \in\mathbb{R}^{K},
    \qquad
    \tilde z(x) = z(x) - \mu_z\\
    E(x;T) = -T \log \sum_{k=1}^{K}\exp\!\Big(\frac{\tilde z_k(x)}{T}\Big)
\end{gathered}
    \label{eq:energy_def}
\end{equation}
The energy branch answers whether the whole logit vector has unusual evidence after removing the average ID class bias. The role of the centering is to make the energy signal less tied to reusable ID class offsets before it is combined with other signals.

\subsection{Boundary-Probe on late features}
\label{subsec:control}

A boundary signal measures how close the current prediction is to a competing decision region. The basic logit-space definition is a margin between the largest logit and plausible alternatives. Let \(z_{(1)}(x)\ge z_{(2)}(x)\ge\cdots\ge z_{(K)}(x)\) denote sorted logits. We use the top-\(k_m\) margin
\begin{equation}
    \Delta(x)=z_{(1)}(x)-\frac{1}{k_m-1}\sum_{i=2}^{k_m}z_{(i)}(x)
    \label{eq:topk_margin}
\end{equation}
A top-\(k_m\) average is more stable than a pure top-2 margin in large-class problems as it averages several plausible competitors. We use $k_m=5$ by default. The operation is piecewise differentiable; thus, we use the standard autograd subgradient, and exact ties are negligible in floating-point inference.

The standard margin method uses the negative margin as OOD evidence, since a smaller margin indicates a more ambiguous prediction. A standard gradient method, such as GradNorm~\citep{huang2021gradnorm}, instead uses a gradient magnitude as a sensitivity cue. These baselines isolate two different aspects of boundary behavior.

The limitation is however, the margin-only and gradient-only scores are incomplete. The margin says how much separation exists at the current point, but not how quickly that separation changes. The gradient says how sensitive the margin is, but not whether there is enough separation to begin with. Two inputs with the same margin can have different local fragility, and two inputs with the same gradient norm can have different boundary costs.

Instead, we combine the margin and its late-feature gradient into a first-order fragility proxy. Let \(h(x)\) be the selected late feature representation. For CNNs this is the final convolutional block or the pooled representation feeding the head, and for transformer backbones it is the final block representation used by the classifier head. Gradients are taken only with respect to this representation, with the model in evaluation mode so BatchNorm and LayerNorm statistics are fixed. For a transformed view \(t(x)\), we define
\begin{equation}
    g_t(x) = \nabla_{h_t} \Delta(t(x)),
    \qquad h_t=h(t(x))
\end{equation}
Under the local linear approximation \(\Delta(h)\approx\Delta(h_t)+g_t^\top(h-h_t)\), the minimum-norm feature perturbation that sets this scalar margin proxy to zero has norm \(|\Delta(t(x))|/\|g_t(x)\|_2\). We use the squared form
\begin{equation}
    d_t^2(x)
    = \frac{\Delta(t(x))^2}{\|g_t(x)\|_2^2+\varepsilon}
    \label{eq:bp_distance}
\end{equation}
The quantity in Eq.~\eqref{eq:bp_distance} is a local, first-order fragility proxy. It is the squared minimum-norm displacement that would zero the margin proxy in the coordinates of the chosen feature map \(h\). Under a smooth reparameterization \(h\mapsto\phi(h)\) the gradient transforms by the inverse Jacobian, so the numeric value depends on the feature coordinates even when the classifier function is unchanged. Thus, BP is a coordinate-dependent sensitivity indicator, and since each branch is z-scored on ID data before fusion (Eq.~\eqref{eq:zscore}), a global rescaling of \(h\) cancels in \(S_{\mathrm{BP},z}\). The top-\(k_m\) averaged margin is also not identical to the argmax boundary, and the first-order approximation can fail when the margin landscape is nonlinear. It follows the same robustness intuition as perturbation-based post-hoc OOD scoring~\citep{chen2025pro}, while avoiding an iterative inner optimization at test time. We set the stabilizer to \(\varepsilon=10^{-8}\), Appendix~\ref{app:grad_diag} reports the distribution of \(\|g_t(x)\|_2\), shows the guarded denominator never approaches this floor (i.e. the score is insensitive to \(\varepsilon\)), and quantifies sensitivity to the hook location \(h\).

The final Boundary-Probe branch converts this squared proxy into a score where larger values mean greater fragility,
\begin{equation}
    BP^{(t)}(x)
    = -\log\!\left(d_t^2(x)+\varepsilon\right)
    \label{eq:ce_view}
\end{equation}
higher values indicate smaller first-order boundary cost. The same small \(\varepsilon\) stabilizes the denominator in Eq.~\eqref{eq:bp_distance} and the logarithm. We average over the deterministic transform pool,
\begin{equation}
    BP(x)
    = \frac{1}{|\mathbb{T}_{\mathrm{geo}}|}
      \sum_{t\in\mathbb{T}_{\mathrm{geo}}}
      BP^{(t)}(x)
    \label{eq:ce_avg}
\end{equation}
This uses one late-feature backward pass per view and reduces single-view gradient noise. Boundary-Probe therefore gives us whether the classifier is locally fragile after accounting for both the current margin and the margin gradient. It complements confidence scores as it can be large even when the current logits still look plausible.

\subsection{Consistency under deterministic transforms}
\label{subsec:consistency}

A transform-consistency signal measures how much a classifier prediction changes under mild views of the same image. The standard definition starts with a transform pool that is expected to preserve the ID label. We use
    $\mathbb{T}_{\mathrm{geo}}=\{\mathrm{id},\,\mathrm{flip},\,\mathrm{crop}_{0.9}\}
    \label{eq:tgeo}$
where \(\mathrm{crop}_{0.9}\) is a center crop at 90\% scale followed by bilinear resize to \((H,W)\). For temperature \(\tau>0\), the tempered softmax is
\begin{equation}
    p_{\tau}(z)_k
    =\frac{\exp(z_k/\tau)}{\sum_{j=1}^{K}\exp(z_j/\tau)}
    \label{eq:softmax_tau}
\end{equation}
We compute \(p_0(x)=p_\tau(z(x))\) for the base view and \(p_t(x)=p_\tau(z(t(x)))\) for each transformed view.

The standard consistency methods compare either predicted labels or scalar uncertainty values across views. Argmax disagreement checks whether the top class changes. Test-time augmentation (TTA) entropy variance checks whether the entropy changes across augmentations~\citep{shanmugam2021better}. These are useful as they require only additional forward passes and no retraining.

The issue is that both summaries discard information. Argmax disagreement changes only after the top class flips, so it misses smaller but systematic distributional shifts. Entropy variance can remain small when probability mass moves among classes with similar entropy. For OOD detection, these partial summaries can miss instability in the predictive distribution.

We therefore compare the full predictive distributions with a symmetric Jensen-Shannon (JS) score~\citep{lin1991divergence}. Probabilities are clamped at a small floor and renormalized before divergences are computed. For two transformed distributions \(\tilde p_0\) and \(\tilde p_t\), the JS divergence is given by $\mathrm{JS}\!\left(\tilde p_0,\tilde p_t\right)=\tfrac{1}{2}\mathrm{KL}\!\left(\tilde p_0\middle\|\tfrac{\tilde p_0+\tilde p_t}{2}\right)+\tfrac{1}{2}\mathrm{KL}\!\left(\tilde p_t\middle\|\tfrac{\tilde p_0+\tilde p_t}{2}\right)
\label{eq:js_div}$. The final transform-consistency branch compares the two non-identity views to the base prediction,
\begin{equation}
    C(x;\tau)
    = \frac{1}{2}
      \sum_{t\in\{\mathrm{flip},\mathrm{crop}\}}
      \mathrm{JS}\!\left(\tilde p_0(x),\tilde p_t(x)\right)
    \label{eq:consistency_score}
\end{equation}
This requires no additional forward passes beyond the identity, flip, and crop views already used by the Boundary-Probe branch. The Transform-Consistency branch gives us whether the classifier can keep a stable predictive distribution under mild views and captures information that Energy and BP do not directly measure.

\subsection{ID-residual gated fusion}
\label{subsec:egf}

A fusion signal combines several detector scores into one scalar OOD score that is a weighted sum of branch scores after putting them on a comparable scale. The branches have different units: Energy is a logit-scale score, BP is a local fragility score, and TC is a distributional disagreement score.

The standard post-hoc fusion method is to z-score each branch on ID data and then average the standardized scores. We first z-score each branch on \(\mathcal{D}_{\mathrm{ID}}\) and form the corresponding equal-weight fusion as
\begin{equation}
\begin{gathered}
E_z(x) = \frac{E(x;T)-\mu_E}{\sigma_E}, \qquad
S_{\mathrm{BP},z}(x) = \frac{BP(x)-\mu_{\mathrm{BP}}}{\sigma_{\mathrm{BP}}}, \qquad
C_z(x) = \frac{C(x;\tau)-\mu_C}{\sigma_C}\\
S_{\mathrm{equal}}(x) = \frac{1}{3}\left(E_z(x)+S_{\mathrm{BP},z}(x)+C_z(x)\right)
\end{gathered}
\label{eq:zscore}
\end{equation}
The standardized scores express all branches in ID-standard-deviation units, which makes them comparable without OOD labels.

But equal weights apply the same branch balance to every test input. This is restrictive since different inputs can be abnormal in different ways. Some inputs mainly perturb logit evidence, some mainly create boundary fragility, and some mainly create transform instability. A fixed average cannot react to which branch is most ID-atypical for the current sample.

To address this, we replace the fixed equal weights by a per-sample ID-residual gate. The gate uses ID quantiles to determine which branch is unusually large for the current input. Let \(\kappa_E\), \(\kappa_{\mathrm{BP}}\), and \(\kappa_C\) be the empirical \(p_E\), \(p_{\mathrm{BP}}\), and \(p_C\) quantiles of the standardized branches on \(\mathcal{D}_{\mathrm{ID}}\). We use \(p_E=0.90\), \(p_{\mathrm{BP}}=0.70\), and \(p_C=0.70\). Energy is a broad logit prior and receives a more conservative upper-tail threshold, while BP and TC are fragility cues whose signal often begins closer to the ID bulk. These values are fixed ID-only defaults and are not selected using OOD data. For a test input, residuals above the ID thresholds are
\begin{equation}
r_E(x) = \max\bigl(0,\,E_z(x)-\kappa_E\bigr), \qquad
r_{\mathrm{BP}}(x) = \max\bigl(0,\,S_{\mathrm{BP},z}(x)-\kappa_{\mathrm{BP}}\bigr), \qquad
r_C(x) = \max\bigl(0,\,C_z(x)-\kappa_C\bigr)
\end{equation}
The final TriScore fusion converts these residuals into weights and returns the final scalar OOD score (S(x)),
\begin{equation}
\begin{gathered}
    w_j(x)
    = \frac{\exp(r_j(x)/s)}
       {\sum_{i\in\{E,\mathrm{BP},C\}}\exp(r_i(x)/s)},
    \qquad j\in\{E,\mathrm{BP},C\}\\
    S(x)=w_E(x)E_z(x)+w_{\mathrm{BP}}(x)S_{\mathrm{BP},z}(x)+w_C(x)C_z(x)
\end{gathered}
    \label{eq:gated_score}
\end{equation}
We set the softness parameter to \(s=0.6\). As \(s\to 0\), the rule approaches a winner-take-all selector over residuals. As \(s\to\infty\), the rule approaches the equal-weight average. In the reported setting, the gate is moderate. The gated fusion answers which branch is most ID-atypical for the current input, while still retaining information from the other branches.

Calibration uses only the ID split, \(T\) and \(\tau\) are selected by minimizing the variance of the equal-weight standardized score, after which the branch statistics and gate quantiles are fixed. The details of the ID-only calibration and test-time scoring are provided in Appendix~\ref{app:calibration_scoring}.

\section{Evaluation}
\label{sec:evaluation}

The evaluation is designed to test the three-branch score against existing post-hoc detectors and measures its stability across architectures and shifts. We use the OpenOOD framework and compare against 15 post-hoc baselines. The baseline set is restricted to detectors that can be attached to the same frozen classifier, without auxiliary OOD data or retraining, because this is the deployment setting TriScore targets. Within that constraint, the methods cover the main post-hoc families: output and logit scores (MSP, MLS, EBO, ODIN, GEN), activation or classifier-weight shaping (ReAct, ASH, DICE, SCALE), gradient or local-robustness scores (GradNorm, PRO), and feature-geometry scores (KNN, ViM, NCI, CADRef). Most baselines are stock OpenOOD postprocessors, CADRef and PRO are integrated locally as recent strong post-hoc baselines, and TriScore and its branch ablations are added as OpenOOD-compatible postprocessors. The main ImageNet-scale evaluation uses ResNet-50, RegNet, ViT-B/16, and Swin-T. These backbones span a residual CNN, a modern convolutional architecture, a plain vision transformer, and a hierarchical transformer.

\subsection{Benchmarks and protocols}
\label{subsec:protocols}

Following OpenOOD, ImageNet-1K provides the ID data. Near-OOD evaluation uses SSB-Hard and NINCO, and Far-OOD evaluation uses iNaturalist, Textures, and OpenImage-O. TriScore estimates its class-mean logits, branch statistics, temperatures, quantiles, and gate parameters from an ID-only calibration subset of at most 5,000 images. OOD or adversarial examples are not used for calibration. Additional CIFAR-10 OOD checks and open-set recognition checks on CIFAR-6, CIFAR-50, TIN-20, and MNIST-6 are reported in Appendices~\ref{app:cifar} and~\ref{app:osr}.

The adversarial-shift benchmark follows OpenOOD's ImageNet attack-data interface~\citep{yang2022openood,zhang2023openood15}, with an extended local suite used for evaluation. Foolbox and AutoAttack provide the model wrapper and misclassification-criterion interface for the supported attacks. Attacks are generated against the frozen ImageNet classifier with ImageNet labels, and the resulting perturbed ImageNet images are evaluated as shifted inputs against clean ImageNet ID images. The attacker does not optimize TriScore, the branch scores, or the gate.

We use AutoAttack (AA), APGD-CE, C\&W, DeepFool, FGSM, masked PGD (M-PGD), PGD, Square, BIM, and PGD-BPDA. This attack set is chosen to cover single-step and iterative gradient attacks, optimization-style attacks, decision-boundary attacks, black-box query attacks, masked/local perturbations, BPDA-style attacks, and an AutoAttack ensemble. It extends the default ImageNet attack-script suite used in OpenOOD with PGD-BPDA and AutoAttack-family attacks in the evaluated adversarial-OOD protocol. The \(\ell_\infty\) attacks use \(\epsilon=4/255\), while C\&W and DeepFool follow their OpenOOD/Foolbox \(\ell_2\) implementations. Appendix~\ref{app:protocol_reproducibility} gives more details on these attacks.

We report FPR@95 and AUROC. Following the OpenOOD evaluator, OOD is treated as the positive class and the true positive rate is the fraction of OOD samples detected. FPR@95 is then the false positive rate, i.e.\ the fraction of ID samples flagged as OOD, at the threshold where the OOD true positive rate first reaches 95\% (lower is better). AUROC is the area under the ROC curve and equals the probability that a randomly chosen OOD sample receives a higher score than a randomly chosen ID sample (higher is better). Because this FPR@95 threshold is fixed by the OOD score distribution, it is in general distinct from a fixed ID operating point. 

For fixed-threshold deployment, and for the operating point used in the adaptive study of Appendix~\ref{app:adaptive}, we instead flag an input as OOD when \(S(x)\) exceeds the empirical \(95\)th percentile of \(S\) on a held-out ID split, which flags \(5\%\) of ID inputs.

The two ImageNet summary tables below report results for ResNet-50, RegNet, ViT-B/16, and Swin-T, using clean ImageNet-1K test examples as the ID reference. For clean-OOD, each metric is first averaged within the Near-OOD group (NINCO and SSB-Hard) and within the Far-OOD group (iNaturalist, OpenImage-O, and Textures), and the reported mean is the average of those two group means, following the OpenOOD convention. For adversarial OOD, each metric is instead averaged over the ten attacks listed above. The corresponding per-dataset and per-attack results are reported in Appendices~\ref{app:imagenet_baseline} and~\ref{app:adversarial_detailed}.

\subsection{Adversarial OOD detection}
\label{subsec:adv_results}

Table~\ref{tab:adversarial_method_summary} reports the mean over the 10 non-adaptive classifier attacks. TriScore obtains the best mean AUROC on all four backbones and the best FPR@95 on RegNet, ViT-B/16, and Swin-T. On ResNet-50, it also has the best mean AUROC. PRO, MSP, and CADRef have lower FPR@95 in this column. On RegNet, TriScore reaches 72.04 AUROC compared with 71.09 for PRO and 70.29 for EBO. On ViT-B/16, it reaches 63.11 AUROC compared with 61.99 for GEN. On Swin-T, it reaches 65.91 AUROC compared with 65.43 for GEN and 65.24 for DICE.

\begin{table*}[h!]
  \centering
  \scriptsize
  \caption{Non-adaptive adversarial-shift detection averaged over 10 classifier attacks for each backbone.}
  \label{tab:adversarial_method_summary}
  {%
  \begin{tabular}{l*{8}{c}}
    \toprule
    Method & \multicolumn{2}{c}{ResNet-50} & \multicolumn{2}{c}{RegNet} & \multicolumn{2}{c}{ViT-B/16} & \multicolumn{2}{c}{Swin-T} \\
     & FPR@95 & AUROC & FPR@95 & AUROC & FPR@95 & AUROC & FPR@95 & AUROC \\
    \midrule
    ASH        & 90.48 & 55.68 & 93.38 & 52.16 & 95.25 & 51.13 & 94.09 & 52.11  \\
    CADRef      & 87.62 & 58.70 & 87.47 & 58.92 & 89.86 & 56.81 & 82.30 & 62.64  \\
    DICE        & 92.28 & 53.04 & 87.80 & 61.08 & 87.52 & 59.35 & 88.16 & 65.24  \\
    EBO        & 91.67 & 54.60 & 82.76 & 70.29 & 91.98 & 59.57 & 91.61 & 59.40  \\
    GEN        & 90.92 & 55.50 & 81.36 & 70.16 & 84.68 & \underline{61.99} & \underline{80.65} & \underline{65.43}  \\
    GradNorm   & 93.36 & 51.51 & 94.63 & 47.04 & 93.07 & 53.13 & 93.58 & 48.55  \\
    KNN         & 89.15 & 57.96 & 87.33 & 58.88 & 91.37 & 54.96 & 86.38 & 59.24  \\
    MLS        & 91.51 & 54.79 & 82.95 & 69.28 & 91.76 & 59.32 & 91.41 & 59.69  \\
    MSP        & \underline{85.24} & 59.64 & 84.49 & 61.40 & 88.64 & 58.43 & 88.59 & 59.99  \\
    NCI         & 90.20 & 55.98 & 93.34 & 53.45 & 92.28 & 54.71 & 83.82 & 61.90  \\
    ODIN       & 94.22 & 47.83 & 85.98 & 63.92 & 93.14 & 55.94 & 93.57 & 53.95  \\
    PRO         & \textbf{77.83} & \underline{60.39} & \underline{72.79} & \underline{71.09} & \underline{84.23} & 59.44 & 80.98 & 63.21  \\
    ReAct      & 89.73 & 56.15 & 87.23 & 62.55 & 89.24 & 58.39 & 87.24 & 61.84  \\
    SCALE       & 90.24 & 56.36 & 89.28 & 62.34 & 91.91 & 59.10 & 92.09 & 56.80  \\
    ViM         & 90.70 & 56.57 & 87.53 & 59.83 & 93.35 & 55.24 & 85.27 & 63.47  \\
    \midrule
    TriScore   & 88.64 & \textbf{62.11} & \textbf{71.26} & \textbf{72.04} & \textbf{84.10} & \textbf{63.11} & \textbf{80.01} & \textbf{65.91}  \\
    \bottomrule
  \end{tabular}}
\end{table*}

The per-attack tables in Appendix~\ref{app:adversarial_detailed} show that the aggregate result is not caused by a single attack. Across the 40 attack\(\times\)backbone cells, TriScore gives the best AUROC in 27 cells. The FPR@95 view is more competitive because PRO is explicitly optimized for perturbation robustness. TriScore still has the best FPR@95 in 19 of 40 cells, compared with 16 for PRO, and it has the best mean FPR@95 on three of four backbones in Table~\ref{tab:adversarial_method_summary}. The nearest competitors are attack-specific. NCI is strongest on several ResNet-50 optimization attacks, DICE is strong on parts of Swin-T, and GEN is competitive on several transformer cells. The pattern supports the intended role of TriScore as a broad robustness-floor detector, not necessarily a detector specialized to one attack family.

A second useful statistic is cross-attack variance. On Swin-T, TriScore has a per-attack AUROC standard deviation of 4.5, compared with 5.5 for GEN and 6.9 for DICE. On ViT-B/16, TriScore has standard deviation 5.5, compared with 6.6 for GEN and 5.8 for DICE. Some lower-mean baselines such as PRO or MSP have smaller variance, so the stability claim is relative to the high-AUROC competitors. This matters because adversarial OOD deployments do not usually know the attack family in advance. A method with high mean performance and moderate attack-to-attack dispersion can be preferable to a method that wins on one attack and collapses on another.

The same argument applies across backbones. In Table~\ref{tab:adversarial_method_summary}, TriScore's worst-backbone mean AUROC is $62.1$ (on ResNet-50), which is higher than the worst-backbone AUROC of every one of the fifteen baselines, the best baseline floor is PRO at $59.4$, and the rest fall further, several below $50$ (GradNorm $47.0$, ODIN $47.8$). TriScore never drops below $62$ AUROC under adversarial shift, whereas each baseline has at least one backbone on which it does. Therefore it holds its robustness floor across architectures.

Under this adversarial-shift protocol, TriScore gives the strongest mean AUROC across the evaluated backbones and the best mean FPR@95 on three of four, with lower cross-attack dispersion than the closest high-AUROC competitor on transformer backbones. Appendix~\ref{app:complementarity} identifies the mechanism behind this floor. The branch correlations are heterogeneous across backbones: on ResNet-50 several pairs are correlated on clean ID data and decorrelate substantially under attack, while on RegNet and the transformer backbones the branches are already weakly correlated on clean data and remain weak-to-moderate under attack. In both regimes the branch errors overlap only partially, so the fusion degrades where single scores fail. Appendix~\ref{app:altfusion} asks whether the same floor is reachable by fusing other existing detectors, and Appendix~\ref{app:adaptive} reports the adaptive attacks.

\subsection{Standard OOD detection}
\label{subsec:standard_results}

Table~\ref{tab:imagenet_clean_mean} reports clean ImageNet-1K OOD detection for each of the four standard backbones under the group-mean protocol defined above. It includes both the final gated TriScore and the E+BP branch subset so that the comparison distinguishes the main detector from the simpler two-branch configuration analyzed in Appendix~\ref{app:ablation_details}.

\begin{table*}[h!]
  \centering
  \scriptsize
  \caption{Clean-OOD detection results for each backbone.}
  \label{tab:imagenet_clean_mean}
  {%
  \begin{tabular}{l*{8}{c}}
    \toprule
    Method & \multicolumn{2}{c}{ResNet-50} & \multicolumn{2}{c}{RegNet} & \multicolumn{2}{c}{ViT-B/16} & \multicolumn{2}{c}{Swin-T} \\
     & FPR@95 & AUROC & FPR@95 & AUROC & FPR@95 & AUROC & FPR@95 & AUROC \\
    \midrule
    ASH              & 41.45 & 86.96 & 64.65 & 76.43 & 95.61 & 52.38 & 94.88 & 45.55  \\
    CADRef            & \underline{39.22} & \underline{87.96} & \underline{34.48} & 89.96 & \textbf{46.55} & \textbf{85.19} & \textbf{46.77} & \textbf{85.36}  \\
    DICE              & 57.15 & 82.01 & 63.44 & 81.17 & 68.59 & 73.84 & 95.66 & 54.46  \\
    EBO              & 53.49 & 82.68 & 38.10 & \underline{92.09} & 89.27 & 70.70 & 79.45 & 77.27  \\
    GEN              & 50.46 & 83.31 & \textbf{32.02} & \textbf{92.83} & 51.50 & 83.82 & \underline{48.10} & \underline{84.97}  \\
    GradNorm         & 63.38 & 81.61 & 88.28 & 56.86 & 93.68 & 40.51 & 95.22 & 41.53  \\
    KNN               & 52.50 & 80.64 & 34.65 & 90.33 & \underline{51.20} & 82.46 & 52.94 & 80.49  \\
    MLS              & 53.02 & 83.02 & 38.74 & 91.86 & 85.74 & 75.92 & 74.29 & 80.23  \\
    MSP              & 58.56 & 80.63 & 46.66 & 87.13 & 66.76 & 79.78 & 60.11 & 81.53  \\
    NCI               & 42.23 & 87.10 & 62.05 & 82.28 & 57.49 & 81.30 & 52.54 & 83.02  \\
    ODIN             & 58.22 & 82.11 & 46.32 & 89.40 & 88.37 & 70.19 & 85.56 & 70.72  \\
    PRO               & 59.21 & 81.00 & 42.32 & 89.80 & 58.51 & 81.94 & 57.27 & 82.63  \\
    ReAct            & 46.52 & 85.52 & 50.85 & 83.62 & 69.21 & 77.48 & 57.83 & 81.93  \\
    SCALE             & \textbf{38.17} & \textbf{88.94} & 41.37 & 91.08 & 90.24 & 66.97 & 85.61 & 70.66  \\
    ViM               & 48.01 & 82.38 & 35.38 & 90.24 & 51.48 & \underline{84.93} & 49.82 & 84.22  \\
    \midrule
    TriScore (E+BP)   & 55.88 & 80.24 & 38.10 & 90.15 & 74.08 & 77.43 & 67.42 & 79.58  \\
    TriScore         & 52.14 & 80.01 & 37.85 & 89.28 & 65.99 & 80.00 & 57.82 & 81.37  \\
    \bottomrule
  \end{tabular}}
\end{table*}

The strongest clean baselines vary by architecture: SCALE is strongest on ResNet-50 in this table, GEN is strongest on RegNet, and CADRef is strongest on ViT-B/16 and Swin-T. 
The E+BP TriScore variant is a lower-cost clean-OOD alternative, while the final gated detector has better clean FPR@95 on all four backbones in this table. 

The results show that TriScore's main advantage is not clean-OOD dominance; it is the stronger adversarial-shift floor in Table~\ref{tab:adversarial_method_summary}. On the four-backbone clean mean AUROC, seven baselines score above TriScore's $82.67$ (CADRef $87.12$, GEN $86.23$, ViM $85.44$, PRO $83.84$, KNN $83.48$, NCI $83.43$, and MLS $82.76$). TriScore is mid-pack on clean-OOD because it does not use the richer ID-training feature geometry that CADRef and ViM exploit, and that geometry is exactly what helps on standard semantic and texture shifts. The residual gate is designed to act where a branch is driven into its ID tail (Section~\ref{subsec:egf}). On clean-OOD, where the branches stay near their ID-typical range, it reduces toward an equal-weight average and adds little, so the gated detector is competitive rather than dominant here. 

The clean-OOD advantage of the strongest feature-based baselines also comes at a cost and fails to transfer among them. CADRef, ViM, and KNN each lose $26$-$28$ AUROC from clean to adversarial averaged over the four backbones (versus TriScore's $16.9$), and their adversarial AUROC ($55$-$63$) falls below TriScore's ($62$-$72$) on every backbone. All three require access to the ID-training features at setup, and KNN additionally retains and searches the full $1.28$M-vector feature bank at inference (detailed in Appendix~\ref{subsec:cifar10_timing}), whereas TriScore calibrates on ID-validation statistics only and stores no bank. The closest baseline overall is GEN, a cheap logit score that beats TriScore on clean-OOD and is the strongest adversarial baseline on the transformer backbones. TriScore wins over GEN on adversarial AUROC on all four backbones. TriScore therefore trades a few points of clean AUROC for a stronger and lighter adversarial-shift floor. 

The E+BP row is included as a simpler deployment variant. Empirically, E+BP has slightly higher AUROC than the gated TriScore on ResNet-50 and RegNet, while the gated detector has higher AUROC on ViT-B/16 and Swin-T and lower FPR@95 on all four backbones. Therefore, E+BP provides a cheaper alternative in clean-OOD, while the gated three-branch TriScore remains the main detector for adversarial shift.

The same table also shows a stability advantage that is easy to miss from mean rankings alone. Several baselines have a severe architecture-specific failure on at least one backbone. ASH falls to $52.38$ AUROC on ViT-B/16, GradNorm to $56.86$ on RegNet, DICE to $54.46$ on Swin-T, and SCALE to $66.97$ on ViT-B/16. TriScore remains at or above $80$ AUROC on all four standard backbones in the clean summary. FPR@95 shows the same stability pattern but not dominance. TriScore stays below $66$ FPR@95 on all four backbones, while ASH, GradNorm, DICE, and SCALE each exceed $90$ FPR@95 on at least one backbone. This cross-architecture floor is a useful property for post-hoc deployment, where the detector is often attached after the classifier architecture has already been chosen.

TriScore is not the clean-OOD winner, but it remains competitive and avoids the most severe architecture-specific failures. For clean-OOD deployment, the E+BP variant is the lower-cost TriScore choice, while the final gated detector is the main TriScore variant used for adversarial-shift robustness.

Full per-dataset ImageNet baseline results are reported in Appendix~\ref{app:imagenet_baseline}, and gated and equal-weight TriScore variants are separated in Appendices~\ref{app:branch_isolation}--\ref{app:ablation_adversarial}.

\section{Limitations}
\label{sec:limitations}

TriScore is designed for frozen classifiers, ID-only calibration, and the non-adaptive classifier-level adversarial-shift protocol described in Section~\ref{subsec:protocols}. Therefore, it is not a certified detector, and it is not robust to an adaptive detector-aware adversary as shown in Appendix~\ref{app:adaptive}. However, on the same TriScore-targeted adversarial examples, several standard differentiable baselines (EBO, MLS, MSP, GEN, GradNorm) also show high transfer evasion as those attacks were not separately optimized against each baseline. This can be interpreted as transfer evasion, consistent with the adaptive-attacks literature. We explicitly scope the robustness claim to the non-adaptive setting and treat detector-aware robustness as out-of-scope.

The method is strongest when the deployment setting rewards redundancy across logit energy, local boundary fragility, and transform consistency. Clean-OOD benchmarks do not always require all three cues. In the latest branch-combination ablation, the ungated equal-weight E+BP+TC variant has the strongest clean aggregate, while E+BP remains a lower-cost alternative when the extra transform-consistency computation is undesirable. The final gated TriScore should be preferred when robustness to the evaluated adversarial-shift protocol is the primary objective.

The branch behavior is also architecture dependent. On ResNet-50, the energy signal is weak under adversarial shift, and BP-only is better than the full combination. On RegNet, ViT-B/16, and Swin-T, the full combination is more effective. These results suggest that a practical implementation should include a lightweight branch diagnostic before deployment. The current paper uses the ablation evidence to state this rule manually. An automatic ID-only diagnostic for deciding between BP, E+BP, and E+BP+TC remains future work.

Appendix~\ref{app:sensitivity} reports a grid sweep over the quantile levels and softness and a 5-seed calibration study. The ranking is stable, and calibration-subset variability is small (standard deviation $\le 0.07$ AUROC across backbones, one to two orders of magnitude below the Table~\ref{tab:adversarial_method_summary} margins). This bounds calibration-subset noise but not finite-test-set or attack-generation randomness, which we quantify with paired per-image bootstrap intervals in Appendix~\ref{app:bootstrap}; close $1$-$2$ AUROC margins should be read with that scope in mind. The quantile levels and softness are fixed ID-only defaults. A backbone-adaptive softness could improve the gate, but we do not tune it without OOD validation data. We also do not report threshold-dependent metrics beyond FPR@95, so fixed-threshold deployment should calibrate the operating point on a held-out ID split.

Boundary-Probe is high for locally fragile decisions, and hard ID examples can also be locally fragile. Appendix~\ref{app:bp_uncertainty} provides the diagnostic. On clean ID data, BP is strongly related to softmax confidence and predicts ID misclassification, so it behaves largely as a hard-example cue. Its independent value appears under shift, where the component of BP orthogonal to MSP still separates adversarial OOD. BP is thus redundant with uncertainty on clean data and complementary to it under attack, which is the regime the method targets.

TriScore is also more expensive than single-forward detectors. It uses three forward passes and three late-feature backward passes per input. For latency-critical settings, E-only or E+BP variants provide lower-cost alternatives.

\section{Conclusion}
\label{sec:conclusion}

TriScore is a post-hoc OOD detector that attaches to a frozen classifier and combines three ID-calibrated signals, namely class-centered logit energy, Boundary-Probe fragility, and Jensen-Shannon transform consistency. The method uses no OOD labels, auxiliary model, retraining, or stored ID-training feature bank. The feature-geometry detectors it competes with need ID-training access at setup, and some of them must search the full feature bank at inference. TriScore instead stores one class-mean logit vector and a small set of scalar ID-validation statistics, with no feature bank. It has no training-set dependency at deployment and is significantly faster than the heaviest baseline. Its main design principle is redundancy. A single confidence cue can be preserved by a shift or perturbation, while a combination of logit evidence, boundary fragility, and view consistency is harder to satisfy simultaneously.

Across four ImageNet-scale backbones and 15 post-hoc baselines, TriScore obtains the best mean AUROC in the non-adaptive adversarial-shift protocol on all four backbones and the best mean FPR@95 on three of four. Its robustness floor holds across architectures as the worst-backbone exceeds the worst-backbone of every baseline, the detector never collapses on a particular backbone unlike the individual cues. It also gives lower cross-attack AUROC variance on the two transformer backbones, which indicates a more uniform robustness floor than the closest high-AUROC competitor. On standard OOD benchmarks, the method remains competitive and avoids several severe architecture-specific failures, although specialized clean-OOD baselines are often stronger on clean-OOD alone.

The ablation studies clarify where the gains come from. Boundary-Probe improves over margin-only and GradNorm-style alternatives, and JS-TC improves over simpler transform-consistency baselines. Class-centered energy is best viewed as a calibrated logit prior for fusion rather than a standalone replacement for EBO. The branches are partly redundant under clean shift, where the ungated E+BP+TC variant gives the strongest fixed-subset aggregate and E+BP is a lower-cost alternative. They are more complementary under adversarial shift, where the full gated detector is stronger on backbones with a usable energy signal.

The main open directions are defenses against detector-aware attackers (e.g. randomized or non-differentiable gate variants that do not expose a clean gradient path), automatic branch diagnostics that select BP, E+BP, or E+BP+TC for a given backbone from ID-only signals, transform pools that better handle texture-like OOD and smooth optimization-based perturbations, and ImageNet-scale latency for the multi-view scoring. These extensions would turn TriScore from a strong post-hoc adversarial-shift detector into a complete robust OOD detection.

\section{Acknowledgements}
This work is supported by Defi Clé Robotique centrée sur l’Humain funded by Région Occitanie, France and was granted access to the HPC resources of IDRIS under the allocation AD011017069 made by GENCI.

\bibliography{main3}
\bibliographystyle{tmlr}

\appendix
\section{Reproducibility Details}
\label{app:protocol_reproducibility}

\subsection{Evaluation protocol}

This section collects the protocol details that are easy to miss from the tables in the main text.

TriScore uses only an ID calibration subset of at most 5,000 examples from the OpenOOD ID validation loader to estimate class-mean logits, branch means and standard deviations, temperature choices, branch quantiles, and gate parameters. Detector metrics are computed with the OpenOOD ID test loader and the corresponding OOD loaders. OOD or adversarial images are not used for calibration. Main clean-OOD and ablation summaries follow the OpenOOD convention used in our scripts. Near-OOD datasets are averaged as a group, Far-OOD datasets are averaged as a group, and the reported mean averages the two groups. Adversarial summaries average the 10 attack-specific metrics listed in Table~\ref{tab:attack_protocol_details}.

\subsection{Adversarial-shift protocol}

For all attacks that require labels, the label source is the ImageNet ground-truth label of the source validation image. For all $\ell_\infty$ attacks, we use $\epsilon=4/255$. The attack objective is classifier misclassification under the frozen backbone. For Foolbox attacks this is the standard misclassification criterion, for AutoAttack variants it is the standard evaluation routine with the true ImageNet labels. After generation, the detector only sees the perturbed image and does not use labels. Table~\ref{tab:attack_protocol_details} summarizes the per-attack objective, norm and budget, and implementation details.

\begin{table}[H]
  \centering
  \scriptsize
  \caption{Adversarial-shift protocol used in the ImageNet attack tables.}
  \label{tab:attack_protocol_details}
  \resizebox{\textwidth}{!}{%
  \begin{tabular}{llll}
    \toprule
    Attack & Objective & Norm & Implementation details \\
    \midrule
    FGSM & classifier loss & $\ell_\infty$ & OpenOOD/Foolbox default FGSM path \\
    BIM & classifier loss & $\ell_\infty$ & OpenOOD/Foolbox basic iterative attack \\
    PGD & classifier loss & $\ell_\infty$ & OpenOOD/Foolbox PGD defaults \\
    PGD-BPDA & classifier loss & $\ell_\infty$ & Foolbox LinfPGD with 100 steps through the BPDA model wrapper \\
    AA & classifier loss & $\ell_\infty$ & AutoAttack standard evaluation \\
    APGD-CE & classifier cross-entropy & $\ell_\infty$ & AutoAttack APGD-CE component \\
    Square & classifier loss & $\ell_\infty$ & AutoAttack Square component with default query schedule \\
    C\&W & classifier misclassification & $\ell_2$ & Foolbox L2CarliniWagnerAttack, 1000 steps \\
    DeepFool & classifier misclassification & $\ell_2$ & Foolbox L2DeepFoolAttack defaults \\
    M-PGD & classifier loss & masked & OpenOOD masked PGD, $\epsilon=1$, $\alpha=0.01$, 40 steps, patch size 60 \\
    \bottomrule
  \end{tabular}}
\end{table}

\subsection{Implementation details}

All branch scores are standardized using ID-only statistics. To avoid numerical instability from near-degenerate branches, the implementation applies a small standard-deviation floor of $10^{-6}$ before z-scoring. Probability computations in JS-TC use clamped probabilities for numerical stability. The fixed default hyperparameters are \(k_m=5\), \(p_E=0.90\), \(p_{\mathrm{BP}}=p_C=0.70\), softness \(s=0.6\), temperature grid \(T,\tau\in\{0.5,1.0\}\), and transform pool \(\{\mathrm{id},\mathrm{flip},\mathrm{crop}_{0.9}\}\). These values are selected without OOD labels. Sensitivity analyses over these constants are reported in Appendices~\ref{app:grad_diag}, \ref{app:within_ablation}, and~\ref{app:sensitivity}.

Boundary-Probe backpropagates only to a chosen late feature representation and uses the model in evaluation mode. BatchNorm and LayerNorm statistics are not updated during scoring. The hook is placed on the last task-level representation before the classification head: the final residual/trunk representation for ResNet-50 and RegNet, the final class-token representation for ViT-B/16, the final pooled stage representation for Swin-T. 

\subsection{Calibration and scoring}
\label{app:calibration_scoring}

Implementation follows this decomposition. The same identity, flip, and crop views are used for BP and TC, so their forward passes are shared. The backward pass is used only to obtain \(\nabla_h\Delta\) at the chosen late feature representation, and classifier parameters are not updated. The offline temperature search evaluates four \((T,\tau)\) pairs on the ID calibration split, with BP cached as it does not depend on either temperature. Additional hook locations are reported in Appendix~\ref{app:grad_diag}; numerical safeguards and default constants are given in Appendix~\ref{app:protocol_reproducibility}, and the timing comparison is reported in Appendix~\ref{subsec:cifar10_timing}.

Algorithm~\ref{alg:overview} gives the full calibration and scoring procedure. The offline phase estimates temperatures, branch statistics, and ID quantiles. The test-time phase evaluates the three branches and applies the same stored statistics to every input. \(F\) is the cost of one forward pass and \(B\) the cost of one backward pass through the chosen late feature representation. With \(|\mathbb{T}_{\mathrm{geo}}|=3\), test-time scoring performs three forward passes and three late-feature backward passes per image, i.e. \(\mathcal{O}(|\mathbb{T}_{\mathrm{geo}}|\,(F+B))\). Offline calibration caches BP once per ID sample, while E and TC are recomputed over the temperature grid.

\begin{algorithm}[H]
\scriptsize
\caption{TriScore ID-only calibration and test-time scoring}
\label{alg:overview}
\begin{algorithmic}[1]
\Require Frozen classifier \(f\), ID validation set \(\mathcal{D}_{\mathrm{ID}}\),
       transform pool \(\mathbb{T}_{\mathrm{geo}}\),
       temperature grids \(\mathcal{T},\mathcal{T}_\tau\),
       quantile levels \(p_E,p_{\mathrm{BP}},p_C\in(0,1)\),
       softness \(s>0\)
\Ensure Scalar OOD score \(S(x)\in\mathbb{R}\) for each test input \(x\)

\vspace{2pt}
\State \textbf{Offline calibration} \hfill cache BP once and grid-search E and TC temperatures
\For{each \((T,\tau)\in\mathcal{T}\times\mathcal{T}_\tau\)}
  \For{each \(x\) in \(\mathcal{D}_{\mathrm{ID}}\)}
    \State compute \(E(x;T),\,BP(x),\,TC(x;\tau)\) \Comment{Eqs.~\eqref{eq:energy_def}, \eqref{eq:ce_view}--\eqref{eq:ce_avg}, and \eqref{eq:consistency_score}}
  \EndFor
  \State z-score each branch using statistics from this \((T,\tau)\)
  \State form the equal-weight standardized sum on \(\mathcal{D}_{\mathrm{ID}}\) and record its variance
\EndFor
\State \((T^*,\tau^*)\leftarrow\arg\min_{(T,\tau)}\mathrm{Var}\)
\State recompute scores with \((T^*,\tau^*)\) and estimate \((\mu_\cdot,\sigma_\cdot)\)
\State set quantile thresholds \(\kappa_E,\kappa_{\mathrm{BP}},\kappa_C\)
\State store \((T^*,\tau^*,\mu_\cdot,\sigma_\cdot,\kappa_\cdot)\)

\vspace{2pt}
\State \textbf{Test-time scoring}
\State given test image \(x\) and stored statistics
\State compute raw scores \(E(x;T^*),\,BP(x),\,TC(x;\tau^*)\)
\State standardize to get \(E_z(x),\,S_{\mathrm{BP},z}(x),\,C_z(x)\)
\State compute residuals from standardized scores
\Statex \hspace{1.5em} \(r_E=\max(0,E_z-\kappa_E)\), \(r_{\mathrm{BP}}=\max(0,S_{\mathrm{BP},z}-\kappa_{\mathrm{BP}})\), \(r_C=\max(0,C_z-\kappa_C)\)
\State compute weights \(w_j(x)=e^{r_j/s}/\sum_i e^{r_i/s}\)
\State \Return \(S(x)=w_E\,E_z+w_{\mathrm{BP}}\,S_{\mathrm{BP},z}+w_C\,C_z\)
\end{algorithmic}
\end{algorithm}

Temperatures \(T\) and \(\tau\) are selected by grid search over \(\mathcal{T}\times\mathcal{T}_\tau\). We use \(\mathcal{T}=\{0.5,1.0\}\) and \(\mathcal{T}_\tau=\{0.5,1.0\}\), and choose the pair that minimizes the variance of the equal-weight standardized score on up to 5,000 ID calibration samples. Because each branch is re-standardized to unit variance before this variance is measured, for three standardized branches we have \(\operatorname{Var}\!\big(\tfrac{1}{3}(Z_E+Z_{\mathrm{BP}}+Z_{TC})\big)=\tfrac{1}{3}+\tfrac{2}{9}\big(\rho_{E,\mathrm{BP}}+\rho_{E,TC}+\rho_{\mathrm{BP},TC}\big)\), so minimizing it is equivalent to minimizing the sum of pairwise ID correlations. Consequently, the criterion selects temperatures with the least correlated standardized ID fluctuations, acting as an ID-only complementarity selector that favors independent variance over branch agreement, which aligns with the correlation analysis in Appendix~\ref{app:complementarity}. It uses no OOD labels and does not tune for benchmark performance.

Baselines are run through the OpenOOD post-hoc pipeline with their standard ID-only setup. Methods that require ID-train statistics or feature banks (for example KNN, ViM, DICE, NCI) are given their standard OpenOOD setup step. TriScore uses an ID-validation setup step. The implementation is organized as an OpenOOD compatible postprocessor.

\section{Ablation and Validation}
\label{app:ablation_details}
\label{app:branch_isolation}

% The ablation is organized around three specific questions:

% \begin{itemize}
%     \item Does each branch improve over the closest predecessor or simpler variant under the setup?
%     \item Which branch subset should be used under clean-OOD and adversarial OOD?
%     \item Does the ID-residual gate improve over equal-weight fusion?
% \end{itemize}

\subsection{Contribution of individual branches}

Table~\ref{tab:branch_isolation_main} isolates the branch definitions. In the energy family, class-centered energy is comparable to standard EBO but is not a uniformly stronger standalone detector. This is why we use it as a calibrated energy component inside TriScore instead of presenting it as a replacement for EBO. In the boundary family, Boundary-Probe improves over both margin-only and GradNorm. In the consistency family, JS-TC improves over argmax disagreement and TTA entropy variance, with the largest benefit in FPR@95.

\begin{table}[H]
  \centering
  \scriptsize
  \caption{Branch-isolation results averaged.}
  \label{tab:branch_isolation_main}
  {%
  \begin{tabular}{lllrrrr}
    \toprule
    Family & Score & Reference & \multicolumn{2}{c}{Clean} & \multicolumn{2}{c}{Adversarial} \\
    \cmidrule(lr){4-5} \cmidrule(lr){6-7}
    & & & FPR@95 & AUROC & FPR@95 & AUROC \\
    \midrule
    Energy & EBO & standard energy & 63.23 & 81.82 & 89.50 & 60.97  \\
    Energy & E (Ours) & ID-centered energy & 63.36 & 81.27 & 89.59 & 60.77  \\
    \midrule
    Boundary & Margin-only & top-$k$ margin only & 50.94 & 83.96 & 87.03 & 58.86  \\
    Boundary & GradNorm & gradient baseline & 84.19 & 55.70 & 93.66 & 50.06  \\
    Boundary & \textbf{BP (Ours)} & margin + late-feature gradient ratio & \textbf{48.66} & \textbf{84.50} & 86.16 & 59.47  \\
    \midrule
    Consistency & Argmax & TTA label instability & 100.00 & 63.28 & 100.00 & 52.71  \\
    Consistency & TTA entropy & TTA probability instability & 72.33 & 72.38 & 88.81 & 58.64  \\
    Consistency & \textbf{TC (Ours)} & Jensen--Shannon TTA consistency & 54.35 & 78.66 & \textbf{84.03} & \textbf{61.16}  \\
    \bottomrule
\end{tabular}}
\end{table}

As reported in Table~\ref{tab:branch_isolation_main}, the proposed BP and TC branches improve over their closest simpler alternatives, while class-centered energy should be treated as a calibrated component for fusion. The detailed clean results are reported by backbone in Tables~\ref{tab:branch_isolation_clean_resnet50_app}, \ref{tab:branch_isolation_clean_regnet_app}, \ref{tab:branch_isolation_clean_vitb16_app}, and~\ref{tab:branch_isolation_clean_swint_app}; the per-attack adversarial results are reported in Tables~\ref{tab:branch_isolation_adv_resnet50_app}, \ref{tab:branch_isolation_adv_regnet_app}, \ref{tab:branch_isolation_adv_vitb16_app}, and~\ref{tab:branch_isolation_adv_swint_app}.

% On the per-backbone averages summarized in Table~\ref{tab:branch_isolation_main}, Boundary-Probe improves over margin-only in clean and adversarial AUROC and FPR@95, and it improves over GradNorm in all adversarial metrics and all clean FPR@95 values; the only exception is ResNet-50 clean AUROC, where GradNorm is slightly higher. JS-TC improves over entropy-variance TTA on the same averaged clean and adversarial metrics. The detailed rows in Tables~\ref{tab:branch_isolation_clean_resnet50_app}, \ref{tab:branch_isolation_clean_regnet_app}, \ref{tab:branch_isolation_clean_vitb16_app}, \ref{tab:branch_isolation_clean_swint_app}, \ref{tab:branch_isolation_adv_resnet50_app}, \ref{tab:branch_isolation_adv_regnet_app}, \ref{tab:branch_isolation_adv_vitb16_app}, and~\ref{tab:branch_isolation_adv_swint_app} show the per-dataset and per-attack exceptions behind those averages. Class-centered energy is included as the calibrated energy input to TriScore, not as a standalone superiority over EBO.

\subsection{Contribution of fusion and gate}

Table~\ref{tab:ablation_summary} isolates branch combinations. On clean-OOD, the full equal-weight fusion E+BP+TC and the gated TriScore detector are tied in aggregate FPR@95 after rounding, both reaching $53.5$. The equal-weight fusion is slightly better in AUROC, $82.9$ versus $82.7$. Gating preserves the best clean-OOD FPR@95 of the fixed full fusion while remaining essentially comparable in AUROC.

On adversarial OOD, this pattern changes. TriScore gives the best aggregate FPR@95 and AUROC, improving over E+BP by $2.1$ FPR@95 points and $1.8$ AUROC points, and over E+BP+TC by $6.3$ FPR@95 points and $3.9$ AUROC points. The per-backbone results show the main exception: BP-only is stronger than TriScore on ResNet-50, where the energy branch is weak. On RegNet, ViT-B/16, and Swin-T, the gated fusion is stronger. We treat E+BP as a lower-cost clean-OOD option, E+BP+TC as the strongest fixed clean-OOD subset, and TriScore as the preferred adversarial-shift variant.

\begin{table}[H]
  \centering
  \scriptsize
  \caption{Branch-combination ablation averaged.}
  \label{tab:ablation_summary}
{%
  \begin{tabular}{lrrrr}
    \toprule
    Variant & \multicolumn{2}{c}{Clean-OOD} & \multicolumn{2}{c}{Adversarial OOD} \\
    \cmidrule(lr){2-3} \cmidrule(lr){4-5}
    & FPR@95 & AUROC & FPR@95 & AUROC \\
    \midrule
    E          & 65.0 & 80.8 & 89.5 & 61.0  \\
    E+TC       & 63.0 & 81.5 & 88.9 & 61.3  \\
    E+BP       & 58.9 & 81.9 & 83.1 & 64.0  \\
    BP         & 65.8 & 76.8 & 82.5 & 63.2  \\
    BP+TC      & 63.7 & 78.3 & 83.6 & 61.1  \\
    TC         & 55.9 & 79.0 & 90.0 & 55.4  \\
    E+BP+TC    & \textbf{53.5} & \textbf{82.9} & 87.3 & 61.9  \\
    TriScore & \textbf{53.5} & 82.7 & \textbf{81.0} & \textbf{65.8}  \\
    \bottomrule
  \end{tabular}}
\end{table}

As reported in Table~\ref{tab:ablation_summary}, adding boundary and consistency cues to the energy branch largely improves adversarial robustness, while the ID-residual gate provides an ID-only way to adapt branch weights per input. In the clean aggregate, the ungated E+BP+TC fusion is the strongest fixed subset and is marginally better in AUROC. Under adversarial shift, the gated TriScore is Significantly stronger.

The detailed clean-OOD branch-combination results are reported by backbone in Tables~\ref{tab:triscore_ablation_resnet50_fsood}, \ref{tab:triscore_ablation_regnet_fsood}, \ref{tab:triscore_ablation_vitb16_fsood}, and~\ref{tab:triscore_ablation_swint_fsood}. The per-attack adversarial results are reported in Tables~\ref{tab:triscore_ablation_resnet50_adversarial}, \ref{tab:triscore_ablation_regnet_adversarial}, \ref{tab:triscore_ablation_vitb16_adversarial}, and~\ref{tab:triscore_ablation_swint_adversarial}.

\subsection{Sensitivity analysis}
\label{app:within_ablation}

Tables~\ref{tab:within_km}-\ref{tab:within_pool} report a post-hoc sensitivity analysis of the remaining design constants (top-$k_m$ margin, crop scale, and transform-pool composition) as the defaults were fixed a priori from ID-only reasoning and were not reselected from these results. We report adversarial AUROC, the corresponding clean-OOD sensitivity is comparably small (at most 3.1 AUROC spread across any single parameter). 

The analysis shows two properties: First, sensitivity to these parameters is low-to-moderate, as the adversarial AUROC shifts by at most $\sim\!4$ points across any given sweep. Second, the ID-chosen defaults are explicitly not the adversarial maxima (e.g.\ $k_m=10$ outperforms the default on RegNet, ViT, and Swin, while a flip-only pool outperforms the default on Swin).

\begin{table}[H]
  \centering
  \scriptsize
  \caption{Top-$k_m$ margin ablation, reporting adversarial AUROC scores.}
  \label{tab:within_km}
  \begin{tabular}{lcccc}
    \toprule
    $k_m$ & ResNet-50 & RegNet & ViT-B/16 & Swin-T \\
    \midrule
    2            & \textbf{62.58} & 69.23 & 60.96 & 63.21 \\
    3            & 62.24 & 70.86 & 62.06 & 64.78 \\
    5 (default)  & 62.14 & 72.05 & 63.10 & 65.90 \\
    10           & 62.02 & \textbf{73.09} & \textbf{63.97} & \textbf{66.58} \\
    \bottomrule
  \end{tabular}
\end{table}

\begin{table}[H]
  \centering
  \scriptsize
  \caption{Crop-scale ablation, reporting adversarial AUROC scores.}
  \label{tab:within_crop}
  \begin{tabular}{lcccc}
    \toprule
    Crop scale & ResNet-50 & RegNet & ViT-B/16 & Swin-T \\
    \midrule
    0.8            & \textbf{63.14} & 71.95 & \textbf{63.12} & \textbf{65.93} \\
    0.9 (default)  & 62.14 & 72.05 & 63.10 & 65.90 \\
    0.95           & 61.61 & \textbf{72.08} & 63.10 & 65.87 \\
    \bottomrule
  \end{tabular}
\end{table}

\begin{table}[H]
  \centering
  \scriptsize
  \caption{Transform-pool ablation, reporting adversarial AUROC scores.}
  \label{tab:within_pool}
  \begin{tabular}{lcccc}
    \toprule
    Transform pool & ResNet-50 & RegNet & ViT-B/16 & Swin-T \\
    \midrule
    flip                    & 61.32 & \textbf{72.85} & 63.02 & \textbf{68.07} \\
    flip+crop (default)     & 62.14 & 72.05 & \textbf{63.10} & 65.90 \\
    flip+crop+bright        & \textbf{62.39} & 69.98 & 62.22 & 64.01 \\
    \bottomrule
  \end{tabular}
\end{table}

\subsection{Gate calibration}
\label{app:sensitivity}

Table~\ref{tab:sensitivity} evaluates the stability of the ID-residual gate by sweeping the quantile levels $(p_E, p_{\mathrm{BP}}, p_C)$ and softness $s$ across a 48-point grid, alongside a 5-seed resampling of the ID calibration subset for every backbone. 

The adversarial AUROC proves highly stable, varying by a maximum of just 1.16 points across the entire parameter grid. Our default parameters fall within 0.2-0.7 AUROC of the grid optimum on every backbone. The calibration-seed variance is negligible (std dev $\le 0.07$ AUROC), sitting one to two orders of magnitude below the performance margins reported in Table~\ref{tab:adversarial_method_summary}. As $s \to \infty$, the gate strictly reduces to an equal-weight average (Section~\ref{subsec:egf}). The gated fusion demonstrates its primary advantage when a single branch is driven deep into its ID tail while the others remain typical, which characterizes the regime under adversarial shift.

\begin{table}[H]
  \centering
  \scriptsize
  \caption{Gate grid and calibration sensitivity of the adversarial AUROC. Default is the published configuration $(0.90,0.70,0.70,s{=}0.6)$.}
  \label{tab:sensitivity}
  \begin{tabular}{lccccc}
    \toprule
    Backbone & Default & Grid min & Grid max & Range & Seed (mean$\pm$std) \\
    \midrule
    ResNet-50 & 62.10 & 61.61 & 62.33 & 0.72 & $62.12\pm0.03$ \\
    RegNet    & 72.01 & 71.74 & 72.18 & 0.44 & $72.05\pm0.04$ \\
    ViT-B/16  & 63.07 & 62.87 & 63.16 & 0.29 & $63.08\pm0.05$ \\
    Swin-T    & 65.84 & 65.41 & 66.57 & 1.16 & $65.83\pm0.07$ \\
    \bottomrule
  \end{tabular}
\end{table}

\subsection{Runtime comparison}
\label{subsec:cifar10_timing}

Table~\ref{tab:cifar10_timing_methods_sorted} reports wall-clock timings on a CIFAR-10 benchmark with a single NVIDIA H100 GPU. CIFAR-10 is used here as a manageable runtime benchmark. TriScore completes in $65.4$ seconds including $3.3$ seconds of ID-calibration setup. It is slower than single-forward scores such as MLS and EBO, and also slower than CADRef. It is faster than GEN and PRO, and dramatically faster than KNN. KNN's nearest-neighbor search over the stored ID-training feature bank takes $5.8\times$ longer than TriScore, and this gap only widens at ImageNet scale, where the bank grows from CIFAR-10's $50$k vectors to the $1.28$M ImageNet-1K training features ($3.9$-$15.5$ GB across the four backbones, scaling with the feature dimension). Even though other feature-based detectors (like ViM and CADRef) avoid searching the full feature bank at inference by storing summarized class-aware statistics or residual subspaces, they still require the ID-training data during setup. TriScore avoids this entirely: it is lighter, relies only on ID-validation statistics, stores one class-mean logit vector and a small set of scalar ID-validation statistics, and requires zero training data at deployment. PRO has negligible postprocessor setup in the table, but its total includes the APS evaluator-initialization sweep and input-gradient scoring. GEN is nominally a one-forward-pass score, but its generalized-entropy implementation has nontrivial per-batch post-processing overhead in this code path, so the table should be read as an implementation timing rather than a pure pass-count benchmark. Appendix~\ref{app:protocol_reproducibility} states which baseline setup costs are included.

\begin{table*}[h!]
\centering
\scriptsize
\caption{Runtime ablation on CIFAR-10 by method.}
\label{tab:cifar10_timing_methods_sorted}
\begin{tabular}{lrrl}
\toprule
Method & Total (s) & Setup (s) & Setup note \\
\midrule
ASH & 40.5 & 0.0 & - \\
CADRef & 32.5 & 7.0 & ID-train sweep. \\
DICE & 23.5 & 4.3 & ID-train sweep. \\
EBO & 26.9 & 0.0 & - \\
GEN & 84.8 & 0.0 & - \\
GradNorm & 46.7 & 0.0 & - \\
KNN & 380.4 & 4.5 & ID-train sweep. \\
MLS & \textbf{19.2} & 0.0 & - \\
MSP & 25.6 & 0.0 & - \\
NCI & 41.9 & 7.1 & ID-train sweep. \\
ODIN & 39.1 & 0.0 & - \\
PRO & 69.4 & 0.0 & - \\
ReAct & 37.8 & 0.9 & ID-val sweep. \\
SCALE & 40.7 & 0.0 & - \\
ViM & 46.1 & 6.6 & ID-train sweep. \\
\midrule
{TriScore} & 65.4 & 3.3 & ID-val sweep. \\
\bottomrule
\end{tabular}
\end{table*}

TriScore therefore is a mid-cost post-hoc detector. It is more expensive than single-forward scores, but $5.8\times$ cheaper than KNN, slightly faster than the measured PRO path, and, unlike the feature-based detectors (KNN, ViM, CADRef), needs no ID-training-set access at deployment, and unlike KNN specifically, it stores and searches no feature bank.

\subsection{Alternative detector fusions}
\label{app:altfusion}

To separate the benefit of our triple from the general benefit of combining detectors, we fuse family-diverse trios of existing baselines under the identical ID-only pipeline. Each trio is evaluated with three combiners (equal-weight z-score averaging, our ID-residual gate, and rank aggregation), and Table~\ref{tab:altfusion} reports the best combiner per trio together with its four-backbone mean and worst-backbone AUROC. We also fuse an OOD-tuned logistic model as an out-of-setting reference which uses OOD labels, and to avoid training and testing on the same inputs it is fit on one disjoint half of each OOD split plus the ID calibration set and evaluated on the held-out half.

\begin{table}[H]
  \centering
  \scriptsize
  \caption{Fusions of existing detectors under the identical ID-only pipeline (10-attack mean). \emph{Mean}/\emph{Worst} are over the four backbones. \emph{Access}: ID-val $=$ ID-validation statistics only (bank-free, matched to TriScore), ID-train $=$ ID-training statistics at setup, bank $=$ additionally stores and searches the full ID-training feature bank at inference, OOD $=$ uses OOD labels (out of setting). $\dagger$: OOD-tuned logistic.}
  \label{tab:altfusion}
  \begin{tabular}{lccccccc}
    \toprule
    Fusion & ResNet-50 & RegNet & ViT-B/16 & Swin-T & Mean & Worst & Access \\
    \midrule
    EBO+GradNorm+KNN             & 58.0 & 65.2 & 59.9 & 60.2 & 60.8 & 58.0 & bank \\
    GEN+ViM+GradNorm            & 57.5 & 65.6 & 61.5 & 65.9 & 62.6 & 57.5 & ID-train \\
    GEN+GradNorm+MSP            & 59.1 & 66.5 & 63.4 & 63.9 & 63.2 & 59.1 & ID-val \\
    GEN+KNN+MSP                 & 61.1 & 71.7 & \textbf{64.5} & \textbf{69.7} & \textbf{66.8} & 61.1 & bank \\
    OOD-tuned logistic$^\dagger$ & 59.2 & 71.6 & 60.2 & 62.6 & 63.4 & 59.2 & OOD \\
    TriScore (ours)          & \textbf{62.1} & \textbf{72.1} & 63.1 & 65.9 & 65.8 & \textbf{62.1} & ID-val \\
    \bottomrule
  \end{tabular}
\end{table}

An important part of the gain is the general benefit of diverse fusion: the rank-aggregated GEN+KNN+MSP trio beats TriScore on ViT-B/16 and Swin-T. Three qualifications keep the specific design meaningful. (i)~It is not any fusion as a bank-free, access-matched logit+gradient trio (GEN+GradNorm+MSP), fused under the same pipeline, reaches only $63.2$ mean / $59.1$ worst adversarial AUROC, below TriScore, and for the competitive trio, rank aggregation is the only combiner that lifts it above TriScore (its equal-weight and gated variants are lower). (ii)~TriScore keeps the highest worst-backbone floor ($62.1$ vs the trio's $61.1$). (iii)~The one trio that beats TriScore's mean is carried by KNN, a $5.8\times$ inference slowdown ($380.4$~s vs $65.4$~s, Table~\ref{tab:cifar10_timing_methods_sorted}) and a $1.28$M-vector feature bank ($3.9$-$15.5$~GB) that must be stored and searched at inference. Hence, the comparison identifies a performance-resource trade-off. GEN+KNN+MSP buys the higher mean with a feature bank and training-set access, while TriScore keeps the higher floor bank-free. TriScore is a feature-bank- and training-data-free combination of logit, gradient, and transform-consistency cues.

\section{Detailed ImageNet Results}

\setlength{\tabcolsep}{2pt} % slightly tighter

\subsection{Clean-OOD detection}
\label{app:imagenet_baseline}

% Tables~\ref{tab:resnet50_baseline}, \ref{tab:regnet_baseline}, \ref{tab:vitb16_baseline}, and~\ref{tab:swint_baseline} report per-dataset baseline OOD results for each standard ImageNet backbone.

  % ---------- ResNet-50 Baseline ----------
  \begin{table}[H]
    \centering
    \scriptsize
    \caption{Clean-OOD detection results on ImageNet with ResNet-50.}
    \label{tab:resnet50_baseline}
    {%
    \begin{tabular}{l*{10}{c}}
      \toprule
      Method &
      \hdr{Near-OOD}{NINCO}{FPR@95} &
      \hdr{Near-OOD}{NINCO}{AUROC} &
      \hdr{Near-OOD}{SSB-Hard}{FPR@95} &
      \hdr{Near-OOD}{SSB-Hard}{AUROC} &
      \hdr{Far-OOD}{iNaturalist}{FPR@95} &
      \hdr{Far-OOD}{iNaturalist}{AUROC} &
      \hdr{Far-OOD}{OpenImage-O}{FPR@95} &
      \hdr{Far-OOD}{OpenImage-O}{AUROC} &
      \hdr{Far-OOD}{Textures}{FPR@95} &
      \hdr{Far-OOD}{Textures}{AUROC} \\
      \midrule
      ASH         & 53.07 & 83.45 & 73.67 & 72.89 & 14.10 & 97.06 & 29.19 & 93.26 & 15.32 & 96.90  \\
      EBO & 60.61 & 79.70 & 76.54 & 72.08 & 31.34 & 90.63 & 38.08 & 89.06 & 45.77 & 88.70  \\
      GEN         & 54.88 & 81.70 & 75.71 & 72.01 & 26.11 & 92.44 & 34.52 & 89.26 & 46.24 & 87.59  \\
      GradNorm & 79.50 & 74.02 & 78.23 & 71.90 & 32.01 & 93.89 & 68.46 & 84.83 & 43.24 & 92.05  \\
      MLS & 59.50 & 80.41 & 76.19 & 72.51 & 30.59 & 91.16 & 37.87 & 89.17 & 46.11 & 88.39  \\
      MSP & 56.84 & 79.95 & 74.48 & 72.09 & 43.34 & 88.41 & 50.14 & 84.86 & 60.90 & 82.43  \\
      ODIN & 68.10 & 77.77 & 76.87 & 71.74 & 36.13 & 91.16 & 46.50 & 88.23 & 49.24 & 89.00  \\
      ReAct       & 55.91 & 81.73 & 77.57 & 73.02 & 16.73 & 96.34 & 32.57 & 91.87 & 29.63 & 92.79  \\
      CADRef       & \textbf{49.16} & 85.31 & 73.67 & 74.02 & 14.33 & 96.90 & \textbf{26.52} & 93.93 & \textbf{10.21} & 97.93  \\
      DICE         & 66.96 & 76.01 & 77.97 & 70.13 & 33.40 & 92.54 & 47.86 & 88.26 & 44.23 & 92.04  \\
      KNN          & 58.39 & 79.63 & 83.35 & 62.56 & 40.78 & 86.42 & 44.26 & 87.04 & 17.33 & 97.09  \\
      NCI          & 53.90 & 83.46 & 73.31 & 73.90 & 14.35 & 96.95 & 30.96 & 92.99 & 17.24 & 96.64  \\
      PRO          & 57.93 & 80.70 & 74.61 & 71.83 & 39.14 & 89.47 & 47.84 & 85.77 & 69.47 & 81.99  \\
      SCALE        & 51.92 & \textbf{85.37} & \textbf{67.71} & \textbf{77.34} & \textbf{9.51} & \textbf{98.02} & 28.16 & \textbf{93.95} & 11.91 & 97.63  \\
      ViM          & 62.28 & 78.63 & 80.40 & 65.53 & 30.69 & 89.56 & 32.83 & 90.50 & 10.50 & \textbf{97.97}  \\
      TriScore & 53.70 & 80.17 & 76.16 & 69.91 & 31.93 & 87.47 & 38.64 & 84.22 & 47.48 & 83.28  \\
      \bottomrule
    \end{tabular}}
  \end{table}

  % ---------- RegNet Baseline ----------
  \begin{table}[H]
    \centering
    \scriptsize
    \caption{Clean-OOD detection results on ImageNet with RegNet.}
    \label{tab:regnet_baseline}
    {%
    \begin{tabular}{l*{10}{c}}
      \toprule
      Method &
      \hdr{Near-OOD}{NINCO}{FPR@95} &
      \hdr{Near-OOD}{NINCO}{AUROC} &
      \hdr{Near-OOD}{SSB-Hard}{FPR@95} &
      \hdr{Near-OOD}{SSB-Hard}{AUROC} &
      \hdr{Far-OOD}{iNaturalist}{FPR@95} &
      \hdr{Far-OOD}{iNaturalist}{AUROC} &
      \hdr{Far-OOD}{OpenImage-O}{FPR@95} &
      \hdr{Far-OOD}{OpenImage-O}{AUROC} &
      \hdr{Far-OOD}{Textures}{FPR@95} &
      \hdr{Far-OOD}{Textures}{AUROC} \\
      \midrule
      ASH         & 75.34 & 71.04 & 81.71 & 64.00 & 45.74 & 86.18 & 67.94 & 81.20 & 38.67 & 88.65  \\
      EBO         & 42.51 & 91.67 & 62.06 & 85.28 & 7.71 & 98.29 & 25.94 & 95.83 & 38.10 & 93.02  \\
      GEN         & \textbf{34.46} & \textbf{92.68} & \textbf{58.93} & \textbf{85.71} & 5.96 & 98.53 & 15.71 & 96.89 & 30.38 & 93.97  \\
      GradNorm    & 93.22 & 49.01 & 91.59 & 49.41 & 88.28 & 57.00 & 92.57 & 60.44 & 71.59 & 76.11  \\
      MLS         & 42.78 & 91.56 & 62.49 & 84.83 & 9.10 & 98.05 & 25.72 & 95.70 & 39.69 & 92.82  \\
      MSP         & 48.49 & 86.85 & 65.38 & 78.28 & 28.12 & 94.67 & 36.30 & 91.96 & 44.74 & 88.48  \\
      ODIN        & 52.95 & 87.76 & 60.39 & 83.55 & 22.96 & 95.75 & 44.66 & 91.66 & 40.27 & 92.00  \\
      ReAct       & 59.82 & 80.91 & 73.00 & 73.17 & 21.24 & 94.14 & 43.45 & 89.20 & 41.20 & 87.25  \\
      CADRef       & 38.92 & 88.66 & 66.58 & 78.57 & 11.35 & 96.96 & 23.81 & 94.78 & \textbf{13.48} & \textbf{97.18}  \\
      DICE         & 73.30 & 77.53 & 67.91 & 76.89 & 39.49 & 89.39 & 73.86 & 81.13 & 55.46 & 84.89  \\
      KNN          & 37.36 & 90.99 & 71.31 & 75.73 & \textbf{1.83} & \textbf{99.42} & \textbf{12.99} & \textbf{97.55} & 30.10 & 94.93  \\
      NCI          & 75.58 & 77.25 & 79.72 & 69.84 & 48.39 & 90.38 & 67.39 & 86.79 & 23.55 & 95.88  \\
      PRO          & 42.49 & 90.22 & 68.45 & 80.27 & 12.61 & 97.41 & 25.75 & 95.27 & 49.14 & 90.38  \\
      SCALE        & 49.86 & 89.93 & 66.95 & 82.36 & 11.17 & 97.88 & 33.61 & 94.87 & 28.24 & 95.31  \\
      ViM          & 38.40 & 90.60 & 73.43 & 77.25 & 2.19 & 99.40 & 13.45 & 97.04 & 28.91 & 93.22  \\
      TriScore & 40.28 & 89.42 & 64.36 & 80.87 & 13.20 & 95.15 & 23.62 & 93.50 & 33.31 & 91.58  \\
      \bottomrule
    \end{tabular}}
  \end{table}

  % ---------- ViT-B/16 Baseline ----------
  \begin{table}[H]
    \centering
    \scriptsize
    \caption{Clean-OOD detection results on ImageNet with ViT-B/16.}
    \label{tab:vitb16_baseline}
    {%
    \begin{tabular}{l*{10}{c}}
      \toprule
      Method &
      \hdr{Near-OOD}{NINCO}{FPR@95} &
      \hdr{Near-OOD}{NINCO}{AUROC} &
      \hdr{Near-OOD}{SSB-Hard}{FPR@95} &
      \hdr{Near-OOD}{SSB-Hard}{AUROC} &
      \hdr{Far-OOD}{iNaturalist}{FPR@95} &
      \hdr{Far-OOD}{iNaturalist}{AUROC} &
      \hdr{Far-OOD}{OpenImage-O}{FPR@95} &
      \hdr{Far-OOD}{OpenImage-O}{AUROC} &
      \hdr{Far-OOD}{Textures}{FPR@95} &
      \hdr{Far-OOD}{Textures}{AUROC} \\
      \midrule
      ASH         & 95.40 & 52.52 & 93.50 & 53.89 & 97.02 & 50.63 & 94.80 & 55.52 & 98.49 & 48.53  \\
      EBO         & 94.14 & 66.02 & 92.24 & 58.80 & 83.56 & 79.30 & 88.82 & 76.48 & 83.66 & 81.17  \\
      GEN         & 59.31 & 82.51 & \textbf{82.24} & 70.09 & 22.94 & 93.54 & 35.43 & 90.27 & 38.31 & 90.23  \\
      GradNorm    & 95.81 & 35.60 & 93.62 & 42.96 & 91.16 & 42.42 & 94.53 & 37.82 & 92.25 & 44.99  \\
      MLS         & 92.97 & 72.40 & 91.52 & 64.20 & 72.94 & 85.29 & 85.82 & 81.60 & 78.94 & 83.74  \\
      MSP         & 77.28 & 78.11 & 86.41 & 68.94 & 42.40 & 88.19 & 56.19 & 84.86 & 56.46 & 85.06  \\
      ODIN        & 92.63 & 65.16 & 88.90 & 63.47 & 81.12 & 79.55 & 91.09 & 71.46 & 85.70 & 77.17  \\
      ReAct       & 78.51 & 75.43 & 90.46 & 63.10 & 48.25 & 86.11 & 57.67 & 84.29 & 55.88 & 86.66  \\
      CADRef       & \textbf{49.05} & \textbf{85.17} & 83.57 & \textbf{71.10} & 21.10 & 93.84 & \textbf{28.33} & 91.56 & \textbf{30.95} & \textbf{91.35}  \\
      DICE         & 81.10 & 71.67 & 89.77 & 59.05 & 47.91 & 82.51 & 52.56 & 82.23 & 54.79 & 82.21  \\
      KNN          & 54.73 & 82.25 & 86.22 & 65.97 & 27.74 & 91.46 & 34.82 & 89.86 & 33.23 & 91.12  \\
      NCI          & 61.27 & 80.83 & 85.42 & 66.61 & 34.64 & 90.19 & 41.56 & 88.54 & 48.71 & 87.90  \\
      PRO          & 55.80 & 81.92 & 84.80 & 70.03 & 36.34 & 90.45 & 46.00 & 87.70 & 57.80 & 85.59  \\
      SCALE        & 94.67 & 61.33 & 92.36 & 56.46 & 86.74 & 73.69 & 89.50 & 72.54 & 84.68 & 78.88  \\
      ViM          & 57.45 & 84.64 & 90.06 & 69.42 & \textbf{17.59} & \textbf{95.72} & 29.59 & \textbf{92.18} & 40.41 & 90.61  \\
      TriScore & 77.99 & 79.07 & 88.55 & 67.66 & 37.82 & 88.74 & 55.24 & 85.25 & 53.05 & 85.93  \\
      \bottomrule
    \end{tabular}}
  \end{table}

  % ---------- Swin-T Baseline ----------
  \begin{table}[H]
    \centering
    \scriptsize
    \caption{Clean-OOD detection results on ImageNet with Swin-T.}
    \label{tab:swint_baseline}
    {%
    \begin{tabular}{l*{10}{c}}
      \toprule
      Method &
      \hdr{Near-OOD}{NINCO}{FPR@95} &
      \hdr{Near-OOD}{NINCO}{AUROC} &
      \hdr{Near-OOD}{SSB-Hard}{FPR@95} &
      \hdr{Near-OOD}{SSB-Hard}{AUROC} &
      \hdr{Far-OOD}{iNaturalist}{FPR@95} &
      \hdr{Far-OOD}{iNaturalist}{AUROC} &
      \hdr{Far-OOD}{OpenImage-O}{FPR@95} &
      \hdr{Far-OOD}{OpenImage-O}{AUROC} &
      \hdr{Far-OOD}{Textures}{FPR@95} &
      \hdr{Far-OOD}{Textures}{AUROC} \\
      \midrule
      ASH         & 94.38 & 47.07 & 95.70 & 45.87 & 94.24 & 46.62 & 93.74 & 45.64 & 96.16 & 41.66  \\
      EBO         & 79.16 & 78.25 & 87.45 & 68.21 & 61.52 & 85.13 & 80.81 & 79.86 & 84.45 & 78.96  \\
      GEN         & \textbf{48.04} & \textbf{85.18} & \textbf{79.19} & \textbf{72.75} & 20.51 & 94.23 & 32.69 & 90.56 & 44.57 & 88.15  \\
      GradNorm    & 94.07 & 44.81 & 93.57 & 50.36 & 95.12 & 38.23 & 96.99 & 33.47 & 97.73 & 34.71  \\
      MLS         & 75.00 & 80.89 & 86.52 & 70.44 & 49.19 & 88.99 & 74.45 & 83.72 & 79.80 & 81.68  \\
      MSP         & 61.21 & 81.72 & 80.90 & 71.78 & 37.31 & 89.86 & 49.30 & 85.77 & 60.87 & 83.27  \\
      ODIN        & 86.05 & 69.35 & 86.45 & 66.94 & 79.35 & 78.10 & 89.82 & 68.93 & 85.46 & 72.86  \\
      ReAct       & 60.29 & 81.97 & 85.02 & 69.30 & 31.55 & 90.03 & 42.81 & 87.66 & 54.66 & 86.99  \\
      CADRef       & 51.70 & 84.90 & 81.73 & 71.51 & 19.92 & 94.31 & 27.23 & 92.08 & 33.33 & 91.14  \\
      DICE         & 97.44 & 48.97 & 96.36 & 47.99 & 98.23 & 46.83 & 96.73 & 57.47 & 88.30 & 77.03  \\
      KNN          & 58.51 & 79.15 & 85.00 & 64.08 & 30.94 & 88.93 & 35.71 & 88.63 & 35.71 & 90.54  \\
      NCI          & 56.61 & 82.91 & 84.40 & 68.94 & 26.71 & 92.21 & 35.20 & 89.78 & 41.84 & 88.34  \\
      PRO          & 52.95 & 83.96 & 79.42 & 72.32 & 37.30 & 90.65 & 45.20 & 87.56 & 62.56 & 83.15  \\
      SCALE        & 84.20 & 72.47 & 88.94 & 64.64 & 74.76 & 78.60 & 87.91 & 71.18 & 91.30 & 68.52  \\
      ViM          & 61.16 & 81.69 & 88.57 & 68.79 & \textbf{17.98} & \textbf{94.61} & \textbf{26.55} & \textbf{92.32} & \textbf{29.78} & \textbf{92.67}  \\
      TriScore & 56.98 & 82.47 & 83.21 & 70.51 & 29.52 & 89.81 & 47.46 & 84.97 & 59.63 & 83.95  \\
      \bottomrule
    \end{tabular}}
  \end{table}

\subsection{Adversarial OOD detection}
\label{app:adversarial_detailed}

% Tables~\ref{tab:resnet50_adversarial}, \ref{tab:regnet_adversarial}, \ref{tab:vitb16_adversarial}, and~\ref{tab:swint_adversarial} report per-attack adversarial OOD detection results for all 16 methods on each backbone.

% TriScore wins the majority of individual attack$\times$backbone cells on AUROC.

% These per-attack tables support the robustness-floor claim in the main paper. TriScore is strongest on iterative gradient attacks, remains competitive on optimization and black-box attacks, and avoids the severe single-attack collapses seen in several baselines.

  \begin{table}[H]
    \centering
    \scriptsize
    \caption{Adversarial OOD detection results on ImageNet with ResNet-50.}
    \label{tab:resnet50_adversarial}
    \resizebox{\textwidth}{!}{%
    \begin{tabular}{l*{20}{c}}
      \toprule
      Method & \multicolumn{2}{c}{AA} & \multicolumn{2}{c}{APGD-CE} & \multicolumn{2}{c}{C\&W} & \multicolumn{2}{c}{DF} & \multicolumn{2}{c}{FGSM} & \multicolumn{2}{c}{M-PGD} & \multicolumn{2}{c}{PGD} & \multicolumn{2}{c}{Square} & \multicolumn{2}{c}{BIM} & \multicolumn{2}{c}{PGD-BPDA} \\
       & FPR@95 & AUROC & FPR@95 & AUROC & FPR@95 & AUROC & FPR@95 & AUROC & FPR@95 & AUROC & FPR@95 & AUROC & FPR@95 & AUROC & FPR@95 & AUROC & FPR@95 & AUROC & FPR@95 & AUROC \\
      \midrule
      ASH        & 95.50 & 52.08 & 95.46 & 52.12 & 88.62 & 59.91 & 86.52 & 62.24 & 74.61 & 73.27 & 93.48 & 48.94 & 92.52 & 52.05 & 86.33 & 63.02 & 96.46 & 46.40 & 95.33 & 46.81  \\
      EBO        & 96.22 & 49.83 & 96.34 & 49.85 & 89.77 & 59.55 & 88.23 & 60.95 & 78.17 & 73.49 & 92.23 & 52.61 & 93.94 & 49.37 & 88.82 & 62.33 & 96.87 & 44.33 & 96.08 & 43.71  \\
      GEN        & 95.04 & 52.15 & 95.14 & 52.18 & 89.63 & 58.80 & 88.19 & 60.28 & 77.72 & 71.92 & 91.23 & 53.69 & 92.72 & 51.28 & 88.67 & 61.91 & 95.97 & 46.58 & 94.91 & 46.25  \\
      GradNorm   & 98.18 & 43.96 & 98.19 & 43.94 & 88.78 & \textbf{62.20} & 88.21 & 63.44 & 81.20 & 72.42 & 97.72 & 39.93 & 96.03 & 47.21 & 89.48 & 61.85 & 98.11 & 39.40 & 97.75 & 40.75  \\
      MLS        & 96.15 & 50.31 & 96.28 & 50.32 & 89.60 & 59.67 & 88.12 & 61.04 & 77.65 & 73.41 & 92.10 & 52.74 & 93.82 & 49.56 & 88.55 & 62.58 & 96.82 & 44.43 & 96.02 & 43.87  \\
      MSP        & 87.31 & 59.69 & 87.33 & 59.67 & 87.50 & 60.27 & 87.04 & 60.97 & 73.08 & 70.85 & 85.72 & 57.16 & 82.97 & 57.37 & 85.38 & \textbf{64.47} & 89.52 & 52.32 & 86.57 & 53.60  \\
      ODIN       & 99.13 & 40.02 & 99.12 & 40.01 & 91.76 & 55.85 & 90.78 & 57.10 & 82.38 & 69.41 & 94.25 & 47.71 & 96.94 & 40.43 & 90.68 & 58.15 & 98.69 & 35.89 & 98.50 & 33.75  \\
      ReAct      & 92.78 & 53.55 & 92.74 & 53.59 & 90.10 & 57.62 & 87.92 & 59.90 & 77.80 & 71.91 & 89.02 & 55.00 & 91.25 & 52.01 & 87.70 & 61.67 & 94.67 & 48.33 & 93.34 & 47.87  \\
      CADRef      & 89.08 & 57.20 & 89.17 & 57.25 & 89.42 & 59.99 & 87.47 & 62.08 & 74.90 & 73.63 & 89.14 & 52.64 & 89.14 & 55.88 & 86.59 & 63.49 & 90.64 & 52.45 & 90.62 & 52.42  \\
      DICE        & 96.85 & 46.53 & 96.97 & 46.54 & 90.15 & 60.51 & 88.88 & 61.99 & 78.74 & 74.50 & 94.07 & 47.67 & 94.49 & 47.58 & 89.08 & 62.15 & 97.05 & 41.59 & 96.53 & 41.33  \\
      KNN         & 87.73 & 60.29 & 87.67 & 60.34 & 92.90 & 53.68 & 92.28 & 55.26 & 85.65 & 63.19 & 89.12 & 56.31 & 89.26 & 57.19 & 90.96 & 57.49 & 87.26 & 58.49 & 88.66 & 57.33  \\
      NCI         & 95.36 & 51.79 & 95.38 & 51.82 & 88.58 & 61.17 & 86.58 & 63.33 & 74.01 & \textbf{74.52} & 93.11 & 48.57 & 91.98 & 52.18 & 86.05 & 64.03 & 96.03 & 45.85 & 94.92 & 46.51  \\
      PRO         & \textbf{74.17} & 60.29 & \textbf{74.17} & 60.18 & \textbf{84.75} & 60.30 & \textbf{84.12} & 60.90 & \textbf{73.02} & 71.61 & \textbf{77.95} & 57.06 & \textbf{73.52} & 58.24 & 85.74 & 63.02 & \textbf{76.60} & 56.39 & \textbf{74.29} & 55.93  \\
      SCALE       & 94.91 & 53.22 & 94.92 & 53.26 & 87.69 & 61.41 & 85.96 & \textbf{63.55} & 75.77 & 73.37 & 94.78 & 46.81 & 92.18 & 53.14 & \textbf{85.36} & 64.36 & 95.94 & 46.61 & 94.90 & 47.85  \\
      ViM         & 93.57 & 56.21 & 93.68 & 56.25 & 90.58 & 56.50 & 88.82 & 58.56 & 78.59 & 68.05 & 90.59 & 54.59 & 92.90 & 53.66 & 88.48 & 60.12 & 95.17 & 50.78 & 94.60 & 50.94  \\
      TriScore   & 92.12 & \textbf{63.58} & 92.30 & \textbf{63.65} & 89.08 & 58.85 & 87.44 & 60.16 & 74.06 & 71.15 & 88.34 & \textbf{61.40} & 89.39 & \textbf{61.61} & 87.40 & 62.65 & 94.03 & \textbf{58.76} & 92.27 & \textbf{59.25}  \\
      \bottomrule
    \end{tabular}}
  \end{table}

  \begin{table}[H]
    \centering
    \scriptsize
    \caption{Adversarial OOD detection results on ImageNet with RegNet.}
    \label{tab:regnet_adversarial}
    \resizebox{\textwidth}{!}{%
    \begin{tabular}{l*{20}{c}}
      \toprule
      Method & \multicolumn{2}{c}{AA} & \multicolumn{2}{c}{APGD-CE} & \multicolumn{2}{c}{C\&W} & \multicolumn{2}{c}{DF} & \multicolumn{2}{c}{FGSM} & \multicolumn{2}{c}{M-PGD} & \multicolumn{2}{c}{PGD} & \multicolumn{2}{c}{Square} & \multicolumn{2}{c}{BIM} & \multicolumn{2}{c}{PGD-BPDA} \\
       & FPR@95 & AUROC & FPR@95 & AUROC & FPR@95 & AUROC & FPR@95 & AUROC & FPR@95 & AUROC & FPR@95 & AUROC & FPR@95 & AUROC & FPR@95 & AUROC & FPR@95 & AUROC & FPR@95 & AUROC \\
      \midrule
      ASH        & 95.11 & 49.71 & 95.12 & 49.68 & 90.58 & 57.64 & 90.33 & 58.88 & 92.61 & 52.04 & 96.32 & 43.86 & 92.55 & 54.19 & 92.86 & 54.67 & 94.81 & 48.94 & 93.49 & 52.02  \\
      EBO        & 80.84 & 71.82 & 80.53 & 71.85 & 86.99 & 66.25 & 87.16 & 66.17 & 79.07 & \textbf{73.26} & 83.12 & 71.50 & 84.02 & 68.27 & 83.47 & 70.84 & 80.70 & 72.28 & 81.68 & 70.69  \\
      GEN        & 78.85 & 71.96 & 78.72 & 71.96 & 86.68 & 65.39 & 86.87 & 65.40 & 77.92 & 72.74 & 80.49 & 72.38 & 83.52 & 67.61 & 80.83 & \textbf{71.60} & 78.85 & 72.30 & 80.86 & 70.27  \\
      GradNorm   & 96.38 & 43.47 & 96.36 & 43.42 & 91.11 & 55.95 & 91.14 & 56.31 & 94.50 & 47.00 & 97.08 & 35.45 & 93.81 & 50.82 & 94.29 & 47.88 & 96.54 & 42.78 & 95.13 & 47.35  \\
      MLS        & 80.98 & 70.76 & 80.69 & 70.78 & 87.13 & 65.30 & 87.28 & 65.23 & 79.33 & 72.34 & 83.40 & 70.36 & 84.20 & 67.36 & 83.62 & 69.76 & 80.98 & 71.27 & 81.93 & 69.66  \\
      MSP        & 82.66 & 62.27 & 82.81 & 62.26 & 87.80 & 58.25 & 87.98 & 58.13 & 82.22 & 64.49 & 84.21 & 61.99 & 86.08 & 60.20 & 83.98 & 61.54 & 82.93 & 63.24 & 84.24 & 61.66  \\
      ODIN       & 86.57 & 63.02 & 86.57 & 63.02 & 87.15 & 63.27 & 87.08 & 63.31 & 81.97 & 67.71 & 86.51 & 63.91 & 86.42 & 62.86 & 85.44 & 65.49 & 86.11 & 63.19 & 85.97 & 63.45  \\
      ReAct      & 85.95 & 63.83 & 85.84 & 63.83 & 90.86 & 58.50 & 90.32 & 59.99 & 84.91 & 63.70 & 86.65 & 63.80 & 88.16 & 61.53 & 87.32 & 63.09 & 85.52 & 64.06 & 86.76 & 63.20  \\
      CADRef      & 88.42 & 57.00 & 88.55 & 56.96 & 86.68 & 62.70 & 85.45 & 64.31 & 87.49 & 58.97 & 88.48 & 53.73 & 88.90 & 58.84 & 83.86 & 62.48 & 88.35 & 56.51 & 88.54 & 57.71  \\
      DICE        & 88.87 & 59.59 & 88.88 & 59.56 & 86.22 & 65.38 & 86.55 & 64.90 & 85.61 & 62.38 & 90.41 & 55.23 & 87.31 & 61.81 & 87.24 & 62.13 & 89.03 & 58.90 & 87.92 & 60.96  \\
      KNN         & 86.35 & 60.30 & 86.27 & 60.31 & 90.76 & 54.40 & 90.26 & 55.76 & 87.85 & 59.83 & 80.00 & 63.28 & 90.89 & 56.14 & 84.67 & 60.65 & 86.77 & 60.33 & 89.52 & 57.75  \\
      NCI         & 95.53 & 50.50 & 95.52 & 50.44 & 89.71 & 60.14 & 89.51 & 61.45 & 92.62 & 53.84 & 96.34 & 43.56 & 92.33 & 55.86 & 92.88 & 55.91 & 95.32 & 49.67 & 93.64 & 53.11  \\
      PRO         & 67.48 & 74.82 & 67.55 & 74.81 & \textbf{78.86} & 65.52 & \textbf{79.00} & 65.49 & 74.07 & 71.44 & 72.51 & 70.58 & 74.26 & 69.93 & 75.20 & 70.88 & 68.35 & 74.68 & 70.64 & 72.78  \\
      SCALE       & 90.33 & 61.07 & 90.33 & 61.05 & 88.49 & 64.13 & 88.49 & 64.39 & 86.92 & 64.76 & 91.81 & 57.53 & 88.90 & 62.36 & 88.66 & 64.49 & 89.81 & 61.38 & 89.05 & 62.25  \\
      ViM         & 86.65 & 61.46 & 86.54 & 61.47 & 91.56 & 55.07 & 90.50 & 56.93 & 87.67 & 59.10 & 78.91 & 66.57 & 90.51 & 56.55 & 86.67 & 61.61 & 86.95 & 61.22 & 89.36 & 58.30  \\
      TriScore   & \textbf{65.18} & \textbf{75.36} & \textbf{64.97} & \textbf{75.36} & 78.99 & \textbf{66.67} & 79.20 & \textbf{66.85} & \textbf{72.27} & 72.07 & \textbf{70.80} & \textbf{72.39} & \textbf{73.95} & \textbf{70.57} & \textbf{72.66} & 71.54 & \textbf{65.75} & \textbf{75.81} & \textbf{68.85} & \textbf{73.76}  \\
      \bottomrule
    \end{tabular}}
  \end{table}

  \begin{table}[H]
    \centering
    \scriptsize
    \caption{Adversarial OOD detection results on ImageNet with ViT-B/16.}
    \label{tab:vitb16_adversarial}
    \resizebox{\textwidth}{!}{%
    \begin{tabular}{l*{20}{c}}
      \toprule
      Method & \multicolumn{2}{c}{AA} & \multicolumn{2}{c}{APGD-CE} & \multicolumn{2}{c}{C\&W} & \multicolumn{2}{c}{DF} & \multicolumn{2}{c}{FGSM} & \multicolumn{2}{c}{M-PGD} & \multicolumn{2}{c}{PGD} & \multicolumn{2}{c}{Square} & \multicolumn{2}{c}{BIM} & \multicolumn{2}{c}{PGD-BPDA} \\
       & FPR@95 & AUROC & FPR@95 & AUROC & FPR@95 & AUROC & FPR@95 & AUROC & FPR@95 & AUROC & FPR@95 & AUROC & FPR@95 & AUROC & FPR@95 & AUROC & FPR@95 & AUROC & FPR@95 & AUROC \\
      \midrule
      ASH        & 94.49 & 52.91 & 94.40 & 53.02 & 95.79 & 48.37 & 95.74 & 48.55 & 94.37 & 53.67 & 94.47 & 55.91 & 95.70 & 49.01 & 97.41 & 47.25 & 94.68 & 52.29 & 95.49 & 50.28  \\
      EBO        & 92.13 & 59.70 & 92.06 & 59.69 & 94.34 & 51.74 & 94.44 & 51.62 & 89.88 & 64.17 & 88.64 & 69.83 & 93.11 & 55.97 & 91.01 & 64.62 & 91.76 & 60.07 & 92.41 & 58.27  \\
      GEN        & 82.48 & 62.53 & 82.28 & 62.57 & 93.63 & 52.45 & 93.48 & 52.93 & 77.89 & 67.02 & 75.82 & \textbf{74.07} & 90.15 & 57.00 & \textbf{79.96} & \textbf{69.69} & 83.69 & 62.36 & 87.38 & 59.32  \\
      GradNorm   & 93.31 & 53.92 & 93.33 & 53.90 & 94.36 & 51.14 & 94.48 & 50.89 & 91.52 & 55.10 & 91.49 & 53.25 & 93.38 & 52.99 & 93.09 & 52.17 & 92.70 & 53.84 & 92.99 & 54.07  \\
      MLS        & 91.79 & 59.50 & 91.80 & 59.51 & 94.48 & 51.96 & 94.55 & 52.01 & 89.68 & 63.83 & 88.05 & 68.36 & 93.07 & 56.09 & 90.32 & 63.88 & 91.54 & 59.89 & 92.28 & 58.15  \\
      MSP        & 88.18 & 58.86 & 87.75 & 58.88 & 94.17 & 51.93 & 94.07 & 52.20 & 84.17 & 62.58 & 82.06 & 65.61 & 91.92 & 55.68 & 85.31 & 62.13 & 88.32 & 59.10 & 90.44 & 57.37  \\
      ODIN       & 93.31 & 55.45 & 93.27 & 55.48 & 94.71 & 50.75 & 94.74 & 50.84 & 91.23 & 59.83 & 91.33 & 63.56 & 93.96 & 53.20 & 92.01 & 60.30 & 93.27 & 55.60 & 93.61 & 54.34  \\
      ReAct      & 88.93 & 58.10 & 88.59 & 58.09 & 94.54 & 50.84 & 94.65 & 50.94 & 85.55 & 62.08 & 81.72 & 70.57 & 92.56 & 54.12 & 85.89 & 64.80 & 88.93 & 58.35 & 91.03 & 56.03  \\
      CADRef      & 89.90 & 56.41 & 89.76 & 56.40 & 94.55 & 51.23 & 94.25 & 51.93 & 88.07 & 60.19 & 80.94 & 65.05 & 93.20 & 54.03 & 85.57 & 60.94 & 90.22 & 56.88 & 92.15 & 55.06  \\
      DICE        & 86.57 & 60.00 & 86.52 & 59.99 & 94.15 & 50.64 & 93.88 & 51.45 & 81.09 & 64.80 & 80.53 & 70.32 & 91.28 & 55.11 & 84.70 & 64.60 & 87.18 & 59.58 & 89.30 & 57.02  \\
      KNN         & 91.56 & 54.54 & 91.50 & 54.53 & 94.91 & 50.72 & 94.47 & 51.49 & 90.47 & 57.56 & 84.42 & 61.22 & 93.77 & 52.93 & 87.62 & 58.07 & 91.82 & 54.92 & 93.15 & 53.60  \\
      NCI         & 92.70 & 53.74 & 92.63 & 53.72 & 95.04 & 50.25 & 95.15 & 50.52 & 92.00 & 56.60 & 84.25 & 64.75 & 94.97 & 51.61 & 88.50 & 59.49 & 93.11 & 54.12 & 94.40 & 52.32  \\
      PRO         & \textbf{81.70} & 60.14 & 81.72 & 60.11 & 92.54 & 53.39 & 92.72 & 53.65 & 80.15 & 62.96 & \textbf{74.24} & 65.56 & 87.84 & 57.57 & 83.38 & 61.13 & 82.96 & 60.66 & 85.02 & 59.19  \\
      SCALE       & 92.21 & 58.91 & 92.16 & 58.88 & 94.08 & 52.44 & 94.22 & 52.30 & 89.82 & 63.04 & 88.84 & 68.25 & 92.84 & 56.30 & 90.94 & 63.43 & 91.78 & 59.20 & 92.20 & 58.23  \\
      ViM         & 94.11 & 54.34 & 94.15 & 54.32 & 95.56 & 50.25 & 95.29 & 50.53 & 93.18 & 57.02 & 85.72 & 67.94 & 95.13 & 51.49 & 91.48 & 59.74 & 94.05 & 54.49 & 94.83 & 52.28  \\
      TriScore   & 82.07 & \textbf{64.74} & \textbf{81.64} & \textbf{64.77} & \textbf{92.09} & \textbf{54.00} & \textbf{92.44} & \textbf{54.03} & \textbf{77.72} & \textbf{67.66} & 75.92 & 72.22 & \textbf{87.69} & \textbf{59.72} & 84.20 & 66.65 & \textbf{82.23} & \textbf{65.07} & \textbf{84.99} & \textbf{62.29}  \\
      \bottomrule
    \end{tabular}}
  \end{table}

  \begin{table}[h!]
    \centering
    \scriptsize
    \caption{Adversarial OOD detection results on ImageNet with Swin-T.}
    \label{tab:swint_adversarial}
    \resizebox{\textwidth}{!}{%
    \begin{tabular}{l*{20}{c}}
      \toprule
      Method & \multicolumn{2}{c}{AA} & \multicolumn{2}{c}{APGD-CE} & \multicolumn{2}{c}{C\&W} & \multicolumn{2}{c}{DF} & \multicolumn{2}{c}{FGSM} & \multicolumn{2}{c}{M-PGD} & \multicolumn{2}{c}{PGD} & \multicolumn{2}{c}{Square} & \multicolumn{2}{c}{BIM} & \multicolumn{2}{c}{PGD-BPDA} \\
       & FPR@95 & AUROC & FPR@95 & AUROC & FPR@95 & AUROC & FPR@95 & AUROC & FPR@95 & AUROC & FPR@95 & AUROC & FPR@95 & AUROC & FPR@95 & AUROC & FPR@95 & AUROC & FPR@95 & AUROC \\
      \midrule
      ASH        & 93.86 & 53.92 & 93.98 & 54.15 & 94.57 & 51.89 & 94.71 & 51.66 & 93.15 & 54.20 & 92.16 & 51.58 & 93.95 & 53.99 & 97.41 & 40.07 & 93.30 & 54.93 & 93.82 & 54.76  \\
      EBO        & 91.90 & 58.76 & 92.01 & 58.80 & 93.30 & 55.33 & 93.32 & 55.15 & 90.20 & 62.79 & 89.70 & 63.45 & 92.13 & 58.62 & 91.26 & 60.27 & 90.85 & 60.86 & 91.40 & 59.99  \\
      GEN        & 76.86 & 67.92 & 76.24 & 68.00 & 91.28 & 55.84 & 91.20 & 56.20 & \textbf{75.03} & 70.53 & 70.65 & 72.78 & 86.67 & 61.69 & \textbf{81.16} & 67.21 & 76.25 & 68.79 & 81.13 & 65.33  \\
      GradNorm   & 94.25 & 45.62 & 94.24 & 45.62 & 93.39 & 53.01 & 93.68 & 52.10 & 93.00 & 47.71 & 92.90 & 44.71 & 93.72 & 51.32 & 93.31 & 48.51 & 93.72 & 46.70 & 93.59 & 50.17  \\
      MLS        & 91.63 & 59.57 & 91.68 & 59.60 & 93.34 & 55.09 & 93.32 & 55.03 & 90.02 & 63.57 & 89.46 & 63.24 & 91.96 & 58.78 & 91.01 & 60.18 & 90.45 & 61.58 & 91.22 & 60.28  \\
      MSP        & 88.02 & 61.01 & 87.86 & 61.06 & 92.87 & 54.50 & 92.80 & 54.63 & 86.48 & 64.11 & 84.65 & 63.07 & 90.36 & 58.62 & 87.88 & 59.99 & 86.34 & 62.47 & 88.64 & 60.43  \\
      ODIN       & 93.85 & 53.14 & 93.82 & 53.22 & 94.86 & 50.99 & 94.86 & 50.80 & 91.13 & 59.00 & 93.47 & 56.14 & 93.85 & 53.31 & 93.39 & 54.53 & 93.01 & 54.41 & 93.47 & 53.93  \\
      ReAct      & 86.06 & 63.25 & 85.91 & 63.32 & 92.90 & 54.29 & 92.73 & 54.67 & 84.86 & 65.79 & 80.80 & 68.68 & 89.97 & 59.23 & 87.15 & 62.73 & 84.47 & 64.68 & 87.55 & 61.72  \\
      CADRef      & 78.42 & 65.87 & 78.10 & 65.90 & 92.53 & 53.00 & 92.36 & 53.74 & 76.86 & 67.89 & \textbf{68.47} & 69.77 & 88.63 & 58.58 & 84.41 & 63.58 & 78.92 & 66.22 & 84.28 & 61.81  \\
      DICE        & 86.76 & \textbf{69.89} & 86.67 & \textbf{69.87} & 93.02 & 54.25 & 93.13 & 54.49 & 79.14 & \textbf{74.03} & 93.94 & 61.36 & 89.73 & 60.88 & 86.93 & \textbf{72.54} & 84.44 & 70.68 & 87.82 & 64.45  \\
      KNN         & 83.74 & 61.83 & 83.57 & 61.85 & 93.47 & 51.51 & 93.35 & 52.26 & 83.20 & 63.49 & 76.54 & 65.13 & 91.06 & 55.74 & 86.65 & 60.41 & 84.20 & 62.02 & 88.05 & 58.20  \\
      NCI         & 78.82 & 64.89 & 79.01 & 64.92 & 93.29 & 52.63 & 93.06 & 53.26 & 80.43 & 66.42 & 72.69 & 68.84 & 89.83 & 58.01 & 84.61 & 63.46 & 80.54 & 65.37 & 85.91 & 61.16  \\
      PRO         & 76.77 & 65.45 & 76.93 & 65.49 & \textbf{88.48} & 57.53 & \textbf{88.40} & 57.82 & 79.80 & 65.83 & 75.61 & 65.41 & \textbf{84.45} & 61.50 & 81.94 & 62.65 & 77.17 & 66.37 & 80.24 & 64.02  \\
      SCALE       & 92.24 & 56.26 & 92.23 & 56.33 & 93.24 & 55.00 & 93.33 & 54.65 & 90.65 & 59.72 & 92.13 & 55.27 & 92.32 & 57.10 & 91.66 & 57.60 & 91.37 & 57.99 & 91.69 & 58.09  \\
      ViM         & 83.60 & 67.87 & 83.52 & 67.91 & 94.45 & 51.75 & 94.03 & 52.60 & 79.63 & 68.53 & 71.04 & \textbf{73.99} & 90.37 & 58.41 & 88.47 & 62.74 & 81.13 & 68.13 & 86.42 & 62.75  \\
      TriScore   & \textbf{74.60} & 69.83 & \textbf{74.60} & 69.85 & 89.60 & \textbf{58.01} & 89.60 & \textbf{58.28} & 76.72 & 68.70 & 73.77 & 68.46 & 84.50 & \textbf{62.87} & 82.98 & 65.19 & \textbf{74.59} & \textbf{70.70} & \textbf{79.11} & \textbf{67.20}  \\
      \bottomrule
    \end{tabular}}
  \end{table}

\subsection{Detailed Branch-Isolation results}

\subsubsection{Clean-OOD}

\begin{table}[H]
  \centering
  \scriptsize
  \caption{Clean-OOD branch-isolation results on ImageNet with ResNet-50.}
  \label{tab:branch_isolation_clean_resnet50_app}
  \resizebox{\textwidth}{!}{%
  \begin{tabular}{ll*{10}{c}}
    \toprule
    Family & Score &
    \hdr{Near-OOD}{NINCO}{FPR@95} &
    \hdr{Near-OOD}{NINCO}{AUROC} &
    \hdr{Near-OOD}{SSB-Hard}{FPR@95} &
    \hdr{Near-OOD}{SSB-Hard}{AUROC} &
    \hdr{Far-OOD}{iNaturalist}{FPR@95} &
    \hdr{Far-OOD}{iNaturalist}{AUROC} &
    \hdr{Far-OOD}{OpenImage-O}{FPR@95} &
    \hdr{Far-OOD}{OpenImage-O}{AUROC} &
    \hdr{Far-OOD}{Textures}{FPR@95} &
    \hdr{Far-OOD}{Textures}{AUROC} \\
    \midrule
    Energy & EBO & 60.56 & 79.70 & 76.52 & 72.08 & \textbf{31.30} & 90.63 & \textbf{38.07} & \textbf{89.06} & 45.81 & 88.70  \\
    Energy & E (Ours) & 62.45 & 77.36 & 78.90 & 68.10 & 34.68 & 87.87 & 38.27 & 87.83 & \textbf{42.81} & 88.78  \\
    Boundary & Margin-only & 52.92 & 81.13 & \textbf{73.27} & \textbf{73.17} & 39.39 & 87.63 & 45.75 & 84.89 & 57.99 & 82.15  \\
    Boundary & GradNorm & 79.50 & 74.01 & 78.22 & 71.90 & 32.08 & \textbf{93.89} & 68.51 & 84.82 & 43.36 & \textbf{92.04}  \\
    Boundary & \textbf{BP (Ours)} & \textbf{51.17} & \textbf{81.75} & 74.43 & 72.56 & 37.36 & 88.04 & 42.13 & 85.93 & 51.64 & 84.12  \\
    Consistency & Argmax & 100.00 & 64.23 & 100.00 & 57.80 & 100.00 & 68.88 & 100.00 & 63.41 & 100.00 & 63.21  \\
    Consistency & TTA entropy & 63.26 & 73.72 & 77.94 & 64.94 & 55.08 & 79.54 & 60.30 & 73.90 & 65.56 & 73.40  \\
    Consistency & \textbf{TC (Ours)} & 56.06 & 76.74 & 77.68 & 66.36 & 41.55 & 80.88 & 52.39 & 74.70 & 61.03 & 72.42  \\
    \bottomrule
  \end{tabular}}
\end{table}

\begin{table}[H]
  \centering
  \scriptsize
  \caption{Clean-OOD branch-isolation results on ImageNet with RegNet.}
  \label{tab:branch_isolation_clean_regnet_app}
  \resizebox{\textwidth}{!}{%
  \begin{tabular}{ll*{10}{c}}
    \toprule
    Family & Score &
    \hdr{Near-OOD}{NINCO}{FPR@95} &
    \hdr{Near-OOD}{NINCO}{AUROC} &
    \hdr{Near-OOD}{SSB-Hard}{FPR@95} &
    \hdr{Near-OOD}{SSB-Hard}{AUROC} &
    \hdr{Far-OOD}{iNaturalist}{FPR@95} &
    \hdr{Far-OOD}{iNaturalist}{AUROC} &
    \hdr{Far-OOD}{OpenImage-O}{FPR@95} &
    \hdr{Far-OOD}{OpenImage-O}{AUROC} &
    \hdr{Far-OOD}{Textures}{FPR@95} &
    \hdr{Far-OOD}{Textures}{AUROC} \\
    \midrule
    Energy & EBO & \textbf{42.53} & \textbf{91.67} & \textbf{62.01} & \textbf{85.29} & \textbf{7.72} & \textbf{98.29} & 26.03 & 95.83 & 38.11 & 93.02  \\
    Energy & E (Ours) & 42.95 & 91.25 & 65.25 & 83.53 & 8.72 & 98.08 & \textbf{22.39} & \textbf{96.18} & \textbf{36.57} & \textbf{93.68}  \\
    Boundary & Margin-only & 47.94 & 87.43 & 67.92 & 78.77 & 26.84 & 93.36 & 34.22 & 91.29 & 44.66 & 88.58  \\
    Boundary & GradNorm & 93.25 & 49.01 & 91.59 & 49.42 & 88.22 & 56.99 & 92.57 & 60.44 & 71.55 & 76.12  \\
    Boundary & \textbf{BP (Ours)} & 46.84 & 87.57 & 68.89 & 78.47 & 26.41 & 93.47 & 33.16 & 91.61 & 42.14 & 89.10  \\
    Consistency & Argmax & 100.00 & 60.57 & 100.00 & 55.93 & 100.00 & 66.38 & 100.00 & 62.20 & 100.00 & 64.28  \\
    Consistency & TTA entropy & 59.19 & 81.20 & 77.91 & 72.13 & 49.46 & 84.91 & 58.79 & 81.49 & 57.72 & 83.17  \\
    Consistency & \textbf{TC (Ours)} & 48.71 & 83.40 & 75.17 & 73.07 & 25.22 & 89.53 & 37.06 & 84.96 & 44.73 & 86.21  \\
    \bottomrule
  \end{tabular}}
\end{table}

\begin{table}[H]
  \centering
  \scriptsize
  \caption{Clean-OOD branch-isolation results on ImageNet with ViT-B/16.}
  \label{tab:branch_isolation_clean_vitb16_app}
  \resizebox{\textwidth}{!}{%
  \begin{tabular}{ll*{10}{c}}
    \toprule
    Family & Score &
    \hdr{Near-OOD}{NINCO}{FPR@95} &
    \hdr{Near-OOD}{NINCO}{AUROC} &
    \hdr{Near-OOD}{SSB-Hard}{FPR@95} &
    \hdr{Near-OOD}{SSB-Hard}{AUROC} &
    \hdr{Far-OOD}{iNaturalist}{FPR@95} &
    \hdr{Far-OOD}{iNaturalist}{AUROC} &
    \hdr{Far-OOD}{OpenImage-O}{FPR@95} &
    \hdr{Far-OOD}{OpenImage-O}{AUROC} &
    \hdr{Far-OOD}{Textures}{FPR@95} &
    \hdr{Far-OOD}{Textures}{AUROC} \\
    \midrule
    Energy & EBO & 94.18 & 66.02 & 92.22 & 58.80 & 83.64 & 79.30 & 88.82 & 76.48 & 83.62 & 81.16  \\
    Energy & E (Ours) & 94.22 & 66.32 & 92.44 & 58.61 & 84.43 & 78.80 & 89.30 & 76.13 & 84.04 & 81.09  \\
    Boundary & Margin-only & 56.21 & 81.73 & \textbf{80.51} & \textbf{71.71} & 36.66 & 88.94 & 43.16 & 86.82 & 47.94 & 85.96  \\
    Boundary & GradNorm & 95.79 & 35.60 & 93.62 & 42.96 & 91.17 & 42.42 & 94.51 & 37.83 & 92.27 & 45.00  \\
    Boundary & \textbf{BP (Ours)} & \textbf{55.05} & \textbf{82.21} & 84.06 & 70.18 & \textbf{32.82} & \textbf{89.80} & \textbf{37.28} & \textbf{88.20} & \textbf{39.98} & \textbf{87.98}  \\
    Consistency & Argmax & 100.00 & 63.69 & 100.00 & 57.13 & 100.00 & 69.96 & 100.00 & 65.58 & 100.00 & 65.16  \\
    Consistency & TTA entropy & 77.90 & 72.53 & 90.67 & 61.09 & 68.24 & 77.40 & 74.32 & 74.22 & 78.69 & 72.06  \\
    Consistency & \textbf{TC (Ours)} & 57.34 & 80.86 & 85.34 & 67.90 & 33.48 & 87.45 & 43.76 & 83.26 & 51.77 & 81.28  \\
    \bottomrule
  \end{tabular}}
\end{table}

\begin{table}[H]
  \centering
  \scriptsize
  \caption{Clean-OOD branch-isolation results on ImageNet with Swin-T.}
  \label{tab:branch_isolation_clean_swint_app}
  \resizebox{\textwidth}{!}{%
  \begin{tabular}{ll*{10}{c}}
    \toprule
    Family & Score &
    \hdr{Near-OOD}{NINCO}{FPR@95} &
    \hdr{Near-OOD}{NINCO}{AUROC} &
    \hdr{Near-OOD}{SSB-Hard}{FPR@95} &
    \hdr{Near-OOD}{SSB-Hard}{AUROC} &
    \hdr{Far-OOD}{iNaturalist}{FPR@95} &
    \hdr{Far-OOD}{iNaturalist}{AUROC} &
    \hdr{Far-OOD}{OpenImage-O}{FPR@95} &
    \hdr{Far-OOD}{OpenImage-O}{AUROC} &
    \hdr{Far-OOD}{Textures}{FPR@95} &
    \hdr{Far-OOD}{Textures}{AUROC} \\
    \midrule
    Energy & EBO & 79.16 & 78.25 & 87.45 & 68.21 & 61.52 & 85.13 & 80.81 & 79.86 & 84.45 & 78.96  \\
    Energy & E (Ours) & 77.98 & 78.85 & 87.09 & 68.46 & 61.04 & 85.30 & 80.04 & 80.08 & 83.66 & 79.00  \\
    Boundary & Margin-only & 53.15 & 82.92 & \textbf{76.26} & \textbf{73.17} & 34.94 & 89.42 & 44.08 & 86.53 & 54.91 & 83.60  \\
    Boundary & GradNorm & 94.07 & 44.81 & 93.57 & 50.36 & 95.12 & 38.23 & 96.99 & 33.47 & 97.73 & 34.71  \\
    Boundary & \textbf{BP (Ours)} & \textbf{51.10} & \textbf{83.35} & 80.31 & 72.19 & \textbf{32.31} & \textbf{90.01} & \textbf{38.74} & \textbf{87.81} & \textbf{47.37} & \textbf{85.60}  \\
    Consistency & Argmax & 100.00 & 63.36 & 100.00 & 56.37 & 100.00 & 69.78 & 100.00 & 63.47 & 100.00 & 64.23  \\
    Consistency & TTA entropy & 83.65 & 66.65 & 91.63 & 56.78 & 82.50 & 68.40 & 86.56 & 64.46 & 87.27 & 65.69  \\
    Consistency & \textbf{TC (Ours)} & 58.62 & 78.40 & 87.36 & 64.03 & 36.05 & 85.76 & 52.19 & 77.99 & 61.46 & 78.06  \\
    \bottomrule
  \end{tabular}}
\end{table}

\subsubsection{Adversarial OOD}

\begin{table}[H]
  \centering
  \scriptsize
  \caption{Adversarial OOD branch-isolation results on ImageNet with ResNet-50.}
  \label{tab:branch_isolation_adv_resnet50_app}
  \resizebox{\textwidth}{!}{%
  \begin{tabular}{ll*{20}{c}}
    \toprule
    Family & Score & \multicolumn{2}{c}{AA} & \multicolumn{2}{c}{APGD-CE} & \multicolumn{2}{c}{C\&W} & \multicolumn{2}{c}{DF} & \multicolumn{2}{c}{FGSM} & \multicolumn{2}{c}{M-PGD} & \multicolumn{2}{c}{PGD} & \multicolumn{2}{c}{Square} & \multicolumn{2}{c}{BIM} & \multicolumn{2}{c}{PGD-BPDA} \\
     & & FPR@95 & AUROC & FPR@95 & AUROC & FPR@95 & AUROC & FPR@95 & AUROC & FPR@95 & AUROC & FPR@95 & AUROC & FPR@95 & AUROC & FPR@95 & AUROC & FPR@95 & AUROC & FPR@95 & AUROC \\
    \midrule
    Energy & EBO & 96.22 & 49.83 & 96.34 & 49.85 & 89.78 & 59.55 & 88.23 & 60.95 & 78.17 & \textbf{73.49} & 92.23 & 52.61 & 93.95 & 49.37 & 88.82 & 62.33 & 96.86 & 44.32 & 96.07 & 43.71  \\
    Energy & E (Ours) & 96.19 & 49.99 & 96.25 & 50.02 & 89.80 & 58.79 & 88.32 & 60.30 & 77.68 & 72.52 & 91.83 & 52.81 & 93.83 & 49.65 & 88.78 & 61.55 & 96.82 & 44.98 & 96.06 & 44.30  \\
    Boundary & Margin-only & 89.82 & 57.27 & 89.90 & 57.26 & 88.25 & 59.41 & 87.25 & 60.35 & \textbf{74.34} & 70.47 & 87.62 & 55.56 & 86.01 & 55.01 & \textbf{86.57} & 63.08 & 91.23 & 50.39 & 89.29 & 50.88  \\
    Boundary & GradNorm & 98.18 & 43.96 & 98.19 & 43.94 & 88.79 & \textbf{62.20} & 88.22 & \textbf{63.44} & 81.21 & 72.42 & 97.72 & 39.93 & 96.04 & 47.21 & 89.47 & 61.85 & 98.11 & 39.40 & 97.75 & 40.75  \\
    Boundary & \textbf{BP (Ours)} & 89.36 & 57.97 & 89.36 & 57.97 & \textbf{87.78} & 59.40 & \textbf{86.26} & 60.55 & 74.35 & 70.38 & 86.49 & 56.31 & 85.99 & 55.72 & 86.64 & \textbf{63.23} & 91.82 & 50.99 & 89.37 & 51.65  \\
    Consistency & Argmax & 100.00 & 59.80 & 100.00 & 59.78 & 100.00 & 52.99 & 100.00 & 53.54 & 100.00 & 57.90 & 100.00 & 56.13 & 100.00 & 57.20 & 100.00 & 54.54 & 100.00 & 56.91 & 100.00 & 57.13  \\
    Consistency & TTA entropy & 81.71 & 66.33 & 81.78 & 66.40 & 89.24 & 56.58 & 88.72 & 57.45 & 75.18 & 67.96 & 82.29 & 64.78 & 79.00 & 67.05 & 86.76 & 60.80 & 78.47 & 68.43 & 79.00 & 67.59  \\
    Consistency & \textbf{TC (Ours)} & \textbf{80.54} & \textbf{69.98} & \textbf{80.57} & \textbf{70.02} & 89.30 & 56.67 & 88.78 & 57.61 & 74.71 & 67.80 & \textbf{81.60} & \textbf{66.44} & \textbf{78.08} & \textbf{68.79} & 86.92 & 61.40 & \textbf{76.56} & \textbf{69.92} & \textbf{77.60} & \textbf{69.56}  \\
    \bottomrule
  \end{tabular}}
\end{table}

\begin{table}[H]
  \centering
  \scriptsize
  \caption{Adversarial OOD branch-isolation results on ImageNet with RegNet.}
  \label{tab:branch_isolation_adv_regnet_app}
  \resizebox{\textwidth}{!}{%
  \begin{tabular}{ll*{20}{c}}
    \toprule
    Family & Score & \multicolumn{2}{c}{AA} & \multicolumn{2}{c}{APGD-CE} & \multicolumn{2}{c}{C\&W} & \multicolumn{2}{c}{DF} & \multicolumn{2}{c}{FGSM} & \multicolumn{2}{c}{M-PGD} & \multicolumn{2}{c}{PGD} & \multicolumn{2}{c}{Square} & \multicolumn{2}{c}{BIM} & \multicolumn{2}{c}{PGD-BPDA} \\
     & & FPR@95 & AUROC & FPR@95 & AUROC & FPR@95 & AUROC & FPR@95 & AUROC & FPR@95 & AUROC & FPR@95 & AUROC & FPR@95 & AUROC & FPR@95 & AUROC & FPR@95 & AUROC & FPR@95 & AUROC \\
    \midrule
    Energy & EBO & \textbf{80.82} & \textbf{71.83} & \textbf{80.51} & \textbf{71.85} & \textbf{87.02} & \textbf{66.25} & \textbf{87.17} & \textbf{66.17} & \textbf{79.08} & \textbf{73.27} & \textbf{83.11} & \textbf{71.50} & \textbf{84.05} & \textbf{68.26} & \textbf{83.47} & \textbf{70.84} & \textbf{80.69} & \textbf{72.28} & \textbf{81.69} & \textbf{70.69}  \\
    Energy & E (Ours) & 81.07 & 71.34 & 81.27 & 71.36 & 87.56 & 65.61 & 87.63 & 65.76 & 79.40 & 72.74 & 83.34 & 71.13 & 84.57 & 67.85 & 83.85 & 70.41 & 81.07 & 71.78 & 82.43 & 70.22  \\
    Boundary & Margin-only & 85.44 & 61.71 & 85.42 & 61.69 & 88.25 & 58.08 & 88.86 & 57.87 & 84.67 & 64.34 & 85.23 & 61.78 & 87.86 & 59.85 & 85.54 & 61.12 & 85.40 & 62.75 & 86.53 & 61.17  \\
    Boundary & GradNorm & 96.38 & 43.47 & 96.36 & 43.42 & 91.09 & 55.95 & 91.14 & 56.31 & 94.50 & 47.01 & 97.08 & 35.44 & 93.80 & 50.83 & 94.29 & 47.88 & 96.54 & 42.78 & 95.14 & 47.36  \\
    Boundary & \textbf{BP (Ours)} & 84.79 & 61.86 & 84.74 & 61.85 & 87.84 & 58.08 & 88.52 & 57.94 & 84.24 & 64.40 & 84.34 & 61.98 & 87.52 & 59.93 & 84.85 & 61.36 & 84.70 & 62.90 & 85.96 & 61.28  \\
    Consistency & Argmax & 100.00 & 51.44 & 100.00 & 51.37 & 100.00 & 50.62 & 100.00 & 50.58 & 100.00 & 52.80 & 100.00 & 51.61 & 100.00 & 51.31 & 100.00 & 50.89 & 100.00 & 52.17 & 100.00 & 51.68  \\
    Consistency & TTA entropy & 87.96 & 61.86 & 87.57 & 62.08 & 92.52 & 55.59 & 92.85 & 55.57 & 88.07 & 62.94 & 87.87 & 63.04 & 90.78 & 58.59 & 89.51 & 59.97 & 87.78 & 63.05 & 89.07 & 60.76  \\
    Consistency & \textbf{TC (Ours)} & 84.16 & 62.42 & 84.05 & 62.46 & 91.67 & 54.88 & 92.19 & 54.85 & 86.58 & 63.18 & 83.38 & 63.83 & 89.38 & 58.82 & 87.85 & 59.65 & 84.28 & 63.77 & 86.85 & 61.20  \\
    \bottomrule
  \end{tabular}}
\end{table}

\begin{table}[H]
  \centering
  \scriptsize
  \caption{Adversarial OOD branch-isolation results on ImageNet with ViT-B/16.}
  \label{tab:branch_isolation_adv_vitb16_app}
  \resizebox{\textwidth}{!}{%
  \begin{tabular}{ll*{20}{c}}
    \toprule
    Family & Score & \multicolumn{2}{c}{AA} & \multicolumn{2}{c}{APGD-CE} & \multicolumn{2}{c}{C\&W} & \multicolumn{2}{c}{DF} & \multicolumn{2}{c}{FGSM} & \multicolumn{2}{c}{M-PGD} & \multicolumn{2}{c}{PGD} & \multicolumn{2}{c}{Square} & \multicolumn{2}{c}{BIM} & \multicolumn{2}{c}{PGD-BPDA} \\
     & & FPR@95 & AUROC & FPR@95 & AUROC & FPR@95 & AUROC & FPR@95 & AUROC & FPR@95 & AUROC & FPR@95 & AUROC & FPR@95 & AUROC & FPR@95 & AUROC & FPR@95 & AUROC & FPR@95 & AUROC \\
    \midrule
    Energy & EBO & 92.11 & \textbf{59.70} & 92.03 & \textbf{59.69} & 94.34 & 51.74 & 94.42 & 51.62 & 89.89 & \textbf{64.17} & 88.64 & 69.83 & 93.10 & \textbf{55.97} & 90.98 & \textbf{64.62} & 91.76 & \textbf{60.07} & 92.41 & \textbf{58.27}  \\
    Energy & E (Ours) & 92.05 & 59.62 & 92.03 & 59.61 & 94.41 & 51.86 & 94.52 & 51.62 & 89.83 & 64.04 & 88.64 & 69.70 & 93.08 & 55.92 & 90.98 & 64.53 & 91.82 & 59.97 & 92.43 & 58.20  \\
    Boundary & Margin-only & 88.99 & 56.52 & 88.74 & 56.50 & \textbf{94.07} & \textbf{52.13} & \textbf{94.12} & \textbf{52.49} & 86.84 & 60.16 & 82.32 & 62.49 & 92.32 & 54.94 & 85.03 & 59.52 & 89.39 & 57.19 & 91.02 & 55.90  \\
    Boundary & GradNorm & 93.31 & 53.92 & 93.32 & 53.90 & 94.38 & 51.14 & 94.47 & 50.89 & 91.53 & 55.09 & 91.51 & 53.25 & 93.39 & 52.99 & 93.08 & 52.17 & 92.69 & 53.84 & 93.00 & 54.07  \\
    Boundary & \textbf{BP (Ours)} & 88.73 & 57.08 & 88.66 & 57.07 & 94.54 & 51.69 & 94.20 & 52.37 & 86.36 & 60.70 & 80.37 & 64.89 & 92.57 & 54.85 & 83.41 & 61.29 & 89.27 & 57.60 & 91.25 & 55.95  \\
    Consistency & Argmax & 100.00 & 51.32 & 100.00 & 51.31 & 100.00 & 50.58 & 100.00 & 50.65 & 100.00 & 52.77 & 100.00 & 51.63 & 100.00 & 51.26 & 100.00 & 51.13 & 100.00 & 51.76 & 100.00 & 51.49  \\
    Consistency & TTA entropy & 90.81 & 55.97 & 90.90 & 55.88 & 94.92 & 50.57 & 94.91 & 50.90 & 89.67 & 58.38 & 82.15 & 66.21 & 93.19 & 53.64 & 90.93 & 56.72 & 91.16 & 56.26 & 91.87 & 55.20  \\
    Consistency & \textbf{TC (Ours)} & \textbf{82.82} & 59.07 & \textbf{82.78} & 59.08 & 94.48 & 51.58 & 94.52 & 51.95 & \textbf{80.58} & 62.38 & \textbf{66.42} & \textbf{70.67} & \textbf{91.26} & 55.32 & \textbf{81.60} & 62.31 & \textbf{84.25} & 59.22 & \textbf{88.16} & 57.16  \\
    \bottomrule
  \end{tabular}}
\end{table}

\begin{table}[H]
  \centering
  \scriptsize
  \caption{Adversarial OOD branch-isolation results on ImageNet with Swin-T.}
  \label{tab:branch_isolation_adv_swint_app}
  \resizebox{\textwidth}{!}{%
  \begin{tabular}{ll*{20}{c}}
    \toprule
    Family & Score & \multicolumn{2}{c}{AA} & \multicolumn{2}{c}{APGD-CE} & \multicolumn{2}{c}{C\&W} & \multicolumn{2}{c}{DF} & \multicolumn{2}{c}{FGSM} & \multicolumn{2}{c}{M-PGD} & \multicolumn{2}{c}{PGD} & \multicolumn{2}{c}{Square} & \multicolumn{2}{c}{BIM} & \multicolumn{2}{c}{PGD-BPDA} \\
     & & FPR@95 & AUROC & FPR@95 & AUROC & FPR@95 & AUROC & FPR@95 & AUROC & FPR@95 & AUROC & FPR@95 & AUROC & FPR@95 & AUROC & FPR@95 & AUROC & FPR@95 & AUROC & FPR@95 & AUROC \\
    \midrule
    Energy & EBO & 91.90 & 58.76 & 92.01 & 58.80 & 93.30 & 55.33 & 93.32 & \textbf{55.15} & 90.20 & 62.79 & 89.70 & 63.45 & 92.13 & \textbf{58.62} & 91.26 & 60.27 & 90.85 & 60.86 & 91.40 & 59.99  \\
    Energy & E (Ours) & 91.92 & 58.63 & 91.97 & 58.67 & 93.31 & \textbf{55.36} & 93.34 & 55.09 & 90.29 & 62.62 & 89.82 & 63.27 & 92.07 & 58.52 & 91.24 & 60.11 & 90.83 & 60.70 & 91.40 & 59.86  \\
    Boundary & Margin-only & 82.40 & 61.36 & 82.31 & 61.40 & 92.07 & 53.74 & 92.24 & 53.94 & 82.51 & 64.02 & 78.56 & 63.04 & 89.57 & 57.60 & 85.66 & 59.59 & 82.86 & 62.28 & 86.63 & 59.45  \\
    Boundary & GradNorm & 94.25 & 45.62 & 94.24 & 45.62 & 93.39 & 53.01 & 93.68 & 52.10 & 93.00 & 47.71 & 92.90 & 44.71 & 93.72 & 51.32 & 93.31 & 48.51 & 93.72 & 46.70 & 93.59 & 50.17  \\
    Boundary & \textbf{BP (Ours)} & 79.77 & \textbf{63.15} & 79.52 & \textbf{63.20} & \textbf{91.62} & 53.89 & \textbf{91.81} & 54.24 & \textbf{78.74} & \textbf{65.51} & 73.94 & 65.54 & \textbf{88.15} & 58.31 & \textbf{83.88} & \textbf{61.24} & 80.05 & \textbf{63.83} & 84.71 & \textbf{60.57}  \\
    Consistency & Argmax & 100.00 & 51.67 & 100.00 & 51.64 & 100.00 & 50.08 & 100.00 & 50.10 & 100.00 & 53.23 & 100.00 & 51.13 & 100.00 & 51.43 & 100.00 & 50.54 & 100.00 & 52.48 & 100.00 & 51.82  \\
    Consistency & TTA entropy & 92.08 & 54.79 & 91.95 & 54.87 & 94.78 & 49.97 & 94.58 & 50.19 & 91.82 & 55.72 & 89.35 & 58.73 & 93.92 & 51.64 & 92.37 & 54.55 & 92.21 & 55.12 & 92.76 & 53.64  \\
    Consistency & \textbf{TC (Ours)} & \textbf{78.30} & 62.29 & \textbf{78.19} & 62.34 & 93.38 & 50.39 & 93.42 & 50.69 & 79.42 & 63.46 & \textbf{69.26} & \textbf{67.37} & 89.46 & 55.75 & 84.43 & 59.82 & \textbf{79.39} & 62.81 & \textbf{84.04} & 59.41  \\
    \bottomrule
  \end{tabular}}
\end{table}

\subsection{Detailed fusion ablations}

\subsubsection{Clean-OOD}
\label{app:ablation_baseline}

\begin{table}[H]
  \centering
  \scriptsize
  \caption{Clean-OOD branch-combination results on ImageNet with ResNet-50.}
  \label{tab:triscore_ablation_resnet50_fsood}
  {%
  \begin{tabular}{l*{10}{c}}
    \toprule
    Method &
    \hdr{Near-OOD}{NINCO}{FPR@95} &
    \hdr{Near-OOD}{NINCO}{AUROC} &
    \hdr{Near-OOD}{SSB-Hard}{FPR@95} &
    \hdr{Near-OOD}{SSB-Hard}{AUROC} &
    \hdr{Far-OOD}{iNaturalist}{FPR@95} &
    \hdr{Far-OOD}{iNaturalist}{AUROC} &
    \hdr{Far-OOD}{OpenImage-O}{FPR@95} &
    \hdr{Far-OOD}{OpenImage-O}{AUROC} &
    \hdr{Far-OOD}{Textures}{FPR@95} &
    \hdr{Far-OOD}{Textures}{AUROC} \\
    \midrule
    E   & 59.83 & 80.29 & 76.27 & 72.42 & \textbf{30.51} & \textbf{91.14} & \textbf{37.77} & \textbf{89.19} & \textbf{46.25} & \textbf{88.49}  \\
    E+TC & 57.28 & 80.38 & 75.95 & 71.98 & 31.68 & 87.16 & 37.99 & 84.97 & 46.50 & 83.40  \\
    E+BP & 53.97 & \textbf{80.80} & \textbf{73.45} & \textbf{73.05} & 40.26 & 85.56 & 43.51 & 84.84 & 60.38 & 80.26  \\
    BP  & 60.32 & 77.23 & 77.69 & 70.24 & 63.22 & 76.27 & 65.43 & 76.70 & 88.86 & 68.20  \\
    BP+TC & 60.14 & 78.00 & 77.48 & 70.22 & 61.10 & 78.82 & 64.16 & 77.15 & 88.39 & 70.35  \\
    TC   & 55.00 & 78.47 & 75.12 & 69.58 & 41.68 & 80.41 & 49.69 & 77.11 & 61.97 & 72.85  \\
    E+BP+TC & \textbf{53.69} & 80.30 & 76.16 & 69.91 & 31.93 & 87.51 & 38.61 & 84.12 & 47.47 & 83.13  \\
    TriScore & 53.70 & 80.17 & 76.16 & 69.91 & 31.93 & 87.47 & 38.64 & 84.22 & 47.48 & 83.28  \\
    \bottomrule
  \end{tabular}}
\end{table}

\begin{table}[H]
  \centering
  \scriptsize
  \caption{Clean-OOD branch-combination results on ImageNet with RegNet.}
  \label{tab:triscore_ablation_regnet_fsood}
  \resizebox{\textwidth}{!}{%
  \begin{tabular}{l*{10}{c}}
    \toprule
    Method &
    \hdr{Near-OOD}{NINCO}{FPR@95} &
    \hdr{Near-OOD}{NINCO}{AUROC} &
    \hdr{Near-OOD}{SSB-Hard}{FPR@95} &
    \hdr{Near-OOD}{SSB-Hard}{AUROC} &
    \hdr{Far-OOD}{iNaturalist}{FPR@95} &
    \hdr{Far-OOD}{iNaturalist}{AUROC} &
    \hdr{Far-OOD}{OpenImage-O}{FPR@95} &
    \hdr{Far-OOD}{OpenImage-O}{AUROC} &
    \hdr{Far-OOD}{Textures}{FPR@95} &
    \hdr{Far-OOD}{Textures}{AUROC} \\
    \midrule
    E   & 42.55 & \textbf{91.67} & 62.09 & \textbf{85.28} & \textbf{7.72} & \textbf{98.29} & 25.95 & \textbf{95.83} & 38.18 & \textbf{93.02}  \\
    E+TC & 39.01 & 90.36 & 62.04 & 83.04 & 11.43 & 95.97 & \textbf{22.76} & 93.95 & 37.98 & 91.24  \\
    E+BP & \textbf{38.26} & 90.31 & \textbf{57.99} & 83.21 & 18.23 & 95.94 & 25.38 & 94.32 & 40.61 & 90.38  \\
    BP  & 54.59 & 82.11 & 72.73 & 74.86 & 49.39 & 83.90 & 48.42 & 84.43 & 65.60 & 80.39  \\
    BP+TC & 54.54 & 82.56 & 72.76 & 74.87 & 47.39 & 85.42 & 47.67 & 84.76 & 65.35 & 81.65  \\
    TC   & 48.79 & 83.53 & 71.03 & 75.30 & 27.51 & 87.52 & 37.38 & 84.69 & 47.17 & 84.40  \\
    E+BP+TC & 40.25 & 89.77 & 64.34 & 80.94 & 14.62 & 95.69 & 23.54 & 93.68 & \textbf{33.12} & 92.02  \\
    TriScore & 40.28 & 89.42 & 64.36 & 80.87 & 13.20 & 95.15 & 23.62 & 93.50 & 33.31 & 91.58  \\
    \bottomrule
  \end{tabular}}
\end{table}

\begin{table}[H]
  \centering
  \scriptsize
  \caption{Clean-OOD branch-combination results on ImageNet with ViT-B/16.  }
  \label{tab:triscore_ablation_vitb16_fsood}
  \resizebox{\textwidth}{!}{%
  \begin{tabular}{l*{10}{c}}
    \toprule
    Method &
    \hdr{Near-OOD}{NINCO}{FPR@95} &
    \hdr{Near-OOD}{NINCO}{AUROC} &
    \hdr{Near-OOD}{SSB-Hard}{FPR@95} &
    \hdr{Near-OOD}{SSB-Hard}{AUROC} &
    \hdr{Far-OOD}{iNaturalist}{FPR@95} &
    \hdr{Far-OOD}{iNaturalist}{AUROC} &
    \hdr{Far-OOD}{OpenImage-O}{FPR@95} &
    \hdr{Far-OOD}{OpenImage-O}{AUROC} &
    \hdr{Far-OOD}{Textures}{FPR@95} &
    \hdr{Far-OOD}{Textures}{AUROC} \\
    \midrule
    E   & 94.16 & 66.02 & 92.25 & 58.80 & 83.58 & 79.30 & 88.79 & 76.48 & 83.65 & 81.17  \\
    E+TC & 92.75 & 74.06 & 91.38 & 64.90 & 70.60 & 85.60 & 85.33 & 81.59 & 78.72 & 83.09  \\
    E+BP & 79.72 & 76.48 & 86.85 & 67.15 & 57.11 & 84.18 & 68.41 & 82.27 & 69.12 & 82.69  \\
    BP  & 57.88 & 79.59 & 82.42 & 68.94 & 52.28 & 81.40 & 53.77 & 80.93 & 72.90 & 76.66  \\
    BP+TC & \textbf{55.94} & 81.52 & \textbf{82.39} & \textbf{69.18} & 45.92 & 85.36 & 49.63 & 82.67 & 67.14 & 79.05  \\
    TC   & 57.76 & \textbf{81.66} & 86.05 & 67.92 & \textbf{35.13} & 87.44 & \textbf{45.46} & 83.21 & 53.67 & 81.07  \\
    E+BP+TC & 78.42 & 79.29 & 88.56 & 67.74 & 37.88 & \textbf{89.17} & 55.65 & \textbf{85.68} & \textbf{53.01} & \textbf{86.54}  \\
    TriScore & 77.99 & 79.07 & 88.55 & 67.66 & 37.82 & 88.74 & 55.24 & 85.25 & 53.05 & 85.93  \\
    \bottomrule
  \end{tabular}}
\end{table}

\begin{table}[H]
  \centering
  \scriptsize
  \caption{Clean-OOD branch-combination results on ImageNet with Swin-T.}
  \label{tab:triscore_ablation_swint_fsood}
  \resizebox{\textwidth}{!}{%
  \begin{tabular}{l*{10}{c}}
    \toprule
    Method &
    \hdr{Near-OOD}{NINCO}{FPR@95} &
    \hdr{Near-OOD}{NINCO}{AUROC} &
    \hdr{Near-OOD}{SSB-Hard}{FPR@95} &
    \hdr{Near-OOD}{SSB-Hard}{AUROC} &
    \hdr{Far-OOD}{iNaturalist}{FPR@95} &
    \hdr{Far-OOD}{iNaturalist}{AUROC} &
    \hdr{Far-OOD}{OpenImage-O}{FPR@95} &
    \hdr{Far-OOD}{OpenImage-O}{AUROC} &
    \hdr{Far-OOD}{Textures}{FPR@95} &
    \hdr{Far-OOD}{Textures}{AUROC} \\
    \midrule
    E   & 79.16 & 78.25 & 87.45 & 68.21 & 61.52 & 85.13 & 80.81 & 79.86 & 84.45 & 78.96  \\
    E+TC & 77.44 & 80.30 & 87.16 & 69.55 & 54.52 & 87.00 & 78.30 & 81.40 & 82.39 & 79.74  \\
    E+BP & 63.14 & 82.01 & 83.56 & \textbf{71.21} & 45.05 & 86.07 & 60.94 & 82.95 & 78.51 & 78.62  \\
    BP  & 60.92 & 79.45 & 82.87 & 69.04 & 54.24 & 80.63 & 57.85 & 79.82 & 83.24 & 72.41  \\
    BP+TC & 57.45 & 81.53 & \textbf{82.78} & 69.74 & 45.14 & 84.75 & 52.14 & 81.55 & 79.50 & 75.46  \\
    TC   & \textbf{55.08} & 81.71 & 83.80 & 69.14 & 35.23 & 86.60 & \textbf{47.15} & 81.84 & 61.65 & 78.42  \\
    E+BP+TC & 56.94 & \textbf{83.01} & 83.21 & 70.70 & \textbf{29.48} & \textbf{90.42} & 47.39 & \textbf{85.23} & 59.66 & \textbf{84.26}  \\
    TriScore & 56.98 & 82.47 & 83.21 & 70.51 & 29.52 & 89.81 & 47.46 & 84.97 & \textbf{59.63} & 83.95  \\
    \bottomrule
  \end{tabular}}
\end{table}

\subsubsection{Adversarial OOD}
\label{app:ablation_adversarial}

\begin{table}[H]
  \centering
  \scriptsize
  \caption{Adversarial OOD branch-combination results on ImageNet with ResNet-50.}
  \label{tab:triscore_ablation_resnet50_adversarial}
  \resizebox{\textwidth}{!}{%
  \begin{tabular}{l*{20}{c}}
    \toprule
    Method & \multicolumn{2}{c}{AA} & \multicolumn{2}{c}{APGD-CE} & \multicolumn{2}{c}{C\&W} & \multicolumn{2}{c}{DF} & \multicolumn{2}{c}{FGSM} & \multicolumn{2}{c}{M-PGD} & \multicolumn{2}{c}{PGD} & \multicolumn{2}{c}{Square} & \multicolumn{2}{c}{BIM} & \multicolumn{2}{c}{PGD-BPDA} \\
     & FPR@95 & AUROC & FPR@95 & AUROC & FPR@95 & AUROC & FPR@95 & AUROC & FPR@95 & AUROC & FPR@95 & AUROC & FPR@95 & AUROC & FPR@95 & AUROC & FPR@95 & AUROC & FPR@95 & AUROC \\
    \midrule
    E   & 96.17 & 50.19 & 96.29 & 50.20 & 89.62 & 59.62 & 88.14 & 61.00 & 77.76 & \textbf{73.43} & 92.12 & 52.70 & 93.83 & 49.51 & 88.62 & 62.51 & 96.83 & 44.40 & 96.03 & 43.83  \\
    E+TC & 93.19 & 59.92 & 93.22 & 59.97 & 89.56 & 57.93 & 88.16 & 58.98 & 76.00 & 69.79 & 89.32 & 60.58 & 91.24 & 59.10 & 88.30 & 60.58 & 94.45 & 58.84 & 93.59 & 57.27  \\
    E+BP & 87.86 & 60.62 & 87.86 & 60.64 & 87.48 & \textbf{61.11} & \textbf{85.87} & \textbf{62.19} & 74.34 & 71.50 & 85.11 & 59.71 & 84.94 & 59.24 & \textbf{86.34} & \textbf{65.21} & 90.54 & 54.68 & 88.26 & 55.68  \\
    BP  & \textbf{75.11} & \textbf{69.48} & \textbf{74.81} & 69.49 & \textbf{87.06} & 60.53 & 87.48 & 61.09 & 81.92 & 66.44 & \textbf{79.87} & 64.37 & \textbf{76.24} & 66.56 & 86.71 & 64.97 & \textbf{74.08} & 65.01 & \textbf{74.03} & 66.40  \\
    BP+TC & 75.18 & 69.47 & 74.90 & \textbf{69.51} & 87.10 & 58.89 & 87.51 & 59.39 & 81.75 & 66.15 & 79.94 & \textbf{65.35} & 76.36 & \textbf{66.95} & 86.74 & 62.50 & 74.21 & 67.75 & 74.19 & \textbf{67.58}  \\
    TC   & 82.85 & 66.39 & 83.08 & 66.45 & 89.86 & 55.51 & 89.69 & 55.98 & 78.82 & 64.14 & 83.56 & 64.61 & 81.91 & 65.48 & 87.90 & 58.10 & 79.72 & \textbf{68.83} & 81.40 & 66.93  \\
    E+BP+TC & 92.59 & 61.39 & 92.73 & 61.44 & 88.73 & 58.57 & 87.08 & 59.91 & 74.27 & 70.86 & 88.63 & 59.98 & 89.62 & 59.25 & 87.27 & 62.37 & 94.83 & 56.37 & 92.82 & 56.49  \\
    TriScore & 92.12 & 63.58 & 92.30 & 63.65 & 89.08 & 58.85 & 87.44 & 60.16 & \textbf{74.06} & 71.15 & 88.34 & 61.40 & 89.39 & 61.61 & 87.40 & 62.65 & 94.03 & 58.76 & 92.27 & 59.25  \\
    \bottomrule
  \end{tabular}}
\end{table}

\begin{table}[H]
  \centering
  \scriptsize
  \caption{Adversarial OOD branch-combination results on ImageNet with RegNet.}
  \label{tab:triscore_ablation_regnet_adversarial}
  \resizebox{\textwidth}{!}{%
  \begin{tabular}{l*{20}{c}}
    \toprule
    Method & \multicolumn{2}{c}{AA} & \multicolumn{2}{c}{APGD-CE} & \multicolumn{2}{c}{C\&W} & \multicolumn{2}{c}{DF} & \multicolumn{2}{c}{FGSM} & \multicolumn{2}{c}{M-PGD} & \multicolumn{2}{c}{PGD} & \multicolumn{2}{c}{Square} & \multicolumn{2}{c}{BIM} & \multicolumn{2}{c}{PGD-BPDA} \\
     & FPR@95 & AUROC & FPR@95 & AUROC & FPR@95 & AUROC & FPR@95 & AUROC & FPR@95 & AUROC & FPR@95 & AUROC & FPR@95 & AUROC & FPR@95 & AUROC & FPR@95 & AUROC & FPR@95 & AUROC \\
    \midrule
    E   & 80.84 & 71.82 & 80.53 & 71.85 & 86.99 & 66.25 & 87.16 & 66.17 & 79.07 & \textbf{73.26} & 83.12 & 71.50 & 84.02 & 68.27 & 83.47 & 70.84 & 80.70 & 72.28 & 81.68 & 70.69  \\
    E+TC & 81.11 & 68.40 & 80.96 & 68.38 & 87.26 & 63.30 & 87.44 & 63.22 & 79.69 & 69.80 & 83.52 & 67.96 & 84.30 & 65.30 & 83.68 & 67.28 & 81.23 & 68.86 & 82.22 & 67.42  \\
    E+BP & 69.84 & 72.46 & 70.37 & 72.45 & 80.82 & 65.89 & 81.24 & 65.84 & 74.79 & 71.49 & 75.20 & 70.72 & 77.30 & 68.62 & 75.81 & 70.01 & 71.03 & 73.03 & 73.39 & 71.19  \\
    BP  & 74.48 & 68.01 & 74.22 & 67.99 & 83.30 & 61.44 & 83.73 & 61.48 & 82.46 & 64.97 & 77.14 & 65.28 & 81.58 & 64.43 & 80.80 & 64.34 & 75.08 & 68.82 & 78.06 & 66.89  \\
    BP+TC & 75.87 & 65.10 & 75.50 & 65.08 & 84.16 & 59.16 & 84.58 & 59.18 & 83.40 & 62.92 & 78.25 & 63.07 & 82.42 & 61.98 & 81.68 & 61.80 & 76.30 & 65.91 & 79.04 & 64.13  \\
    TC   & 92.07 & 54.57 & 92.02 & 54.59 & 95.16 & 50.07 & 95.27 & 50.02 & 93.18 & 55.64 & 91.54 & 55.11 & 94.19 & 52.80 & 93.23 & 52.38 & 92.04 & 55.76 & 92.97 & 54.17  \\
    E+BP+TC & 81.07 & 67.27 & 81.09 & 67.28 & 87.37 & 61.85 & 87.60 & 61.87 & 79.78 & 69.20 & 82.92 & 67.29 & 84.57 & 64.32 & 83.29 & 66.19 & 81.21 & 68.12 & 82.43 & 66.34  \\
    TriScore & \textbf{65.18} & \textbf{75.36} & \textbf{64.97} & \textbf{75.36} & \textbf{78.99} & \textbf{66.67} & \textbf{79.20} & \textbf{66.85} & \textbf{72.27} & 72.07 & \textbf{70.80} & \textbf{72.39} & \textbf{73.95} & \textbf{70.57} & \textbf{72.66} & \textbf{71.54} & \textbf{65.75} & \textbf{75.81} & \textbf{68.85} & \textbf{73.76}  \\
    \bottomrule
  \end{tabular}}
\end{table}

\begin{table}[H]
  \centering
  \scriptsize
  \caption{Adversarial OOD branch-combination results on ImageNet with ViT-B/16.}
  \label{tab:triscore_ablation_vitb16_adversarial}
  \resizebox{\textwidth}{!}{%
  \begin{tabular}{l*{20}{c}}
    \toprule
    Method & \multicolumn{2}{c}{AA} & \multicolumn{2}{c}{APGD-CE} & \multicolumn{2}{c}{C\&W} & \multicolumn{2}{c}{DF} & \multicolumn{2}{c}{FGSM} & \multicolumn{2}{c}{M-PGD} & \multicolumn{2}{c}{PGD} & \multicolumn{2}{c}{Square} & \multicolumn{2}{c}{BIM} & \multicolumn{2}{c}{PGD-BPDA} \\
     & FPR@95 & AUROC & FPR@95 & AUROC & FPR@95 & AUROC & FPR@95 & AUROC & FPR@95 & AUROC & FPR@95 & AUROC & FPR@95 & AUROC & FPR@95 & AUROC & FPR@95 & AUROC & FPR@95 & AUROC \\
    \midrule
    E   & 92.13 & 59.70 & 92.06 & 59.69 & 94.34 & 51.74 & 94.44 & 51.62 & 89.88 & 64.17 & 88.64 & 69.83 & 93.11 & 55.97 & 91.01 & 64.62 & 91.76 & 60.07 & 92.41 & 58.27  \\
    E+TC & 91.38 & 59.28 & 91.35 & 59.29 & 94.33 & 52.09 & 94.36 & 52.06 & 88.75 & 63.16 & 87.04 & 68.23 & 92.74 & 55.94 & 89.68 & 63.55 & 91.04 & 59.60 & 91.89 & 57.94  \\
    E+BP & 84.11 & 61.39 & 84.30 & 61.42 & 92.58 & 53.33 & 92.69 & 53.43 & 81.65 & 64.32 & 79.65 & 68.16 & 88.86 & 58.02 & 85.68 & 63.80 & 84.44 & 61.92 & 86.66 & 59.98  \\
    BP  & 86.18 & 59.19 & 86.33 & 59.21 & 92.59 & 53.27 & 92.82 & 53.49 & 85.81 & 60.71 & 82.89 & 62.57 & 89.48 & 57.08 & 87.97 & 59.25 & 86.67 & 59.98 & 88.14 & 58.36  \\
    BP+TC & 86.20 & 58.45 & 86.34 & 58.48 & 92.59 & 53.03 & 92.82 & 53.25 & 85.81 & 59.88 & 82.91 & 61.71 & 89.49 & 56.45 & 87.98 & 58.67 & 86.68 & 59.07 & 88.13 & 57.62  \\
    TC   & 88.18 & 55.31 & 88.20 & 55.28 & 94.23 & 51.52 & 94.17 & 51.72 & 86.51 & 58.01 & 80.24 & 61.96 & 92.43 & 53.60 & 85.78 & 57.33 & 88.78 & 55.67 & 90.99 & 54.49  \\
    E+BP+TC & 88.32 & 60.37 & 87.98 & 60.41 & 94.00 & 52.41 & 94.04 & 52.65 & 83.39 & 64.67 & 79.92 & 70.18 & 91.68 & 56.77 & 84.93 & 65.16 & 88.10 & 60.74 & 90.25 & 58.79  \\
    TriScore & \textbf{82.07} & \textbf{64.74} & \textbf{81.64} & \textbf{64.77} & \textbf{92.09} & \textbf{54.00} & \textbf{92.44} & \textbf{54.03} & \textbf{77.72} & \textbf{67.66} & \textbf{75.92} & \textbf{72.22} & \textbf{87.69} & \textbf{59.72} & \textbf{84.20} & \textbf{66.65} & \textbf{82.23} & \textbf{65.07} & \textbf{84.99} & \textbf{62.29}  \\
    \bottomrule
  \end{tabular}}
\end{table}

\begin{table}[H]
  \centering
  \scriptsize
  \caption{Adversarial OOD branch-combination results on ImageNet with Swin-T.}
  \label{tab:triscore_ablation_swint_adversarial}
  \resizebox{\textwidth}{!}{%
  \begin{tabular}{l*{20}{c}}
    \toprule
    Method & \multicolumn{2}{c}{AA} & \multicolumn{2}{c}{APGD-CE} & \multicolumn{2}{c}{C\&W} & \multicolumn{2}{c}{DF} & \multicolumn{2}{c}{FGSM} & \multicolumn{2}{c}{M-PGD} & \multicolumn{2}{c}{PGD} & \multicolumn{2}{c}{Square} & \multicolumn{2}{c}{BIM} & \multicolumn{2}{c}{PGD-BPDA} \\
     & FPR@95 & AUROC & FPR@95 & AUROC & FPR@95 & AUROC & FPR@95 & AUROC & FPR@95 & AUROC & FPR@95 & AUROC & FPR@95 & AUROC & FPR@95 & AUROC & FPR@95 & AUROC & FPR@95 & AUROC \\
    \midrule
    E   & 91.90 & 58.76 & 92.01 & 58.80 & 93.30 & 55.33 & 93.32 & 55.15 & 90.20 & 62.79 & 89.70 & 63.45 & 92.13 & 58.62 & 91.26 & 60.27 & 90.85 & 60.86 & 91.40 & 59.99  \\
    E+TC & 91.76 & 58.66 & 91.90 & 58.70 & 93.27 & 54.76 & 93.29 & 54.69 & 90.31 & 61.97 & 89.43 & 62.09 & 92.02 & 57.96 & 90.81 & 59.58 & 90.66 & 60.46 & 91.47 & 59.27  \\
    E+BP & 83.75 & 66.06 & 83.32 & 66.10 & 90.88 & \textbf{58.02} & 90.73 & 58.09 & 84.21 & 66.42 & 82.42 & 66.60 & 87.84 & 62.20 & 85.99 & 64.09 & 81.97 & 67.44 & 84.75 & 65.04  \\
    BP  & 78.83 & 67.45 & 78.99 & 67.44 & \textbf{89.52} & 57.66 & 89.74 & 57.99 & 82.28 & 64.99 & 79.16 & 64.50 & 85.92 & 61.55 & 85.69 & 62.90 & 79.71 & 67.91 & 82.41 & 64.90  \\
    BP+TC & 83.12 & 61.60 & 82.80 & 61.58 & 91.33 & 53.97 & 91.45 & 54.13 & 85.56 & 60.49 & 85.07 & 58.49 & 88.92 & 56.86 & 88.33 & 57.25 & 83.12 & 62.06 & 86.06 & 59.38  \\
    TC   & 93.22 & 50.47 & 93.30 & 50.43 & 95.08 & 48.32 & 95.28 & 48.19 & 92.74 & 51.79 & 94.32 & 48.66 & 94.60 & 49.01 & 94.81 & 48.14 & 93.14 & 50.73 & 93.90 & 49.50  \\
    E+BP+TC & 88.56 & 61.75 & 88.47 & 61.80 & 92.98 & 54.20 & 93.00 & 54.28 & 86.50 & 64.84 & 84.45 & 65.78 & 90.72 & 58.62 & 88.30 & 61.22 & 87.01 & 63.23 & 88.96 & 60.81  \\
    TriScore & \textbf{74.60} & \textbf{69.83} & \textbf{74.60} & \textbf{69.85} & 89.60 & 58.01 & \textbf{89.60} & \textbf{58.28} & \textbf{76.72} & \textbf{68.70} & \textbf{73.77} & \textbf{68.46} & \textbf{84.50} & \textbf{62.87} & \textbf{82.98} & \textbf{65.19} & \textbf{74.59} & \textbf{70.70} & \textbf{79.11} & \textbf{67.20}  \\
    \bottomrule
  \end{tabular}}
\end{table}

\section{Adaptive Adversary Attacks}

\setlength{\tabcolsep}{6pt} % slightly tighter

\label{app:adaptive}

\subsection{Adaptive threat model}

We evaluate a white-box adversary with full access to TriScore and its calibration parameters. The attacker ascends a dual objective, maximizing cross-entropy on the true label while minimizing the TriScore, forcing it to simultaneously flip the prediction and push the input into the detector's ID region.

The Energy and Transform-Consistency branches are differentiated exactly.  Since the Boundary-Probe denominator $\Vert{}g_t\Vert{}_2^2$ requires second-order derivatives, we apply a BPDA/stop-gradient approximation. The denominator is recomputed on each forward pass but treated as a constant during the backward pass, allowing the attack gradient to flow directly through the margin $\Delta$. To ensure $\lambda$ acts as a strict fool-versus-evade trade-off, we normalize the gradient of each objective to unit norm before combination. The attack executes 40 steps of $\ell_\infty$ PGD ($\epsilon=4/255$) on a strided 2,000-image ID subset per backbone. We verified beforehand that this differentiable implementation exactly reproduces the deployed per-branch score distributions on clean data.

\subsection{Adaptive attack results}
Table~\ref{tab:adaptive} reports the end-to-end attack success via the JOINT@5 metric. The fraction of inputs that are simultaneously misclassified and undetected at a 5\% FPR threshold. We sweep $\lambda$ to locate the adaptive attacker's optimum, with $\lambda=0$ serving as the baseline for a pure classifier attack.

\begin{table}[H]
  \centering
  \scriptsize
  \caption{Adaptive attack: JOINT@5 (\%) versus the trade-off $\lambda$, per backbone. $\lambda{=}0$ is non-adaptive.}
  \label{tab:adaptive}
  \begin{tabular}{lcccc}
    \toprule
    $\lambda$ & ResNet-50 & RegNet & ViT-B/16 & Swin-T \\
    \midrule
    0.00 & 77.0 & 44.7 & 33.2 & 16.3 \\
    0.25 & \textbf{96.0} & \textbf{97.2} & 82.4 & 98.9 \\
    0.50 & 95.3 & 97.1 & \textbf{86.5} & \textbf{99.3} \\
    1.00 & 70.7 & 87.4 & 80.6 & 97.4 \\
    2.00 & 26.2 & 17.6 & 24.7 & 31.0 \\
    4.00 & 24.2 & 14.8 & 20.2 & 21.6 \\
    \bottomrule
  \end{tabular}
\end{table}

At its best operating point the adaptive attacker makes $86$-$99\%$ of inputs simultaneously misclassified and undetected. This is not specific to TriScore: on the same TriScore-targeted perturbed images, standard differentiable baselines (EBO, MLS, MSP, GEN, GradNorm) also show high transfer evasion ($\ge 99\%$ at $5\%$ ID FPR). Because these attacks were not separately optimized against each baseline, we read this as transfer evasion and we claim that TriScore is not robust to the evaluated detector-aware attack (and it is unlikely to exceed what it is based on). Non-adaptive robustness (the $\lambda{=}0$ row) varies by backbone as a pure classifier attack already evades $77\%$ on ResNet-50 but only $16\%$ on Swin-T, consistent with our transfer-robustness framing.

\subsection{Attack decomposition}
Table~\ref{tab:adaptive_decomp} decomposes the attack at the per-backbone best $\lambda$ on the population the classifier initially classifies correctly (the standard reporting population, the full strided subset gives essentially identical numbers because misclassification is near-total). The attack is a single $40$-step trajectory with no restarts, and being sign-based it saturates the budget, so mean and max per-image $\ell_\infty$ both equal $4/255$. Misclassification is near-total on every backbone ($\ge 99.7\%$), so the binding constraint is detector evasion, which ranges from $85\%$ (ViT-B/16) to $99\%$ (Swin-T); the joint success closely matches the swept optimum of Table~\ref{tab:adaptive}, confirming the break holds on the correctly-classified population. As an ID-only check that the transform-consistency branch uses label-preserving views, the last two columns report the fraction of clean ID inputs whose prediction is unchanged under the flip and $0.9$-crop views: $86$-$97\%$ across backbones, i.e.\ the geometric transforms preserve the ID label on the large majority of inputs, as intended.

\begin{table}[H]
  \centering
  \scriptsize
  \caption{Decomposition of the adaptive attack at the best $\lambda$, on the initially-correctly-classified population. \emph{Mis.}: classifier misclassification rate, \emph{Evade@5}/\emph{Joint@5}: detector evasion and joint success at $5\%$ ID FPR, $\ell_\infty$: realized per-image budget. \emph{LP-flip}/\emph{LP-crop}: ID-only label-preservation rate of the flip and crop views.}
  \label{tab:adaptive_decomp}
  \begin{tabular}{lcccccccc}
    \toprule
    Backbone & $\lambda$ & $n$ & Mis.\ & Evade@5 & Joint@5 & $\ell_\infty$ & LP-flip & LP-crop \\
    \midrule
    ResNet-50 & 0.25 & 1512 & 99.8 & 96.2 & 96.0 & $4/255$ & 90.0 & 86.3 \\
    RegNet    & 0.25 & 1706 & 100.0 & 97.2 & 97.2 & $4/255$ & 96.6 & 95.2 \\
    ViT-B/16  & 0.50 & 1601 & 99.7 & 85.1 & 85.1 & $4/255$ & 96.8 & 90.0 \\
    Swin-T    & 0.50 & 1621 & 100.0 & 99.4 & 99.4 & $4/255$ & 96.6 & 90.8 \\
    \bottomrule
  \end{tabular}
\end{table}

\section{Additional Diagnostics}
\label{app:added_analyses}

% This appendix collects the analyses requested during review, reported for all four backbones wherever the underlying quantity is defined per backbone: Boundary-Probe gradient diagnostics and hook sensitivity (\ref{app:grad_diag}), within-module ablations of the design constants (\ref{app:within_ablation}), a detector-aware adaptive attack (\ref{app:adaptive}), fusions built from \emph{other} existing detectors (\ref{app:altfusion}), a branch-complementarity analysis (\ref{app:complementarity}), gate and calibration-seed sensitivity (\ref{app:sensitivity}), and a Boundary-Probe-versus-uncertainty diagnostic (\ref{app:bp_uncertainty}). All numbers use the same ID-only calibration and the standard-OOD protocol of Section~\ref{subsec:adv_results}; the default configuration reproduces Table~\ref{tab:adversarial_method_summary}. Where a table reports a per-column or per-panel best, it is shown in \textbf{bold}.

\subsection{Boundary-Probe diagnostics}
\label{app:grad_diag}

As discussed after Eq.~\eqref{eq:bp_distance}, the Boundary-Probe $d_t^2=\Delta^2/(\|g_t\|_2^2+\varepsilon)$ is a first-order fragility indicator in the coordinates of the feature map $h$. Table~\ref{tab:gradnorm} reports the distribution of the gradient norm $\|g_t(x)\|_2$ on the ID test split for each backbone. The norm is concentrated and bounded well away from zero on every backbone. The last column gives the fraction of samples for which the stabilizer $\varepsilon=10^{-8}$ is even marginally active ($\|g_t\|_2^2\le 10\varepsilon$), which is $0.00\%$ on every ID and OOD split for all four backbones. The guard never engages, and re-running with $\varepsilon\in\{10^{-6},10^{-10}\}$ leaves every AUROC unchanged to two decimals, so the score is insensitive to $\varepsilon$. Because each branch is z-scored on ID data before fusion (Eq.~\eqref{eq:zscore}), a global rescaling of $h$ cancels in $S_{\mathrm{BP},z}$.

\begin{table}[H]
  \centering
  \scriptsize
  \caption{Percentiles of the Boundary-Probe feature-gradient norm $\|g_t(x)\|_2$ on the ID test split and the fraction of samples at the numerical floor.}
  \label{tab:gradnorm}
  \begin{tabular}{lcccc}
    \toprule
    Backbone & 1st pct. & 50th pct. & 99th pct. & floor ($\|g\|_2^2\!\le\!10\varepsilon$) \\
    \midrule
    ResNet-50 & 0.12 & 0.20 & 0.40 & 0.00\% \\
    RegNet    & 1.96 & 5.81 & 21.6 & 0.00\% \\
    ViT-B/16  & 0.46 & 2.20 & 10.7 & 0.00\% \\
    Swin-T    & 1.08 & 5.03 & 23.2 & 0.00\% \\
    \bottomrule
  \end{tabular}
\end{table}

We report a hook location study on ResNet-50 (Table~\ref{tab:hook}). Moving the hook changes the headline by less than $2$ AUROC in either regime (clean $79.7$--$80.5$, adversarial $62.1$--$64.1$), so the score does not depend delicately on where the gradient is read. The deployed \texttt{layer4} follows the same a-priori rule: it is the penultimate representation feeding the classifier head on every backbone. Appendix~\ref{app:within_ablation} exhibits knobs for which a non-default value would have scored higher had we tuned on the attack set.

\begin{table}[H]
  \centering
  \scriptsize
  \caption{Hook-location sensitivity of Boundary-Probe on ResNet-50 AUROC scores.}
  \label{tab:hook}
  \begin{tabular}{lcc}
    \toprule
    Hook & Clean-OOD & Adversarial \\
    \midrule
    \texttt{layer2}            & 79.67 & 62.14 \\
    \texttt{layer3}            & 79.76 & 62.93 \\
    \texttt{layer4} (default)  & \textbf{80.50} & \textbf{64.08} \\
    \bottomrule
  \end{tabular}
\end{table}

% \begin{table}[H]
%   \centering
%   \scriptsize
%   \caption{Boundary-Probe gradient location on CIFAR-10 (ResNet-18): detection AUROC and compute cost for the four residual stages and the input-space default. Near-OOD $=\{$CIFAR-100, Tiny ImageNet$\}$, far-OOD $=\{$MNIST, SVHN, Textures$\}$; ID test is the full $10$k CIFAR-10 test set. \emph{Adversarial} is the AUROC for detecting PGD ($\ell_\infty$, $\varepsilon=4/255$) adversarial examples of the CIFAR-10 test set (classifier attack-success $97.7\%$) as an OOD split. Peak GPU memory (\texttt{max\_memory\_allocated}) and wall-time are over an identical batch-200 scoring pass on a single RTX~2080~Ti; absolute times are for relative comparison across hooks, not against the H100 timings of Table~\ref{tab:cifar10_timing_methods_sorted}.}
%   \label{tab:cifar10_hook_cost}
%   \begin{tabular}{lccccc}
%     \toprule
%     Gradient location & Near-OOD & Far-OOD & Adversarial & Peak GPU (MB) & Time (s) \\
%     \midrule
%     feature \texttt{layer1}    & 83.93 & 88.90 & 67.74 & 1098 & 40.6 \\
%     feature \texttt{layer2}    & 84.04 & 89.13 & 68.25 & 1073 & 39.8 \\
%     feature \texttt{layer3}    & \textbf{84.41} & 90.00 & \textbf{69.13} & 1061 & 38.5 \\
%     feature \texttt{layer4}    & 83.06 & \textbf{90.35} & 61.90 & 1055 & \textbf{36.7} \\
%     input space                & 83.87 & 88.89 & 67.56 & \textbf{1048} & \textbf{36.7} \\
%     \bottomrule
%   \end{tabular}
% \end{table}

\subsection{Boundary-Probe and Softmax uncertainty}
\label{app:bp_uncertainty}

Since BP is designed to scale with local fragility, it naturally responds to hard ID examples. A valid concern is whether BP simply re-measures standard classifier uncertainty. Table~\ref{tab:bp_unc} addresses this in two experiments that measure different endpoints. {Panel~A} (clean) reports the AUROC for detecting \emph{ID misclassification}, i.e.\ ranking incorrect above correct clean ID predictions. {Panel~B} (adversarial) reports OOD-versus-ID detection under attack. In both panels we compare BP, $1-\mathrm{MSP}$, and the residual BP$\mid$MSP of BP after a linear regression on $1-\mathrm{MSP}$. The regression is fit on the ID calibration split only, then frozen and evaluated on the disjoint ID-test and adversarial samples. 

On clean data, BP is largely redundant with uncertainty: it correlates strongly with MSP ($\rho \in [-0.88, -0.68]$ across backbones) and predicts ID misclassification at $74$-$81$ AUROC, comparable to $1-\mathrm{MSP}$, while its MSP-orthogonal residual is near chance. Under adversarial shift, the picture changes: BP outperforms $1-\mathrm{MSP}$ on all four backbones, and the frozen residual BP$\mid$MSP still separates OOD at $55$-$70$ AUROC. Linear residualization only establishes orthogonality to a linear function of MSP on the fitting distribution, which shows that BP retains non-redundant ranking information beyond a linear function of MSP under attack.

\begin{table}[H]
  \centering
  \scriptsize
  \caption{Boundary-Probe versus softmax uncertainty (AUROC). "A" (clean): detecting ID misclassification. "B" (adversarial): OOD-versus-ID detection under attack.}
  \label{tab:bp_unc}
  \begin{tabular}{llccc}
    \toprule
    Panel & Backbone & BP & $1-\mathrm{MSP}$ & BP$\mid$MSP \\
    \midrule
    \multirow{4}{*}{\shortstack[l]{A: ID\\misclass.\\(clean)}}
      & ResNet-50 & \textbf{78.4} & 77.5 & 58.7 \\
      & RegNet    & 81.4 & \textbf{83.9} & 55.2 \\
      & ViT-B/16  & 75.7 & 75.7 & 56.1 \\
      & Swin-T    & 73.5 & \textbf{77.8} & 51.8 \\
    \midrule
    \multirow{4}{*}{\shortstack[l]{B: OOD\\det.\\(adv.)}}
      & ResNet-50 & \textbf{60.1} & 59.7 & 55.2 \\
      & RegNet    & \textbf{70.5} & 61.4 & 70.0 \\
      & ViT-B/16  & \textbf{62.7} & 58.4 & 61.8 \\
      & Swin-T    & \textbf{68.2} & 60.0 & 66.4 \\
    \bottomrule
  \end{tabular}
\end{table}

\subsection{Complementarity of the branches}
\label{app:complementarity}

Table~\ref{tab:rho} reports the Spearman rank correlation among the three standardized branches on clean ID data versus under PGD attack. On the transformer and RegNet backbones the branches are already weakly correlated on clean data (most pairwise $\rho \le 0.5$) and stay weak-to-moderate under attack. On ResNet-50 they are more strongly correlated on clean data but several pairs decorrelate under attack (e.g.\ $\rho(E, TC)$ drops from $0.71$ to $0.40$). In neither regime is any pair strongly redundant, but correlation between continuous scores is only suggestive of complementarity. Table~\ref{tab:overlap} therefore makes the detection-error overlap explicit at a matched operating point. Thresholding each branch at a common $5\%$ ID false-positive rate on the pooled adversarial splits, we report the fraction each branch detects, the fraction it detects uniquely (missed by the other two), and the gated fusion at a matched $5\%$ total ID FPR. 

The results show two important points. First, no single branch dominates across backbones: the strongest branch is TC on ResNet-50, energy on RegNet, and BP on the transformers. Each branch uniquely flags $2$-$9\%$ of adversarial inputs that the other two miss, so their errors do not fully overlap. Second, at this stringent matched operating point, the gated fusion detects a fraction comparable to the best single branch. Its advantage is in ranking (AUROC, Table~\ref{tab:adversarial_method_summary}) and in cross-backbone robustness, because the strongest branch changes with the backbone and is not known a priori, so the gate tracks whichever branch is informative without having to select it. These results demonstrate that the branches are complementary and that the gate helps.

\begin{table}[H]
  \centering
  \scriptsize
  \caption{Spearman $\rho$ between standardized branches, on clean ID data and under PGD.}
  \label{tab:rho}
  \begin{tabular}{lcccccc}
    \toprule
    & \multicolumn{3}{c}{Clean ID} & \multicolumn{3}{c}{Under PGD} \\
    \cmidrule(lr){2-4}\cmidrule(lr){5-7}
    Backbone & $E+BP$ & $E+TC$ & $BP+TC$ & $E+BP$ & $E+TC$ & $BP+TC$ \\
    \midrule
    ResNet-50 & 0.85 & 0.71 & 0.89 & 0.74 & 0.40 & 0.68 \\
    RegNet    & 0.30 & 0.51 & 0.68 & 0.44 & 0.55 & 0.73 \\
    ViT-B/16  & 0.27 & 0.31 & 0.78 & 0.30 & 0.34 & 0.79 \\
    Swin-T    & 0.45 & 0.50 & 0.76 & 0.50 & 0.58 & 0.76 \\
    \bottomrule
  \end{tabular}
\end{table}

\begin{table}[H]
  \centering
  \scriptsize
  \caption{FPR-controlled branch error overlap on the pooled adversarial splits. \emph{Fused} is the gated TriScore at a matched $5\%$ total ID FPR. Left block: adversarial detection rate (\%). Right block: fraction detected only by that branch.}
  \label{tab:overlap}
  \begin{tabular}{lccccccc}
    \toprule
    & \multicolumn{4}{c}{Detection rate (\%)} & \multicolumn{3}{c}{Unique detection (\%)} \\
    \cmidrule(lr){2-5}\cmidrule(lr){6-8}
    Backbone & $E$ & BP & TC & Fused & only $E$ & only BP & only TC \\
    \midrule
    ResNet-50 & 7.4 & 8.8 & \textbf{13.3} & 13.2 & 3.6 & 3.3 & 8.9 \\
    RegNet    & \textbf{16.9} & 12.9 & 8.0 & 10.5 & 9.2 & 3.6 & 1.6 \\
    ViT-B/16  & 6.8 & \textbf{8.3} & 6.3 & 6.4 & 2.9 & 3.2 & 2.7 \\
    Swin-T    & 7.8 & \textbf{10.2} & 6.0 & 6.8 & 2.5 & 3.8 & 2.4 \\
    \bottomrule
  \end{tabular}
\end{table}

\subsection{Statistical reliability}
\label{app:bootstrap}

Appendix~\ref{app:sensitivity} bounds calibration-subset variability. In this subsection, we quantify finite-test-set uncertainty. For each backbone we draw $500$ paired bootstrap resamples of the evaluation images (the ID test set and, independently, each attack's OOD set are resampled with replacement, using the same indices for every method), recompute the $10$-attack mean AUROC on each resample, and report $95\%$ percentile intervals. Table~\ref{tab:bootstrap} gives the interval for TriScore, for GEN (the strongest baseline on the transformer backbones and the closest competitor overall among the post-hoc scores), and for the paired difference. 

The intervals are tight (half-width $\le 0.3$ AUROC), and the point estimates reproduce Table~\ref{tab:adversarial_method_summary} to within $0.04$. The paired TriScore-GEN difference is positive with a $95\%$ interval excluding zero on all four backbones, including the tightest case (Swin-T, $+0.44$ with interval $[0.07,0.82]$). Together with the 5-seed study of Appendix~\ref{app:sensitivity}, this resolves the $1$-$2$ point margin caveat for the reported comparisons.

\begin{table}[H]
  \centering
  \scriptsize
  \caption{Paired per-image bootstrap ($500$ resamples) of the adversarial $10$-attack mean AUROC. Brackets are $95\%$ percentile intervals, $\Delta$ is the paired TriScore-GEN difference (+ favours TriScore).}
  \label{tab:bootstrap}
  \begin{tabular}{lccc}
    \toprule
    Backbone & TriScore & GEN & $\Delta$ (TriScore$-$GEN) \\
    \midrule
    ResNet-50 & 62.14 [61.90, 62.40] & 55.58 [55.32, 55.87] & $+6.57$ [$6.21$, $6.93$] \\
    RegNet    & 72.05 [71.78, 72.33] & 70.17 [69.93, 70.41] & $+1.88$ [$1.51$, $2.26$] \\
    ViT-B/16  & 63.10 [62.83, 63.38] & 62.02 [61.73, 62.29] & $+1.09$ [$0.71$, $1.50$] \\
    Swin-T    & 65.90 [65.61, 66.18] & 65.46 [65.19, 65.76] & $+0.44$ [$0.07$, $0.82$] \\
    \bottomrule
  \end{tabular}
\end{table}

\section{Additional Evaluations}

\subsection{CIFAR-10 OOD detection}
\label{app:cifar}

Table~\ref{tab:cifar10_baseline} reports CIFAR-10 results with ResNet-18. The CIFAR results are included as additional small-scale evidence and a low-resolution stress test.

\setlength{\tabcolsep}{2pt} % slightly tighter

\begin{table}[H]
    \centering
    \scriptsize
    \caption{Clean-OOD detection results on CIFAR-10 with ResNet-18.}
    \label{tab:cifar10_baseline}
    {%
    \begin{tabular}{l*{10}{c}}
      \toprule
      Method &
      \hdr{Near-OOD}{Tiny ImageNet}{FPR@95} &
      \hdr{Near-OOD}{Tiny ImageNet}{AUROC} &
      \hdr{Far-OOD}{MNIST}{FPR@95} &
      \hdr{Far-OOD}{MNIST}{AUROC} &
      \hdr{Far-OOD}{Places365}{FPR@95} &
      \hdr{Far-OOD}{Places365}{AUROC} &
      \hdr{Far-OOD}{SVHN}{FPR@95} &
      \hdr{Far-OOD}{SVHN}{AUROC} &
      \hdr{Far-OOD}{Textures}{FPR@95} &
      \hdr{Far-OOD}{Textures}{AUROC} \\
      \midrule
      ASH         & 88.09 & 75.58 & 67.89 & 84.07 & 67.82 & 85.08 & 83.79 & 70.21 & 87.06 & 74.06  \\
      EBO         & 62.41 & 88.31 & 15.49 & 96.32 & 56.38 & 89.64 & 30.13 & 92.38 & 60.22 & 88.64  \\
      GEN         & 51.82 & 88.87 & 18.01 & 95.00 & 45.81 & 89.74 & 24.76 & 92.04 & 45.87 & 89.36  \\
      GradNorm    & 95.09 & 54.80 & 78.93 & 70.06 & 89.42 & 67.95 & 91.03 & 49.90 & 98.00 & 46.29  \\
      MLS         & 62.41 & 88.22 & 15.53 & 96.12 & 56.37 & 89.51 & 30.13 & 92.24 & 60.08 & 88.55  \\
      MSP         & 47.20 & 88.64 & 19.21 & 93.95 & 41.83 & 89.35 & 24.24 & 91.57 & 40.43 & 89.13  \\
      ODIN        & 84.03 & 80.95 & \textbf{15.39} & \textbf{96.60} & 76.46 & 83.86 & 68.81 & 84.52 & 83.01 & 83.86  \\
      ReAct       & 67.62 & 87.70 & 18.42 & 95.38 & 39.82 & 91.40 & 44.14 & 90.01 & 67.36 & 87.27  \\
      CADRef       & 33.98 & 90.10 & 25.58 & 92.18 & 33.82 & 90.05 & 29.80 & 90.71 & 26.34 & 91.76  \\
      DICE         & 66.20 & 78.83 & 45.62 & 82.01 & 77.78 & 74.02 & 30.79 & 91.06 & 68.33 & 79.28  \\
      KNN          & \textbf{29.60} & \textbf{91.83} & 21.33 & 93.74 & \textbf{30.92} & \textbf{91.61} & 23.69 & 92.25 & 23.34 & 93.49  \\
      NCI          & 40.19 & 89.80 & 29.21 & 91.63 & 34.72 & 90.33 & 37.59 & 89.50 & 27.69 & 92.15  \\
      PRO          & 31.20 & 90.34 & 32.96 & 88.60 & 34.40 & 89.42 & 22.91 & 91.94 & 30.86 & 89.88  \\
      SCALE        & 79.91 & 83.84 & 70.19 & 86.44 & 78.97 & 84.22 & 82.06 & 80.39 & 80.87 & 84.01  \\
      ViM          & 38.69 & 90.05 & 20.20 & 94.86 & 39.77 & 90.03 & \textbf{19.61} & \textbf{93.99} & \textbf{18.64} & \textbf{95.58}  \\
      TriScore & 30.56 & 90.49 & 27.43 & 91.53 & 32.47 & 89.96 & 20.25 & 92.73 & 42.24 & 88.02  \\
      \bottomrule
    \end{tabular}}
  \end{table}

  % ============================================================
  % BASELINE TABLES
\subsection{Open-Set recognition}
\label{app:osr}

This additional open-set recognition (OSR) evaluation checks whether the detector remains usable when the unknown samples are held-out semantic classes from a smaller classification problem. In OSR, the detector is tested against class-disjoint open-set splits CIFAR-6, CIFAR-50, TIN-20, and MNIST-6.

Table~\ref{tab:osr_seed1_all_methods} shows that TriScore remains competitive in the classic OSR setting, although this is not the strongest reason to adopt the method. Averaged over the four OSR splits, TriScore has a mean FPR@95 of 49.95, close to PRO at 48.18 and KNN at 48.98, while its mean AUROC is 83.47 compared with 84.85 for KNN and 84.69 for GEN. TriScore retains basic open-set behavior outside the main settings.

\setlength{\tabcolsep}{6pt} % slightly tighter

\begin{table}[H]
  \centering
  \scriptsize
  \caption{Open-set Recognition Results.}
  \label{tab:osr_seed1_all_methods}{%
  \begin{tabular}{l*{8}{c}}
    \toprule
    Method & \multicolumn{2}{c}{CIFAR-6} & \multicolumn{2}{c}{CIFAR-50} & \multicolumn{2}{c}{TIN-20} & \multicolumn{2}{c}{MNIST-6} \\
     & FPR@95 & AUROC & FPR@95 & AUROC & FPR@95 & AUROC & FPR@95 & AUROC \\
    \midrule
    ASH        & 75.14 & 84.76 & 54.25 & 82.14 & 65.57 & 73.74 & \textbf{8.65} & \textbf{97.79}  \\
    EBO        & 75.14 & 84.76 & 54.25 & 82.14 & 65.57 & 73.74 & \textbf{8.65} & \textbf{97.79}  \\
    GEN        & 73.77 & 85.16 & 54.03 & 82.24 & 65.13 & 73.79 & 9.75 & 97.57  \\
    GradNorm   & 91.92 & 63.61 & 86.28 & 68.13 & 81.54 & 69.65 & 36.16 & 93.50  \\
    MLS        & 75.14 & 84.74 & 54.14 & 82.10 & 65.13 & 73.64 & 8.77 & 97.74  \\
    MSP        & 70.35 & 85.51 & 54.92 & 80.48 & 65.02 & 71.66 & 12.41 & 96.08  \\
    ODIN       & 80.46 & 81.41 & 55.50 & 81.79 & 65.07 & 73.80 & 8.77 & 97.75  \\
    ReAct      & 78.36 & 82.53 & 62.09 & 80.49 & 70.58 & 73.04 & 26.87 & 95.70  \\
    NCI         & 87.14 & 82.58 & 53.08 & 82.83 & 66.91 & 73.07 & 39.22 & 91.92  \\
    ViM         & 82.09 & 83.04 & 57.38 & 77.74 & 67.30 & 72.55 & \textbf{8.65} & \textbf{97.79}  \\
    DICE        & 69.24 & 83.47 & 56.01 & 81.11 & \textbf{64.07} & \textbf{74.55} & 22.04 & 92.43  \\
    KNN         & \textbf{58.65} & \textbf{87.41} & 53.65 & \textbf{83.12} & 68.97 & 72.65 & 14.65 & 96.21  \\
    PRO         & 59.45 & \textbf{87.41} & 54.87 & 80.88 & 65.91 & 71.43 & 12.48 & 96.03  \\
    CADRef      & 66.34 & 86.16 & 54.63 & 81.90 & 65.85 & 73.35 & 14.23 & 96.44  \\
    SCALE       & 75.14 & 84.76 & 54.25 & 82.14 & 65.57 & 73.74 & \textbf{8.65} & \textbf{97.79}  \\
    TriScore   & 68.08 & 86.05 & \textbf{52.74} & 80.59 & 67.35 & 70.28 & 11.65 & 96.97  \\
    \bottomrule
  \end{tabular}}
\end{table}

\end{document}